\pdfoutput=1

\PassOptionsToPackage{sort,sort&compress}{natbib}

\documentclass[11pt]{article}
\usepackage[hyperref]{acl}
\usepackage[utf8]{inputenc}
\usepackage{graphicx}
\usepackage{svg}
\usepackage{times}
\usepackage{latexsym}
\usepackage[T1]{fontenc}
\usepackage{amsmath,amsfonts,amssymb,bm}
\usepackage{microtype}
\usepackage{inconsolata}

\usepackage[capitalise,nameinlink,noabbrev]{cleveref}
\usepackage{tabularx}
\usepackage{booktabs}
\usepackage{siunitx}
\usepackage{caption}
\usepackage{multirow}
\usepackage{wrapfig}
\usepackage{pifont,fontawesome}
\usepackage{import}
\usepackage{longtable}
\usepackage{etoolbox}
\usepackage{placeins}

\newcommand{\cmark}{\faCheck}%
\newcommand{\xmark}{\faTimes}%

\newcommand{\crefsubtable}[2]{\hyperref[#1]{\cref*{#1}#2}}

\usepackage{ulem}
\usepackage{pgffor}


\robustify\bfseries
\robustify\uline
\def\Uline#1{#1\llap{\uline{\phantom{#1}}}}

\makeatletter
\patchcmd{\@@setcref}         {??}{\color{orange} ??}{}{}
\patchcmd{\@@setcref}         {??}{\color{orange} ??}{}{}
\patchcmd{\@@setcrefrange}    {??}{\color{orange} ??}{}{}
\patchcmd{\@@setcrefrange}    {??}{\color{orange} ??}{}{}
\patchcmd{\@@setcrefrange}    {??}{\color{orange} ??}{}{}
\patchcmd{\@@setcrefrange}    {??}{\color{orange} ??}{}{}
\patchcmd{\@@setcrefrange}    {??}{\color{orange} ??}{}{}
\patchcmd{\@@setcrefrange}    {??}{\color{orange} ??}{}{}
\patchcmd{\@@setnamecref}     {??}{\color{orange} ??}{}{}
\patchcmd{\@@setnamecref}     {??}{\color{orange} ??}{}{}
\patchcmd{\@@setcpageref}     {??}{\color{orange} ??}{}{}
\patchcmd{\@@setcpageref}     {??}{\color{orange} ??}{}{}
\patchcmd{\@@setcpagerefrange}{??}{\color{orange} ??}{}{}
\patchcmd{\@@setcpagerefrange}{??}{\color{orange} ??}{}{}
\patchcmd{\@@setcpagerefrange}{??}{\color{orange} ??}{}{}
\patchcmd{\@@setcpagerefrange}{??}{\color{orange} ??}{}{}
\patchcmd{\@@setcpagerefrange}{??}{\color{orange} ??}{}{}
\patchcmd{\@@cref}            {??}{\color{orange} ??}{}{}
\makeatother

\crefformat{section}{#2§#1#3}
\crefformat{subsection}{#2§#1#3}

\usepackage{enumitem}
\setlist{itemsep=0pt,parsep=2pt, leftmargin=\labelwidth}

\def\eqref#1{equation~\ref{#1}}
\def\1{\bm{1}}

\def\vp{{\bm{p}}}

\DeclareMathAlphabet{\mathsfit}{\encodingdefault}{\sfdefault}{m}{sl}
\SetMathAlphabet{\mathsfit}{bold}{\encodingdefault}{\sfdefault}{bx}{n}

\def\sA{{\mathbb{A}}}

\title{Investigating the Role of Instruction Variety and Task Difficulty in Robotic Manipulation Tasks}

\author{Amit Parekh, {\bf Nikolas Vitsakis,} {\bf Alessandro Suglia,} {\bf Ioannis Konstas} \\ Heriot-Watt University \\ \texttt{\{amit.parekh, nv2006, a.suglia, i.konstas\}@hw.ac.uk}}

\usepackage{xspace,abbrevs}

\newabbrev{\GDG}{\textit{Gobbledygook}}
\newabbrev{\GDGWords}{\textit{Gobbledygook Words}}
\newabbrev{\GDGTokens}{\textit{Gobbledygook Tokens}}
\newabbrev{\LM}{Language Model (LM)}[LM]
\newabbrev{\VIMABench}{\textsc{VIMA-Bench}}
\newabbrev{\VL}{Vision+Language}
\newabbrev{\LLMs}{Large Language Models (LLMs)}[LLMs]

\makeatletter
\renewcommand\maybe@space@{%
  \maybe@ictrue % <= this is new
  \expandafter   \@tfor
    \expandafter \reserved@a
    \expandafter :%
    \expandafter =%
                 \nospacelist
                 \do \t@st@ic
  \ifmaybe@ic % <= this is new
    \space
  \fi
}
\makeatother

\begin{document}
\maketitle

\begin{abstract}

Evaluating the generalisation capabilities of multimodal models based solely on their performance on out-of-distribution data fails to capture their true robustness. This work introduces a comprehensive evaluation framework that systematically examines the role of instructions and inputs in the generalisation abilities of such models, considering architectural design, input perturbations across language and vision modalities, and increased task complexity. The proposed framework uncovers the resilience of multimodal models to extreme instruction perturbations and their vulnerability to observational changes, raising concerns about overfitting to spurious correlations. By employing this evaluation framework on current Transformer-based multimodal models for robotic manipulation tasks, we uncover limitations and suggest future advancements should focus on architectural and training innovations that better integrate multimodal inputs, enhancing a model's generalisation prowess by prioritising sensitivity to input content over incidental correlations.%
\footnote{Code available \url{https://github.com/amitkparekh/CoGeLoT}.}

\end{abstract}

\begin{figure*}[tb]
    \centering
    \includegraphics[width=\textwidth]{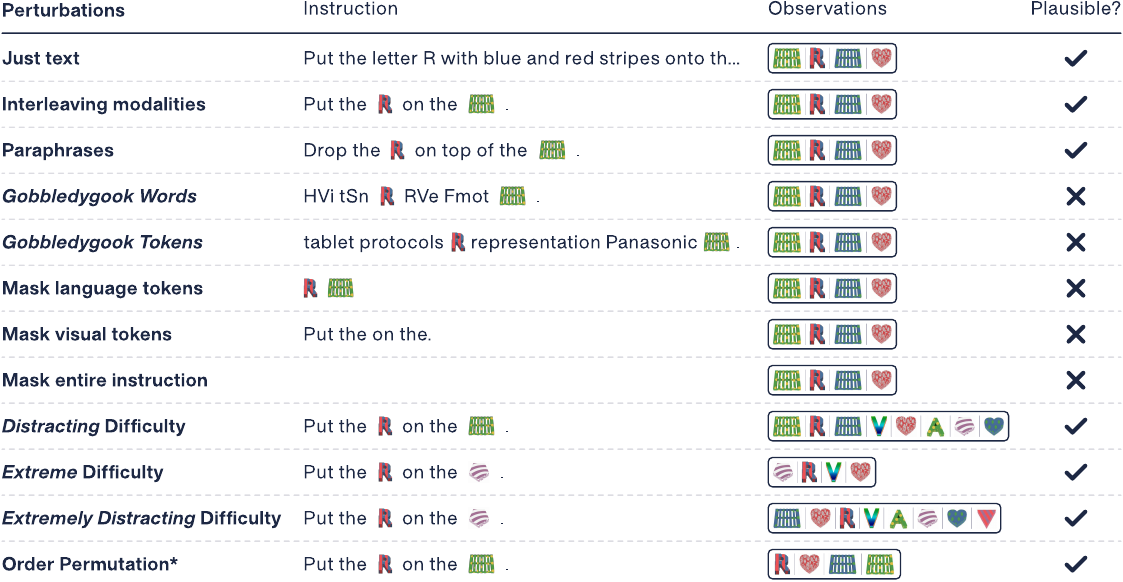}
    \caption{
    Our %proposed 
    evaluation framework. Each perturbation affects the instruction or observation inputs, which can be linguistic, visual, or a combination of both. The plausibility of a perturbation relates to a model's expected performance. Sensitivity to unreasonable conditions (\xmark) indicates that a model should not perform the task successfully given the perturbation, while plausible perturbations (\cmark) suggest that it should still perform successfully.
    }
    \label{fig:perturbation-table}
\end{figure*}

\section{Introduction}

Designing artificial agents to follow natural language instructions---to understand and act within the context of their environment---is a long-term goal of artificial intelligence \citep{winograd1972understanding}. 
An artificial agent should generalise to unseen scenarios by combining concepts and skills underpinning its training data in novel ways \citep{lake2017building}.

Previous work which proposed several language-guided tasks for tackling this challenge, largely focused on generalising to environments with different scenes from the training ones (e.g., \mbox{ALFRED}; \citealp{shridhar2020alfred}). 
However, relying solely on language for embodied action execution tasks can be inefficient, especially in collaborative settings with high ambiguity, such as visually cluttered scenes \cite{chiyah:sigdial2023,LiEtAl2023MasteringRobotManipulation,Chiyah-Garcia2024RepairsBlockWorld}. 
Multimodal prompts---instructions which interleave vision and language tokens---represent a way to specify commands which can be more flexible and specific than can be explained using text only \citep{Ma2024SurveyVisionLanguageActionModels,JiangEtAl2023VIMAGeneralRobot,Stone2023OpenWorldObjectManipulation}.
This capability is crucial for realistic human-robot collaboration tasks 
and can be viewed as analogous to pointing at objects within a scene \citep{Chen2021YouRefItEmbodiedReference, Islam2022CAESAREmbodiedSimulator}.
For this reason, \citet{JiangEtAl2023VIMAGeneralRobot} presented \VIMABench, the first benchmark aimed at studying several axes of generalisation involving novel concepts and tasks, with models receiving instructions combining both language and visual referents.

Many other benchmarks test for generalisation by solely looking at held-out examples \citep{OpenX-EmbodimentCollaboration2024OpenXEmbodimentRobotic,Stone2023OpenWorldObjectManipulation}. However, as highlighted by \citet{HupkesEtAl2023TaxonomyReviewGeneralization}, generalisation should be evaluated across multiple dimensions when creating truly robust models, capable of performing safely in varied environments.  
Inspired by these ideals, we assess generalisation along key axes such as structural, compositional, and robustness through specific covariate shifts (i.e., \textit{input perturbations}) as outlined in \cref{fig:perturbation-table}. Specifically, we looked at 1) an extensive set of linguistic perturbations on instructions, such as paraphrasing, corrupting the language content, and replacing visual referents with language descriptions; 2) masking entire modalities within instructions; 3) introducing visual perturbations by permuting object order; and 4) increasing the difficulty of the tasks (e.g., placing distractors between source and target). We categorise each perturbation as either \textit{plausible} (e.g., paraphrases) or \textit{unrealistic} (e.g., nonsensical instructions). We expect models to be robust to plausible inputs while dropping performance when faced with unrealistic inputs.

To implement this formalisation, we use \VIMABench which, unlike other state-of-the-art benchmarks such as ALFRED~\citep{shridhar2020alfred}, CLIPort~\citep{shridhar2022cliport}, ARNOLD~\citep{gong2023arnold}, and Ravens~\citep{ZengEtAl2021TransporterNetworksRearranging}, provides several advantages: 1) it covers the majority of robotic manipulation tasks, 2) it offers more fine-grained levels for assessing the systematic generalisation of models; and 3) it represents a benchmark that allows for careful examinations of specific architecture and training regimes. 
For this reason, this paper builds on the controllability of \VIMABench to extensively study the impact that properties of multimodal prompts and visual representations have on model performance.

We applied our novel evaluation setup on multiple state-of-the-art architectures 
commonly used for different Embodied AI tasks and datasets \citep{JiangEtAl2023VIMAGeneralRobot,OctoModelTeam2023OctoOpenSourceGeneralist,OpenX-EmbodimentCollaboration2024OpenXEmbodimentRobotic,shridhar2022cliport, reed2022generalist,Zhao2023LearningFineGrainedBimanual}.\footnote{While feasible, we refrain from applying our evaluation framework on larger Vision and Language Models (VLMs) such as LLaVa \citep{liu2024visual}, as we focus on models of a similar size to VIMA, which are amenable to on-device processing when deployed on real-world robots.}
We uncover several deficiencies of current ``generalist agents'' including 1) insensitivity to language perturbations, as they still perform several tasks when provided with gibberish instructions; and 2) inability to handle tasks of increasing difficulty, potentially including more distractors in the visual scene. 
These findings aim to shed light on state-of-the-art model performance and call for more research on systematically assessing model robustness with adequate tasks and settings that are indicative of the generalisation capacities of Embodied AI models designed to safely and effectively complete tasks in the real world in collaboration with humans.

\section{Related Work}

\paragraph{Language-driven Embodied AI}
Embodied AI focuses on designing agents that are embodied in some environment (simulated or real) and generates actions to complete a given task, whose objective is typically specified in natural language \citep{das2018embodied}.
Tasks for Embodied AI have been formulated in different ways depending on the degree of complexity of the action space. 
For example, Vision+Language Navigation \citep[VLN;][]{Anderson2018VisionandLanguageNavigationInterpreting,Thomason2020VisionandDialogNavigation} requires agents to generate navigation actions to follow natural language instructions and reach some destination in the environment. 
With more sophisticated 3D simulated environments such as AI2Thor \citep{kolve2017ai2}, more recent works also define several tasks involving object interaction \citep[e.g.,][]{shridhar2020alfred,Gao2023AlexaArenaUserCentric,shridhar2022cliport,Stone2023OpenWorldObjectManipulation}.

\paragraph{Language in Robotic Manipulation Tasks}
Language plays a crucial role in many Embodied AI tasks, providing an interface for task learning \citep{laird2017interactive}, with many Embodied AI tasks that require language instructions which are typically hand-crafted via templates (e.g., \VIMABench, CLIPort, ARNOLD) or crowdsourced (e.g., ALFRED).
However, benchmarks often focus on evaluating generalisation using held-out episodes \citep{OpenX-EmbodimentCollaboration2024OpenXEmbodimentRobotic} and do not thoroughly evaluate the importance of language \citep{Stone2023OpenWorldObjectManipulation, OpenX-EmbodimentCollaboration2024OpenXEmbodimentRobotic,OctoModelTeam2023OctoOpenSourceGeneralist}. 
For instance, models trained on \mbox{ALFRED} have shown to be insensitive to language instructions \citep{akula2022alfred}, while nonsensical instructions have even improved downstream performance on the VLN benchmark \citep{zhu2023does}. 

We focus on tabletop robotic manipulation tasks with natural language instructions to measure performance on a well-scoped action execution task. This allows for assessment of visual grounding capabilities from grounding instructions in the real world,  while also removing the extra complexity of sophisticated skills (e.g., SLAM) required for navigation tasks \citep{Anderson2018VisionandLanguageNavigationInterpreting} or the need to predict fine-grained joint control by relying on inverse-kinematics \citep{Ma2024SurveyVisionLanguageActionModels,OpenX-EmbodimentCollaboration2024OpenXEmbodimentRobotic,OctoModelTeam2023OctoOpenSourceGeneralist, ZengEtAl2021TransporterNetworksRearranging}.

\paragraph{Assessing Generalisation and Robustness}

Embodied AI systems must generalise to any complex and novel tasks they might face \citep{duan2022survey}, making robustness a highly-desired characteristic in models, illustrating how well they can ignore spurious correlations and generalise to new domains and tasks \citep{akula2022alfred,gong2023arnold,HupkesEtAl2023TaxonomyReviewGeneralization}.
Embodied AI benchmarks often assess generalisation through seen/unseen scenes \citep[e.g.,][]{shridhar2020alfred,Gao2023AlexaArenaUserCentric,Zheng2022VLMbenchCompositionalBenchmark}, assuming that all tasks the agent must complete and the objects the agent must interact with are fully specified at training time.
While recent benchmarks evaluate models on unseen objects and scenes \citep{gong2023arnold,Stone2023OpenWorldObjectManipulation,OpenX-EmbodimentCollaboration2024OpenXEmbodimentRobotic}, there is no notion of systematic or compositional generalisation to new concepts, affordances \citep{Suglia2020CompGuessWhatMultitaskEvaluation,Pantazopoulos2022CombineDescribeEvaluating}, or novel tasks \citep{chung2022scaling}.

Although models trained on realistic simulations can transfer learned behaviours to real-world environments \citep{OctoModelTeam2023OctoOpenSourceGeneralist,OpenX-EmbodimentCollaboration2024OpenXEmbodimentRobotic}, they remain sensitive to distributional shifts in visual inputs %combinatorial generalisation 
\citep{Li2024EvaluatingRealWorldRobot}, an issue that persists even when training data includes perturbations \citep{Pumacay2024COLOSSEUMBenchmarkEvaluating}.
Furthermore, while models can adapt to relocated targets, they struggle with mid-trajectory linguistic shifts, such as swapping directions from ``left'' to ``right'' \citep{Anwar2024ContrastSetsEvaluating}. 
Our work extends these findings by examining model behaviour under extreme instruction perturbations, providing insights into how models handle challenging and unconventional scenarios.

\section{Experimental Setup}

\paragraph{Evaluation Data}

We use \VIMABench to compare model performance across various skills, tasks, and levels of systematicity, as it is best suited for evaluating the role instructions play in generalising from multimodal prompts.%
\footnote{\citet{JiangEtAl2023VIMAGeneralRobot} did not release a reproducible benchmark; \cref{app:cannot-reproduce} details how we remedied this issue.}
Specifically, we %use \VIMABench to 
assess the compositional generalisation capabilities at four distinct levels of systematicity \citep{Pantazopoulos2022CombineDescribeEvaluating,Hupkes2020CompositionalityDecomposedHow}: object pose sensitivity (L1), combinatorial generalisation (L2), novel objects (L3), and novel tasks (L4). 
See \cref{app:environment-details} for environment and evaluation details.

\paragraph{Models}

We compare four model architectures: 
encoding visual representations with either object-centric or image-patches; and
conditioning prompts on the state through either cross-attention or concatenation 
\citep{Ma2024SurveyVisionLanguageActionModels}.
All models are trained on multimodal instructions with interleaved visual and linguistic features. 
Multimodal instructions are encoded through a frozen pretrained T5 language model \citep{RaffelEtAl2020ExploringLimitsTransfer}, where encoded visual features are injected into the embedding space of the language model \citep{Driess2023PaLMEEmbodiedMultimodal,TsimpoukelliEtAl2021MultimodalFewShotLearning,Ma2024SurveyVisionLanguageActionModels}.
Visual features are implicitly encoded through embedding image frames per observation---more adaptable, more efficient, and % better
outperforming explicit symbolic representations \citep{Song2024SceneDrivenMultimodalKnowledge,Gadre2022ContinuousSceneRepresentations}.
% \vspace{-2ex}
For each observation, the model predicts an action %$a_t$ 
defining a linear movement between two end-effector poses in $\textbf{SE}(3)$---each representing position and rotation in 3D space.
See \cref{app:training} for further training and implementation details.

\section{The Evaluation Framework}

We systematically perturb model inputs at test time to investigate the importance of visual and linguistic information in multimodal prompts. This approach helps us understand how input characteristics contribute to a model's task comprehension. Full per-task results are reported in \cref{app:further-experimental-results}.

\begin{table}[tb]
\centering
\footnotesize
\sisetup{table-format=2.1}
\renewcommand{\arraystretch}{1.25}
\begin{tabularx}{\linewidth}{X SSSS @{}}
\toprule
& {L1} & {L2} & {L3} & {L4} \\
\midrule
\multicolumn{5}{@{}l}{\textit{(a) Trained on Original; Evaluated on Original}} \\
Cross-Attn + Obj-Centric & \bfseries 79.3 & \bfseries 78.8 & \uline{72.3} & \uline{48.6} \\
Cross-Attn + Patches & 63.0 & 62.0 & 44.9 & 13.9 \\
Concatenate + Obj-Centric  & \bfseries 79.2 & \bfseries 78.8 & \bfseries 77.1 & \bfseries 49.2 \\
Concatenate + Patches & 68.0 & 66.3 & 52.9 & 23.4 \\
\midrule[0.2pt] 
\multicolumn{5}{@{}l}{\textit{(b) Trained on Original; Evaluated on Paraphrases}} \\
Cross-Attn + Obj-Centric & \bfseries 78.6 & \bfseries 77.6 & \bfseries 69.8 & \bfseries 47.1\\
Cross-Attn + Patches & 61.1 & 58.5 & 45.3 & 16.8\\
Concatenate + Obj-Centric & \uline{71.5} & \uline{72.2} & \uline{62.7} & \uline{43.0} \\
Concatenate + Patches & 61.3 & 57.0 & 46.0 & 20.5\\
\midrule[0.2pt] 
\multicolumn{5}{@{}l}{\textit{(c) Trained on Paraphrases; Evaluated on Original}} \\
Cross-Attn + Obj-Centric & \bfseries 82.7 &\bfseries 81.8 & \bfseries 77.4 & 48.0 \\
Cross-Attn + Patches & 63.9 & 63.0 & 49.5 & 20.4 \\
Concatenate + Obj-Centric & \uline{80.4} & \uline{78.2} & \uline{74.8} & \bfseries 49.0 \\
Concatenate + Patches & 67.1 & 62.8 & 52.0 & 19.8 \\
\midrule[0.2pt] 
\multicolumn{5}{@{}l}{\textit{(d) Trained on Paraphrases; Evaluated on Paraphrases}} \\
Cross-Attn + Obj-Centric &\bfseries 77.4 & \bfseries 77.5 & \bfseries 70.8 & \bfseries 48.6 \\
Cross-Attn + Patches & 62.2 & 61.0 & 45.7 & 16.1 \\
Concatenate + Obj-Centric & \uline{68.8} & 67.2 & 59.6 & 46.0 \\
Concatenate + Patches & 67.2 &\uline{67.8} & \uline{60.5} & \uline{46.9} \\
\bottomrule
\end{tabularx}
\caption{Average success rate per level for each model when trained or evaluated on either original or paraphrased multimodal instructions.}
\label{tab:perf-paraphrases}
\end{table}

\subsection{Substitutivity in Instructions}

We explore how resilient models with multimodal prompts are to substitutivity \citep{Hupkes2020CompositionalityDecomposedHow} by comparing performance on paraphrased instructions: a meaning-preserving operation. We expect robust models to perform similarly to these \textit{plausible} inputs.\footnote{\cref{app:paraphases} details how paraphrases were created.}
We also replace visual referents with textual descriptors to assess how models %trained on multimodal prompts 
map visuals to the language embedding space.\footnote{We focus on single-object referents, excluding scene or frame referents. See \cref{app:textual-details} for implementation details.}

\paragraph{Baseline}
\hyperref[tab:perf-paraphrases]{\cref*{tab:perf-paraphrases}a} shows model performance for each combination of prompt-conditioning method and visual encoder when trained and evaluated using the default instructions from \VIMABench.
The best-performing approach uses cross-attention to condition prompts on object-centric observations, outperforming image patches. 
Performance for each model is similar at L1--2 but worsens at L3--4, indicating their inability to generalise to new objects and tasks.

\paragraph{Evaluating on Paraphrases}
\hyperref[tab:perf-paraphrases]{\cref*{tab:perf-paraphrases}b} shows that models trained on original instructions are predominantly robust to substitutivity in instructions; however, models do exhibit a small performance loss with cross-attention affected less than those using concatenation. This robustness to paraphrased instructions likely stems from using T5 \citep{RaffelEtAl2020ExploringLimitsTransfer} as the frozen pretrained language model \citep{TsimpoukelliEtAl2021MultimodalFewShotLearning}; \citet{RaffelEtAl2020ExploringLimitsTransfer} demonstrate that T5 exhibits strong performance on GLUE/SuperGLUE \citep{Wang2019GLUEMultiTaskBenchmark,Wang2019SuperGLUEStickierBenchmark}.
The lack of syntactic or lexical diversity in the \VIMABench inputs, suggests that the models might overfit to the surface form rather than learning to generalise to new sentences.

\paragraph{Training on Paraphrases}

\hyperref[tab:perf-paraphrases]{\cref*{tab:perf-paraphrases}c} and \hyperref[tab:perf-paraphrases]{\labelcref*{tab:perf-paraphrases}d} show that training on linguistically-diverse instructions can improve performance for models that use cross-attention to condition prompts or use object-centric visual features. 
However, performance worsens when evaluated on paraphrased instructions.
Taken together, this suggests that training on diverse instructions helps models better connect the semantics of a multimodal instruction over its surface form, aiding in generalisation to novel scenes. However, poor performance on L3 shows that models struggle more with unseen objects. 
Furthermore, using image patches and concatenation performs better when trained and evaluated on diverse instructions over any training/evaluation condition. This suggests that these architectures are more resilient to unseen objects and unseen tasks allowing for better generalisation in more complex settings.

\begin{table}[tb]
\centering
\footnotesize
\sisetup{table-format=2.1}
\renewcommand{\arraystretch}{1.25}
\begin{tabularx}{\linewidth}{X SSSS @{}}
\toprule
& {L1} & {L2} & {L3} & {L4} \\
\midrule
\multicolumn{5}{@{}l}{\textit{With Visual Referents*}} \\
Cross-Attn + Obj-Centric & \uline{87.3} & \bfseries 86.4 & \uline{75.8} & \bfseries 49.2 \\
Cross-Attn + Patches & 78.3 & 77.5 & 54.6 & 17.8 \\
Concatenate + Obj-Centric  & \bfseries 88.9 & \bfseries 86.4 & \bfseries 81.2 & 48.5 \\
Concatenate + Patches & 79.9 & 75.1 & 55.0 & 15.2 \\
\midrule[0.2pt]
\multicolumn{5}{@{}l}{\textit{Replace Visual Referents with Descriptors*}} \\
Cross-Attn + Obj-Centric & \bfseries 87.9 & \bfseries 87.2 & \bfseries 73.3 & \bfseries 49.0 \\
Cross-Attn + Patches & 46.8 & 44.7 & 38.2 & 25.5 \\
Concatenate + Obj-Centric  & \uline{79.4} & \uline{78.1} & \uline{70.0} & \uline{38.5} \\
Concatenate + Patches & 56.4 & 50.6 & 52.0 & 25.8 \\
\bottomrule
\end{tabularx}
\caption{Average success rate per level when visual referents are replaced with textual descriptors during evaluation only. Models trained on paraphrased multimodal instructions---using visual referents. \textit{*\,Not all tasks included; see \cref{app:textual-details} for details.}}
\label{tab:perf-obj-text-para}
\end{table}

\paragraph{Replacing Visual Referents with Descriptors}

\cref{tab:perf-obj-text-para} shows object-centric models perform comparably when replacing objects with natural language descriptors, suggesting that models have learned to map visual features within the language model's embedding space \citep{TsimpoukelliEtAl2021MultimodalFewShotLearning,Driess2023PaLMEEmbodiedMultimodal}.
Furthermore, cross-attention outperforms concatenation, indicating it better preserves relationships between natural language descriptors and visual referents.
Additionally, both models that use image patches perform notably worse on L1--3. 
When using patches, all visuals provided to the model in the prompt---be it a single object or a frame---are encoded into a fixed number of patches, whereas object-centric methods encode one object per token. Due to the cardinality of this mapping, the former is a more difficult task than the latter.

\subsection{Perturbations of Instruction Syntax}

\begin{figure}[tb]
    \centering
    \includegraphics[width=\linewidth]{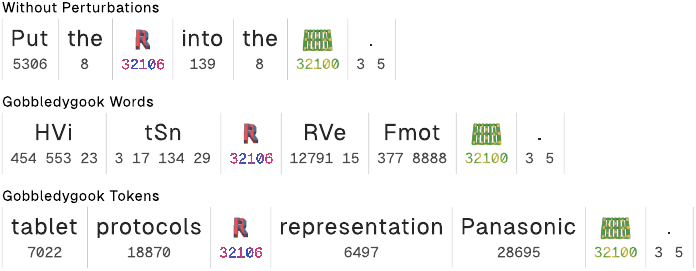}
    \caption{Illustration of language perturbations \textbf{challenging model sensitivity to language content in multimodal instructions}: \GDGWords (random characters, increased token length) and \GDGTokens (random words, same sequence length).}
    \label{fig:lang-perturbation-example}
\end{figure}

We introduce two methods to distort language within a multimodal prompt: \GDGWords and \GDGTokens. 
As shown in \cref{fig:lang-perturbation-example}, each method removes information from the language modality differently without affecting the visual referents.
\GDGTokens preserves the tokenised sequence length, while \GDGWords maintains the word count but increases the tokenised sequence length (see \cref{app:gdg} for implementation details).
As \GDG perturbations are unrealistic, we expect performance to plummet to near-random chance.\footnote{We derive random chance in \cref{app:random-chance}.} 
Furthermore, while the \GDG perturbations removes signal from the linguistic channel, it does not remove text from the instruction. While irrelevant to the task, they are still provided to, and considered by, the language model. 
To investigate the contribution of each modality further, we compare their individual impact on the overall model performance.

\begin{table}[tb]
\centering
\footnotesize
\sisetup{table-format=2.1}
\renewcommand{\arraystretch}{1.25}
\begin{tabularx}{\linewidth}{X SSSS @{}}
\toprule
& {L1} & {L2} & {L3} & {L4} \\
\midrule
\multicolumn{5}{@{}l}{\textit{Without Gobbledygook*}} \\
Cross-Attn + Obj-Centric & \bfseries 82.7 &\bfseries 81.8 & \bfseries 77.4 & 48.0 \\
Cross-Attn + Patches & 63.9 & 63.0 & 49.5 & 20.4 \\
Concatenate + Obj-Centric & \uline{80.4} & \uline{78.2} & \uline{74.8} & \bfseries 49.0 \\
Concatenate + Patches & 67.1 & 62.8 & 52.0 & 19.8 \\
\midrule[0.2pt] 
\multicolumn{5}{@{}l}{\GDGTokens} \\
Cross-Attn + Obj-Centric & \bfseries 56.7 & \uline{54.5} & \uline{36.6} & \uline{22.9} \\
Cross-Attn + Patches & 45.2 & 45.9 & 34.0 & 15.1 \\
Concatenate + Obj-Centric  & \bfseries 56.7 & \bfseries 55.3 & \bfseries 45.8 & \bfseries 26.4 \\
Concatenate + Patches & 45.9 & 44.3 & 32.9 & 20.0 \\
\midrule[0.2pt] 
\multicolumn{5}{@{}l}{\GDGWords} \\
Cross-Attn + Obj-Centric &  \bfseries 50.8 & \bfseries 51.8 & \bfseries 39.9 & \bfseries 33.8 \\
Cross-Attn + Patches &  \uline{46.7} & \uline{48.4} & 33.9 & 18.6 \\
Concatenate + Obj-Centric  &  44.8 & 44.5 & \uline{35.4} & \uline{23.9} \\
Concatenate + Patches & 44.3 & 42.7 & 31.0 & 19.0 \\
\bottomrule
\end{tabularx}
\caption{Average success per level after applying each \GDG perturbation to the original multimodal instructions, showing all models outperforming random chance. Models trained on multimodal paraphrased instructions. \textit{*\,Copied from \crefsubtable{tab:perf-paraphrases}{c}.}}
\label{tab:perf-gdg-on-para}
\end{table}

\paragraph{\GDG Perturbations}
\cref{tab:perf-gdg-on-para} shows that \GDG perturbations degrade performance across architectures, but not to random chance, implying that models rely on other cues to infer tasks despite nonsensical instructions.  

When applying \GDGTokens, object-centric features outperform image-patches, regardless of the prompt-conditioning method used. This implies that object-centric features provide a stronger signal for models to infer the desired task without explicit direction. 
While we would expect similar performance drops across both perturbation methods, object-centric models exhibit poorer performance with \GDGWords compared to \GDGTokens.
With \GDGWords, conditioning with cross-attention helps models uncover the task at lower levels, but cross-attention with patches struggles more with novel tasks and objects, possibly indicating overfitting. 
This problem might arise because the decoder uses absolute positional embeddings, which are known to poorly extrapolate to longer sequences \citep{Press2022TrainShortTest,Sun2022LengthExtrapolatableTransformer}.

\begin{table}[tb]
\centering
\footnotesize
\sisetup{table-format=2.1}
\renewcommand{\arraystretch}{1.25}
\begin{tabularx}{\linewidth}{X SSSS @{}}
\toprule
& {L1} & {L2} & {L3} & {L4} \\
\midrule
\multicolumn{5}{@{}l}{\textit{No Tokens Masked*}} \\
Cross-Attn + Obj-Centric & \bfseries 82.7 &\bfseries 81.8 & \bfseries 77.4 & 48.0 \\
Cross-Attn + Patches & 63.9 & 63.0 & 49.5 & 20.4 \\
Concatenate + Obj-Centric & \uline{80.4} & \uline{78.2} & \uline{74.8} & \bfseries 49.0 \\
Concatenate + Patches & 67.1 & 62.8 & 52.0 & 19.8 \\
\midrule[0.2pt] 
\multicolumn{5}{@{}l}{\textit{Mask Language Tokens}} \\
Cross-Attn + Obj-Centric & \uline{36.3} & \uline{35.1} & 19.1 & 14.8 \\
Cross-Attn + Patches & 26.3 & 26.5 & 20.7 & 11.2 \\
Concatenate + Obj-Centric & \bfseries 39.0 & \bfseries 39.0 & \bfseries 28.8 & \bfseries 25.9 \\
Concatenate + Patches & 30.2 & 29.0 & \uline{24.5} & 16.2 \\
\midrule[0.2pt] 
\multicolumn{5}{@{}l}{\textit{Mask Visual Referents}} \\
Cross-Attn + Obj-Centric & 63.6 & 62.6 & \bfseries 56.4 & \bfseries 47.9 \\
Cross-Attn + Patches & \uline{64.8} & \uline{63.0} & 49.6 & 20.6 \\
Concatenate + Obj-Centric & 59.8 & 58.9 & 53.2 & \bfseries 47.8 \\
Concatenate + Patches & \bfseries 67.1 & \bfseries 63.7 & \uline{52.8} & 23.0 \\
\bottomrule
\end{tabularx}
\caption{Average success rate per level after \textbf{masking tokens from one modality} within the multimodal instruction. Models trained on paraphrased instructions and evaluated on original instructions. \textit{*\,Copied from \crefsubtable{tab:perf-paraphrases}{c}.}}
\label{tab:perf-mask-modalities-on-para}
\end{table}

\paragraph{Comparing Modalities}

We investigate whether models rely equally on both modalities by masking one modalities at test time---a perturbation that should significantly decrease performance. 
\cref{tab:perf-mask-modalities-on-para} shows that when masking one modality, performance across all models and levels is above random chance, indicating that models continue to determine the task. 
Notably, performance suffers more when masking language tokens than when the visual referents are masked. While this indicates that models may rely more heavily on the language content, we would expect that applying \GDG perturbations to lead to a comparable drop in performance as masking out the language tokens entirely. Since this is not the case, we instead hypothesise that we can attribute the observed differences to the nature of autoregressive modelling and the order in which modalities are arranged in the instructions. Specifically, as all instructions begin with language tokens, models may struggle with sequences that do not start this way.

\subsection{Are Models Relying on Heuristics?}

When provided with incomplete instructions, humans often combine available information with heuristics to act rationally in the face of uncertainty \citep{Simon1955BehavioralModelRational,Gigerenzer1996ReasoningFastFrugal}. 
Similarly, models may rely on heuristics---combining \textit{any} available information with prior knowledge and world understanding---to infer appropriate actions and complete tasks. Furthermore, when given the opportunity, models may attempt to recover from mistakes through trial and error.

\begin{table}[tb]
\centering
\footnotesize
\sisetup{table-format=2.1}
\renewcommand{\arraystretch}{1.25}
\begin{tabularx}{\linewidth}{X SSSS @{}}
\toprule
& {L1} & {L2} & {L3} & {L4} \\
\midrule
\multicolumn{5}{@{}l}{\textit{(a) Mistakes Allowed*}} \\
Cross-Attn + Obj-Centric & \bfseries 82.7 &\bfseries 81.8 & \bfseries 77.4 & 48.0 \\
Cross-Attn + Patches & 63.9 & 63.0 & 49.5 & 20.4 \\
Concatenate + Obj-Centric & \uline{80.4} & \uline{78.2} & \uline{74.8} & \bfseries 49.0 \\
Concatenate + Patches & 67.1 & 62.8 & 52.0 & 19.8 \\
\midrule[0.2pt] 
\multicolumn{5}{@{}l}{\textit{(b) No Mistakes Allowed}} \\
Cross-Attn + Obj-Centric & \uline{70.3} & \uline{69.7} & \bfseries 67.9 & \bfseries 46.8 \\
Cross-Attn + Patches & 58.2 & 57.3 & 44.2 & 15.9 \\
Concatenate + Obj-Centric  & \bfseries 72.2 & \bfseries 71.4 & \uline{65.7} & \uline{45.5} \\
Concatenate + Patches & 61.2 & 57.6 & 46.0 & 12.9 \\
\bottomrule
\end{tabularx}
\caption{Average success rate per level when models must solve tasks either \textbf{with or without mistakes permitted}. Models trained on paraphrased instructions and evaluated on original instructions. \textit{*\,Copied from \crefsubtable{tab:perf-paraphrases}{c}.}}
\label{tab:perf-no-mistakes-on-para}
\end{table}

\paragraph{Models Try to Recover from Mistakes}

\crefsubtable{tab:perf-no-mistakes-on-para}{b} shows object-centric representations outperform models encoding visuals with image-patches, showing that these models are better at solving the tasks without any errors. 
However, the performance across all levels and models is lower compared to the more lenient time limit (\crefsubtable{tab:perf-no-mistakes-on-para}{a}), indicating that given additional time, models explore alternative actions, often successfully. 
While beneficial, extended time may lead to misleading conclusions when evaluating model performance under unreasonable conditions or nonsensical instructions, as models simply have more time to perform suboptimal action sequences that eventually lead to success.
We attribute this behaviour to recovery demonstrations from \VIMABench (see discussion in \cref{app:data-not-optimal}).

\begin{table}[tb]
\centering
\footnotesize
\sisetup{table-format=2.1}
\renewcommand{\arraystretch}{1.25}
\begin{tabularx}{\linewidth}{X SSSS @{}}
\toprule
& {L1} & {L2} & {L3} & {L4} \\
\midrule
\multicolumn{5}{@{}l}{\textit{Mask Instructions; Mistakes Allowed}} \\
Cross-Attn + Obj-Centric &  \bfseries 52.4 & \bfseries 50.5 & \bfseries 40.6 & \bfseries 33.0 \\
Cross-Attn + Patches & 26.5 & 25.8 & 20.2 & 11.9 \\
Concatenate + Obj-Centric & \uline{30.9} & \uline{30.5} & 22.1 & 14.1 \\
Concatenate + Patches & \uline{30.4} & 29.4 & \uline{25.0} & \uline{15.6} \\
\midrule[0.2pt] 
\multicolumn{5}{@{}l}{\textit{Mask Instructions; No Mistakes Allowed}} \\
Cross-Attn + Obj-Centric & \bfseries 45.2 & \bfseries 43.9 & \bfseries 33.1 & \bfseries 27.4 \\
Cross-Attn + Patches & 19.5 & 18.7 & 14.5 & 7.6 \\
Concatenate + Obj-Centric & 7.0 & 7.2 & 3.5 & 2.2 \\
Concatenate + Patches & \uline{22.5} & \uline{22.2} & \uline{18.2} & \phantom{0}\uline{8.8} \\
\bottomrule
\end{tabularx}
\caption{Average success rate per level with \textbf{instructions entirely masked} at test-time. Models trained on paraphrased instructions and evaluated with original instructions before masking. Performing above random chance indicates that the model is using other information solve each task.}
\label{tab:perf-no-instructions-trained-para}
\end{table}

\paragraph{Models Act Without Instructions}\label{sec:no-prompt}

\cref{tab:perf-no-instructions-trained-para} reveals that models continue to perform tasks when instructions are entirely removed, which suggests that models learn to rely on heuristics from observations.
Concatenation with object-centric visual features exhibits worse performance, indicating a higher sensitivity to the presence of an instruction, which is a desirable characteristic.
Additionally, cross-attention with object-centric features outperforms concatenation, despite both models using the same visual encoding method. 
Model performance across all levels is greater when they can recover from errors, suggesting that models will persistently attempt to solve the task if uninterrupted.
This behaviour raises important safety concerns: models are acting without clear direction, and yet somehow find the right answer. 
Furthermore, this difference highlights the effects of how instructions are conditioned on observations, especially when those instructions are masked.

\begin{figure*}[tb]
    \centering
    \includegraphics[width=\textwidth]{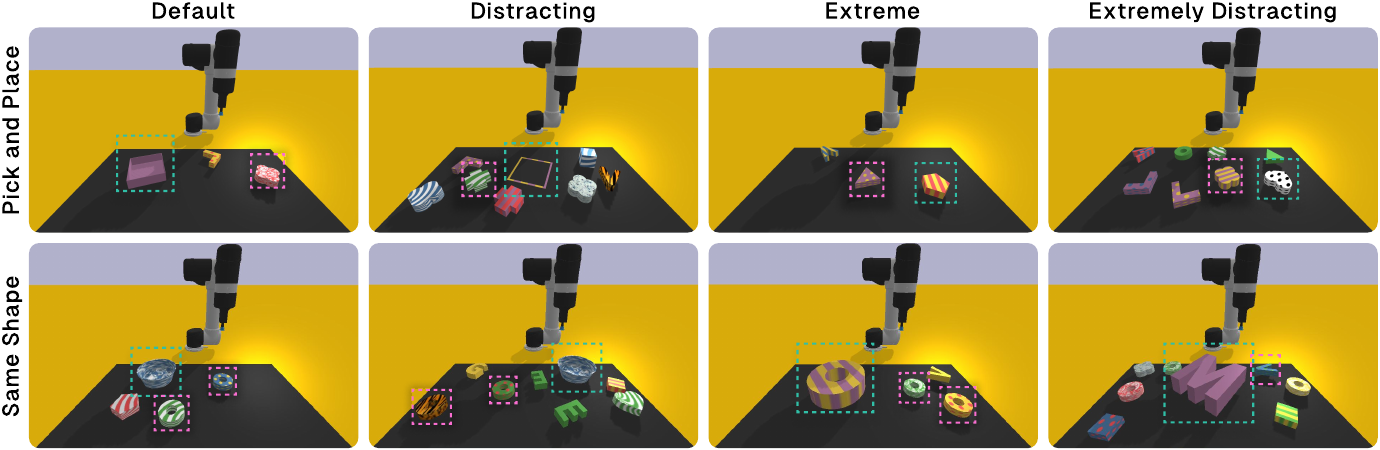}
    \caption{Difficulty level comparisons to default (first column). \textit{Distracting} add visual clutter; \textit{Extreme} changes parameters, complexity, and affordances; and, \textit{Extremely Distracting} combines both. Top row: T1 (\textit{``pick and place into the container''}). Bottom row: T15 (\textit{``place all objects with the same shape as the container into it''}). For illustration purposes, we denote target containers with a green dashed box and target objects with pink dashed box.}
    \label{fig:difficulty-example}
\end{figure*}

\subsection{Task Complexity}

As each architecture can infer the correct task without instruction, it implies that they rely on cues solely from the observations, as that is the only other source of input into the model. 
We test this in two ways: 1) by introducing distractors with the \textit{Distracting} difficulty level, or 2) by increasing task difficulty with the \textit{Extreme} difficulty level. 
Distractors are objects similar to the target objects in either texture or shape and ``task difficulty increases'' are specific to each task. %(outlined in \cref{app:difficulty-levels}).
The \textit{Extreme} level assesses a model reliance on object affordances when reasoning about actions \cite{Lohmann2020LearningObjectsLearning}. 
These new difficulty levels are \textit{plausible}: agents should be able to disregard unnecessary details and focus on task-critical objects or aspects. 
\cref{fig:difficulty-example} provides an example, with further details in \cref{app:difficulty-levels}.

\begin{table}[tb]
\centering
\footnotesize
\sisetup{table-format=2.1}
\renewcommand{\arraystretch}{1.25}
\begin{tabularx}{\linewidth}{X SSSS @{}}
\toprule
& {L1} & {L2} & {L3} & {L4} \\
\midrule
\multicolumn{5}{@{}l}{\textit{(a) Distracting}} \\
Cross-Attn + Obj-Centric & \Uline{53.8} & \Uline{52.4} & \Uline{46.6} & \Uline{34.8} \\
Cross-Attn + Patches & 27.9 & 27.4 & 18.2 & 3.8 \\
Concatenate + Obj-Centric  & \bfseries 60.2 & \bfseries 59.8 & \bfseries 53.3 & \bfseries 39.0 \\
Concatenate + Patches & 29.2 & 27.0 & 18.4 & 5.1 \\
% \midrule[0.2pt] 
\multicolumn{5}{@{}l}{\textit{(b) Extreme}} \\
Cross-Attn + Obj-Centric &  \bfseries 53.1 &\bfseries 53.5 & \bfseries55.5 & \bfseries36.6 \\
Cross-Attn + Patches & 13.0 & 12.5 & 9.7 & 9.2 \\
Concatenate + Obj-Centric  & \Uline{22.5} & \Uline{23.0} & \Uline{23.2} & \Uline{10.5} \\
Concatenate + Patches & 16.2 & 15.0 & 12.1 & 12.1 \\
% \midrule[0.2pt] 
\multicolumn{5}{@{}l}{\textit{(c) Extremely Distracting}} \\
Cross-Attn + Obj-Centric & \bfseries30.2 & \bfseries30.7 & \bfseries33.0 & \bfseries31.8 \\
Cross-Attn + Patches & 4.5 & 3.8 & 2.1 & 2.9 \\
Concatenate + Obj-Centric  & \Uline{14.6} & \Uline{14.3} & \Uline{10.8} & \Uline{8.5} \\
Concatenate + Patches &5.7 & 5.2 & 2.8 & 4.1 \\
\midrule[0.2pt] 
\multicolumn{5}{@{}l}{\textit{(d) Distracting; Mask Instructions}} \\
Cross-Attn + Obj-Centric & \bfseries33.0 & \bfseries32.2 & \bfseries21.0 & \bfseries21.9 \\
Cross-Attn + Patches & \Uline{6.7} & \Uline{7.3} & 2.4 & \Uline{2.6} \\
Concatenate + Obj-Centric  & 5.9 & 4.5 & 1.2 & 0.6 \\
Concatenate + Patches &  \Uline{6.5} & 6.7 & \Uline{3.4} & 1.6 \\
% \midrule[0.2pt] 
\multicolumn{5}{@{}l}{\textit{(e) Extreme; Mask Instructions}} \\
Cross-Attn + Obj-Centric &  \bfseries15.2 & \bfseries15.5 & \bfseries17.4 & \bfseries11.2 \\
Cross-Attn + Patches & \Uline{5.7} & \Uline{4.6} & \Uline{2.7} & 4.6 \\
Concatenate + Obj-Centric  & 4.5 & 3.3 & 2.0 & 2.2 \\
Concatenate + Patches & 3.9 & 4.1 & \Uline{2.7} & \Uline{7.4} \\
% \midrule[0.2pt] 
\multicolumn{5}{@{}l}{\textit{(f) Extremely Distracting; Mask Instructions}} \\
Cross-Attn + Obj-Centric & \bfseries10.1 &\bfseries 8.7 &\bfseries 8.7 &\bfseries 10.9 \\
Cross-Attn + Patches & 2.9 & 2.8 & 0.4 & \Uline{3.6} \\
Concatenate + Obj-Centric  & \Uline{4.3} & \Uline{4.3} & \Uline{2.0} & 1.4 \\
Concatenate + Patches & 1.7 & 1.6 & 0.4 & 1.2 \\
\bottomrule
\end{tabularx}
\caption{
Average success rates \textbf{across difficulty levels}. Models trained on paraphrased instructions and evaluated with original instructions \textit{without any mistakes}.}
\label{tab:perf-difficulty-on-para}
\end{table}

\begin{figure}[tb]
    \centering
    \includegraphics[width=1\linewidth]{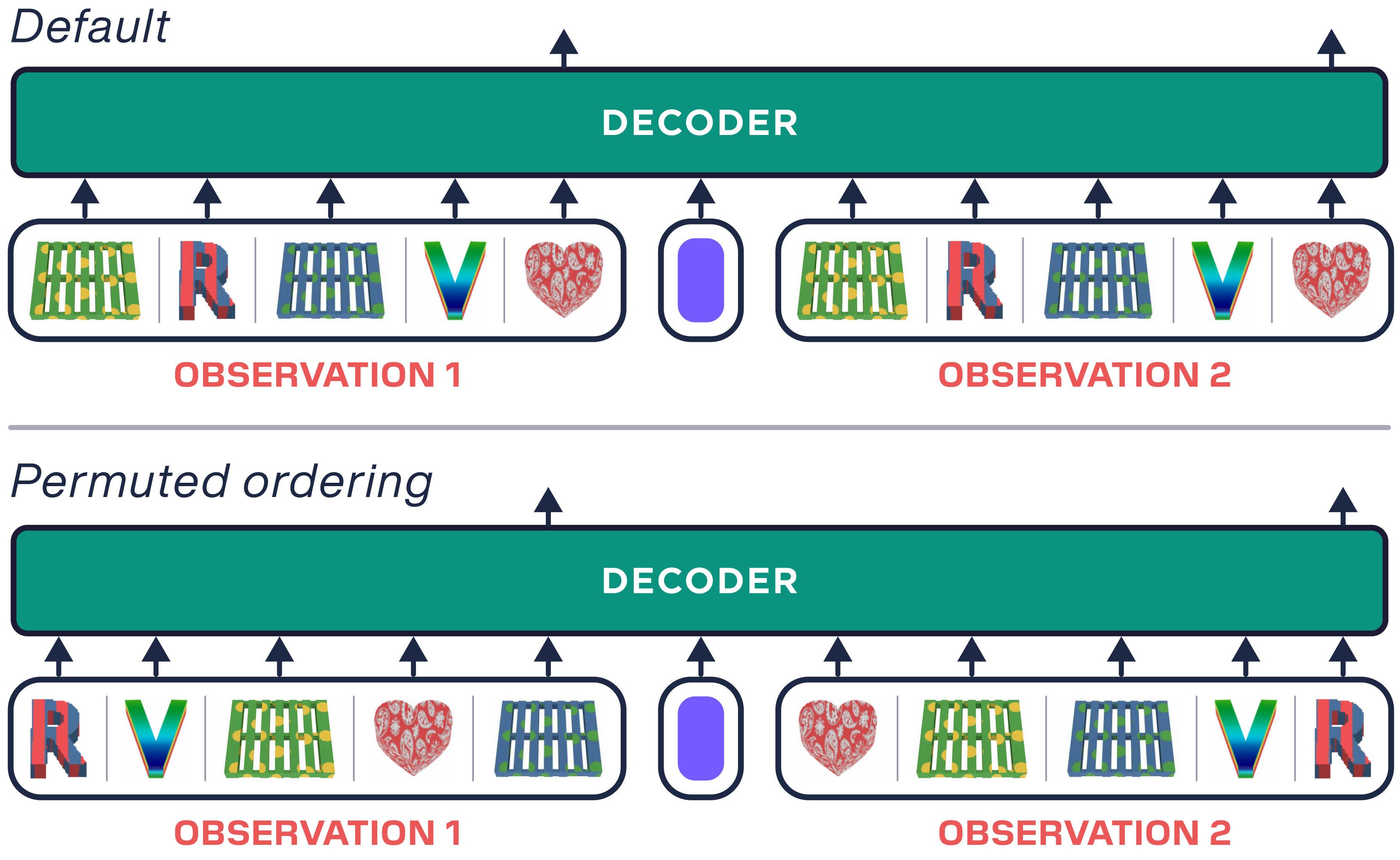}
    \caption{Illustration comparing default and permuted object tokens per observation. In the default ordering (top), tokens in each observation follow the same pattern: the container object first, the target object second, and then any distractor objects. The permuted ordering (bottom) randomises the order differently for each observation in the same sequence.
    }
    \label{fig:shuffle-example}
\end{figure}

\cref{tab:perf-difficulty-on-para} presents results on our novel evaluation set. 
Models using patches likely perform poorly due to their inability to represent objects in complex scenes, a known limitation of Transformer-based vision encoders \citep{darcet2023vision, pantazopoulos2024lost}. Recent work has proposed several solutions to favour suitable object-centric representation learning \citep{locatello2020object}. 
While increasing the resolution per patch or image %is one possible solution that has been shown to 
might improve performance  \citep{Karamcheti2024PrismaticVLMsInvestigating,Liu2024ImprovedBaselinesVisual}, it can increase the number of tokens in the decoder, potentially introducing new issues such as increased computational complexity \citep{Lin2022SurveyTransformers} or inference time \citep{Firoozi2024FoundationModelsRobotics}.
The \textit{Extreme} difficulty level, which changes expected affordances of objects (e.g., using non-container objects to place objects on) impacts patch-based models more significantly than object-centric models. 
This indicates that patch-based models are less robust when objects are used in unexpected ways, while object-centric models adapt better to these changes.
At the \textit{Extremely Distracting} difficulty level, patch-based models struggle substantially, indicating their inability to handle both altered object affordances and excessive visual clutter. This decline highlights limitations of patch-based models in complex, yet plausible, scenarios.

As task complexity increases, we expect the model to be increasingly reliant on an instruction to be able to solve the task without error. \hyperref[tab:perf-difficulty-on-para]{Tables \labelcref*{tab:perf-difficulty-on-para}d--f} show that with masked instructions, all models except \textit{Cross-Attn + Obj-Centric} plummet, though not to random chance. This suggests that instructions are crucial in more complex settings, there remains some chance that the model may successfully solve the task.\footnote{See \cref{app:per-task-difficulty-masked} for additional analysis into why average performance is above random chance.}
As  \textit{Cross-Attn + Obj-Centric} outperforms all other models without instruction, it suggests that it is using heuristics from the environment to determine and solve the task.

\begin{table}[tb]
\centering
\footnotesize
\sisetup{table-format=2.1}
\renewcommand{\arraystretch}{1.3}
\begin{tabularx}{\linewidth}{X SSSS @{}}
\toprule
& {L1} & {L2} & {L3} & {L4} \\
\midrule
\multicolumn{5}{@{}l}{\textit{Permute Object Tokens; Mistakes Allowed}} \\
Cross-Attn + Obj-Centric & \bfseries 40.9 & 39.1 & 33.3 & 11.8 \\
Concatenate + Obj-Centric & \bfseries 40.6 & \bfseries 40.6 & \bfseries 36.3 & \bfseries 14.5 \\
% \midrule[0.2pt] 
\multicolumn{5}{@{}l}{\textit{Permute Object Tokens; No Mistakes Allowed}} \\
Cross-Attn + Obj-Centric &24.9 & 24.6 & 20.7 & 5.9 \\ 
Concatenate + Obj-Centric & \bfseries 27.6 & \bfseries 27.8 & \bfseries 24.8 & \bfseries 8.2 \\
\midrule[0.2pt] 
\multicolumn{5}{@{}l}{\textit{Permute Object Tokens; Distracting}} \\
Cross-Attn + Obj-Centric & \bfseries 14.5 & \bfseries 14.3 & \bfseries 12.0 & \bfseries 1.2 \\
Concatenate + Obj-Centric & 13.3 & 12.5 & \bfseries 12.0 & \bfseries 1.4 \\
% \midrule[0.2pt] 
\multicolumn{5}{@{}l}{\textit{Permute Object Tokens; Extreme}} \\
Cross-Attn + Obj-Centric & \bfseries 12.0 & \bfseries 12.7 & \bfseries 10.8 & 6.1 \\
Concatenate + Obj-Centric & 7.7 & 7.3 & 7.2 & \bfseries 7.1 \\
\midrule[0.2pt] 
\multicolumn{5}{@{}l@{}}{\textit{Train + Eval with Permutation; Mistakes Allowed}} \\
Cross-Attn + Obj-Centric & 59.7 & 42.1 & 38.1 & \bfseries 14.4 \\ 
Concatenate + Obj-Centric &  \bfseries 70.6 & \bfseries 49.9 & \bfseries 44.7 & \bfseries 14.5 \\
% \midrule[0.2pt] 
\multicolumn{5}{@{}l}{\textit{Train + Eval with Permutation; No Mistakes Allowed}} \\
Cross-Attn + Obj-Centric & 50.3 & 34.1 & 30.1 & \bfseries 10.0 \\
Concatenate + Obj-Centric &  \bfseries 58.4 & \bfseries 41.0 & \bfseries 34.5 & 8.1 \\
% \midrule[0.2pt] 
\bottomrule
\end{tabularx}
\caption{Average success rate per level when \textbf{evaluated with permuted object tokens.} All models are trained with paraphrased instructions and evaluated with original instructions.}
\label{tab:perf-eval-shuffle-on-para}
\end{table}

\subsection{Order Permutations}

Object-centric models outperform others, but how they succeed without instruction remains unclear, possibly due to cues from observation encoding.
We explore whether permuting the order of object tokens when provided in the model's input affects model performance 
(see \cref{fig:shuffle-example} for example permutations). 
We assume that Transformer-based models using object-centric tokens should be invariant to order permutations \citep{Carion2020EndtoEndObjectDetection}.

Instead, as shown in \cref{tab:perf-eval-shuffle-on-para}, we note that permuting the order of object-centric tokens causes performance on the default difficulty level to half. 
Exploring how well models perform without the opportunity to recover from mistakes halves this result further. This indicates that when they do not rely on spurious correlations, models try to recover from mistakes until an episode terminates.
Further proof of this is that performance degrades as the environment becomes more complex: both with more objects present (\textit{Distracting}) and when various affordances are not as expected (\textit{Extreme}).

Similar to findings from \citet{Carion2020EndtoEndObjectDetection}, Transformer-based models are vulnerable to order permutations. When trained on these permutations, model performance improves, however, it is not at the same level as \cref{tab:perf-paraphrases}, suggesting that a considerable proportion of model performance stems from learned spurious correlations.

% --------
\section{Conclusion}

We define an evaluation framework for Embodied AI grounded in the generalisation framework from \cite{HupkesEtAl2023TaxonomyReviewGeneralization}. Specifically, we assess generalisation across important axes by means of specific multimodal input perturbations including paraphrases, replacing visual referents with descriptors, and manipulating the instruction syntax as well as entire input modalities. We instantiate this evaluation framework in \VIMABench to assess the robustness of state-of-the-art models.%Embodied AI architectures. 

Overall, our findings indicate that while substitutivity can lead to performance gains, language perturbations do not impact performance as expected. To further explore this effect, we evaluate whether models rely on heuristics to complete tasks by removing individual modalities. We show that models perform tasks even without instructions by relying on spurious correlations within observations, as learned during training. We further prove this effect by showing that performance decreases when the number of objects in an environment increases, and agents can no longer randomly perform the correct sequence of actions.

Taken together, our findings suggest that it is important to define evaluation frameworks like ours that can assess generalisation across multiple axes in order to have a more reliable characterisation of the overall model performance. 
In future work, we aim to apply this evaluation framework systematically to other benchmarks as well to discover important architectural insights that will guide the next generation of Embodied AI models.

% --------
% \clearpage

\section*{Limitations \& Ethical Considerations}

\paragraph{Limited in Embodied AI}

This study aims to provide Embodied AI researchers with an experimental evaluation framework for studying generalisation capabilities of robot policies via an extensive set of multimodal input perturbations. We have instantiated this framework using \VIMABench. \VIMABench was created to evaluate robot manipulation tasks in a controlled setting with a focus on compositional generalisation skills. To date, many proposed embodied AI tasks require several skills, such as navigation and manipulation. We focus on manipulation skills as they remove an extra degree of complexity found in navigation tasks that require more sophisticated skills (e.g., SLAM). Further, tabletop manipulation allows us to focus on problems in grounding language instructions in the real world to assess visual grounding capabilities. 

The architectures used in this work are also used in more realistic benchmarks \citep[e.g.,][]{OpenX-EmbodimentCollaboration2024OpenXEmbodimentRobotic}. Therefore, this provides the possibility to study architectures used for embodied AI tasks under very strict conditions without being influenced by differences in robotic platforms and embodiments.

The main contribution of our paper is to assess to what extent this is true and to shed light on the weaknesses of current Transformer-based action policies. 
Additionally, we believe that our framework is generic enough to be applied to other datasets considering that it analyses model performance using core concepts of systematic generalisation \cite{HupkesEtAl2023TaxonomyReviewGeneralization}.

\paragraph{Choice of Perturbations on Visual Observations}

In this work, we focus primarily on perturbations that directly affect how models make decisions. However, a possible avenue for future work would be to explore how robust models are to other factors such as camera choice and background colours \citep{Pumacay2024COLOSSEUMBenchmarkEvaluating}. 
In robotic manipulation tasks, the camera's distance from the robot is often constant \citep{ZengEtAl2021TransporterNetworksRearranging,shridhar2022cliport,OctoModelTeam2023OctoOpenSourceGeneralist}. Changing the camera's position relative to the robot after training would introduce confounds and increase downstream difficulties, unless trained to do so \citep[e.g.,][]{Grauman2024EgoExo4DUnderstandingSkilled, Pumacay2024COLOSSEUMBenchmarkEvaluating}.
When deploying models, it is crucial to test them under varying light levels and background colours. Reducing light levels can impede the model’s ability to perceive objects. Therefore, using ground-truth segmentation masks in low-light conditions is ecologically invalid; requiring a new model to extract segmentation masks at risk of introducing new confounds and potential issues like sensitivity to light or camera limitations.

\paragraph{Safety Concerns with Embodied AI}

The aim of Embodied AI is to build artificial agents that can collaborate and enhance the human experience via either offering companionship \citep{strohmann2023toward, deng2019embodiment} or performing tasks \citep{takeda2019accountable, duan2022survey}. As explained by \citet{duan2022survey}, the latter is tested via simulations which attempt to create ecologically valid frameworks to evaluate agent performance before deployment in a real-world setting. Through this lens, the findings shown in this paper are particularly worrisome, as the shortcomings that we describe indicate issues with the evaluation process itself. This could mean that embodied agents previously evaluated as successful in their generalisation capabilities may fail outside of a simulated environment, increasing the chance to harm humans.

While our framework explains how to thoroughly and systematically assess the training and evaluation of an embodied agent, it is important to note that while our exploration is extensive, there are still aspects that fall outside of the scope of this paper. Our future work aims to apply our framework to a wider array of environments. This will allow us to provide the research community with a more systematic evaluation approach aimed at pinpointing edge cases and limitations of Embodied AI systems, paving the way to a more robust solution for Sim2Real transfer.

% --------

\section*{Acknowledgements}

This work was supported by the Edinburgh International Data Facility (EIDF) and the Data-Driven Innovation Programme at the University of Edinburgh. We are also grateful for the Heriot-Watt University high-performance computing facility (DMOG) and associated support services.
We extend our gratitude to the members of the Interaction Lab at Heriot-Watt University for their valuable feedback. 
In particular, we would like to express our sincere thanks to Sabrina McCallum, Malvina Nikandrou, and Georgios Pantazopoulos, whose insightful feedback on earlier versions was instrumental in improving the quality and clarity of our work. 
Finally, we appreciate the constructive feedback from all anonymous reviewers, which helped us improve this paper.

% --------
% \clearpage
\bibliography{bib}

\renewcommand{\thetable}{\Alph{section}.\arabic{table}}
\renewcommand{\thefigure}{\Alph{section}.\arabic{figure}}

\clearpage
\appendix

\setcounter{table}{0}
\setcounter{figure}{0}
\section{Training Details}\label{app:training}

\begin{table}[tb]
\centering
\footnotesize
\renewcommand{\arraystretch}{1.3}
\begin{tabularx}{\linewidth}{@{}XX@{}}
    \toprule
    Hyperparameter & Value \\
    \midrule
    Pretrained Language Model & \texttt{t5-base} \citep{RaffelEtAl2020ExploringLimitsTransfer} \\
    Optimizer & AdamW \citep{Loshchilov2019DecoupledWeightDecay} \\
    Dropout & 0.1 \\
    Weight Decay & 0 \\
    Gradient Clip Threshold & 1.0 \\ 
    Maximum Learning Rate & 1e-4 \\
    Minimum Learning Rate & 1e-7 \\ 
    Warmup steps & 7K (896K examples) \\
    Cosine Annealing steps & All remaining steps \\
    Training epochs & 10 \\
    Total examples seen & \num{6099200} \\
    Examples per optimizer step & 128 \\ 
    \bottomrule
\end{tabularx}
\caption{Hyperparameters using during model training for each model.}
\label{tab:hyperparams}
\end{table}

\subsection{Policy Definition}

In the environment, models must learn a non-Markovian policy $\pi : \mathcal{P} \times \mathcal{H} \rightarrow \mathcal{A}$, which is essential for completing tasks that rely on previous observations (such as tasks 5 and 16).
The policy $\pi$ maps a multimodal instruction $\vp \in \mathcal{P}$ and a history trajectory of observations and actions $h_t \in \mathcal{H}$ up to some discrete time step $t$ to the two-pose action primitive $a_t = (\mathcal{T}_{\mathrm{start}},\mathcal{T}_{\mathrm{end}}) \in \mathcal{A}$.

A multimodal instruction $\vp$ is an ordered sequence $(x_1,\dots,x_l)$ of length $l$, where each element $x_i$ can either be a word $w_i$ or a visual representation of an object or frame of a scene $v_i$. 
Observations provided to the model are denoted as $o_t \in \Omega$, where $t$ represents the time step of the observation in the sequence.

Each action $a_t$ defines a linear movement between two end effector poses---where the robot arm moves linearly from the start pose $\mathcal{T}_{\mathrm{start}}$ to the end pose $\mathcal{T}_{\mathrm{end}}$ before retracting.
Each pose is defined in the special Euclidean group $\textbf{SE}(3)$ and represented as the state vector $(x,y,z,qw,qx,qy,qz)$, where $x,y,z$ are Cartesian coordinates and $qw,qx,qy,qz$ are quaternion components representing the orientation of the end effector.

The history trajectory $h_t$ consists of pairs of past observations and actions up to time step $t$, with the final element being the observation at time step $t$. Formally, each history trajectory is structured as $h_t = (o_0, a_0, o_1,\dots,a_{t-1},o_t)$. 
Consequently, the history trajectory space for time step $t$ can be defined as $\mathcal{H} = (\Omega \times \mathcal{A})^t \times \Omega$.

\paragraph{Training objective}

Similar to \citet{JiangEtAl2023VIMAGeneralRobot}, the model is trained through behaviour cloning of expert demonstrations \citep{Duan2017OneShotImitationLearning} that minimises a loss function for a trajectory of $T$ actions given by \cref{eq:loss-fn}:%
\begin{equation}\label{eq:loss-fn}
L(\theta) = \frac{1}{T} \sum^{T}_{t=0} \log\pi_\theta(a_t | \vp, h_t)
\end{equation}
Notably, the loss function was modified to prevent the model from being influenced by the trajectory length \citep{Pantazopoulos2023MultitaskMultimodalPrompted}.

\subsection{Implementation Details}

To allow for a fair comparison, all model code uses the code provided from \citet{JiangEtAl2023VIMAGeneralRobot}. Various alterations were made to capture metrics and improve performance, however all architectures are identical. Hyperparameters per component follow that stated in Appendix C in \citet{JiangEtAl2023VIMAGeneralRobot}.

Following \citet{Brohan2023RT1RoboticsTransformer} and \citet{JiangEtAl2023VIMAGeneralRobot}, each coordinate of the pose is predicted separately into one-of-$n$ bins. We follow \citet{JiangEtAl2023VIMAGeneralRobot}, where each coordinate per pose is discretised into 50 bins, with the exception of the $y$-position which is discretised into 100 bins. For each action dimension, the bin width is uniform across the total action space of the environment.

\subsection{Training Hyperparameters}\label{app:hparams}

To control for possible confounding variables across all models, we use the same training hyperparameters from Appendix D in \citet{JiangEtAl2023VIMAGeneralRobot} and from the various GitHub issues. We report a comprehensive table of hyperparameters in \cref{tab:hyperparams}. Across all models that were trained, these hyperparameters were kept constant and no hyperparameter sweeps were performed. 
All models were trained for 10 epochs and we used the checkpoint created at the end of epoch 10.

\paragraph{Computation Budget}
All models were trained using four~NVIDIA A100 40GB GPUs, with each run taking approximately 10 hours. Each evaluation run on the environment took approximately 2 hours and did not require the use of any GPUs. Therefore, the total computational budget for this work is 480 GPU hours.

\paragraph{Pretrained Language Model}
Following \citet{JiangEtAl2023VIMAGeneralRobot} and \citet{OctoModelTeam2023OctoOpenSourceGeneralist}, we also use the pretrained encoder from \texttt{t5-base} \citep{RaffelEtAl2020ExploringLimitsTransfer} as the pretrained language model that encodes multimodal instructions. Additionally, following \citet{JiangEtAl2023VIMAGeneralRobot} and \citet{TsimpoukelliEtAl2021MultimodalFewShotLearning}, we unfreeze the last two layers of the T5 encoder during training.

\paragraph{Learning Rate Schedule}
While our training process is similar to \citet{JiangEtAl2023VIMAGeneralRobot}, preliminary experiments showed that using a cosine annealing learning rate schedule that reduced the learning rate to the end of the 10th epoch performed better than annealing to 17K steps and training the model at \num{e-7} for 5 epochs.

\subsection{Training Components from Scratch}

Following \citet{JiangEtAl2023VIMAGeneralRobot}, the instruction encoder was the only pretrained component---using \texttt{t5-base} \citep{RaffelEtAl2020ExploringLimitsTransfer}; \textit{all other components} were trained from scratch. 

\paragraph{Segmentation Masks}
We used the ground-truth segmentation masks during training and evaluation over a trained object detector model because there is minimal performance difference between using a ground truth predictor and one that was trained for the task \citep{JiangEtAl2023VIMAGeneralRobot,OctoModelTeam2023OctoOpenSourceGeneralist}. As a result, this allows us to control for possible confounding variables from propagated errors.

\setcounter{table}{0}
\setcounter{figure}{0}
\section{Environment Details}\label{app:environment-details}

In this section, we further outline details of \VIMABench from \citet{JiangEtAl2023VIMAGeneralRobot}.
Built on top of the Ravens simulator \citep{ZengEtAl2021TransporterNetworksRearranging}, \VIMABench contains 17 tabletop object manipulation tasks to assess the capabilities learned by VLMs through a four-level protocol that evaluates their systematic generalisation capabilities.
All models are trained using behavioural cloning from 50K expert demonstrations for each of 13 tasks, with 4 tasks held out for zero-shot evaluation.

\subsection{Skills Models Must Learn to Perform}

One of the benefits of \VIMABench is that models must learn skills either in isolation or in combination with other skills, which is a desirable capability of intelligent systems \citep{lake2017building}. 

\begin{enumerate}
    \item \textbf{Simple Object Manipulation.} Picking up objects from a name or a visual representation, and placing them in specific locations and positions. 
    \item \textbf{Visual Goal Completion.} Manipulating objects to match the scene in the provided frame.
    \item \textbf{Visual Memory}. After performing actions, remember the previous state of the workspace and perform an action given information from that time. 
    \item \textbf{Visual Reasoning}. Only performing actions on objects that have the same colours/shapes as in the instruction.
    \item \textbf{One-Shot Imitation}. Imitate the actions necessary to make the workspace look like a given sequence of frames. 
    \item \textbf{Novel Concept Grounding}. The prompt contains unfamiliar words like \textit{``dax''} which are explained through visual referents and used within an instruction similar to multimodal in-context learning \citep{zhang2023multimodal}.
\end{enumerate}

\subsection{Different Levels of Generalisation}\label{app:gen-levels}

\VIMABench uses tiers of generalisation levels to enable more precise assessment of a model's capabilities in the environment by testing its adaptability conditions unseen during training that are either object or instruction specific, as described below:

\paragraph{Placement Generalisation (L1)} 

Object poses---starting positions and orientation---are novel. Failure at this level indicates that model learning is not invariant to object poses, and therefore indicates the model is unable to generalise beyond how objects are positioned in training data.

\paragraph{Combinatorial Generalisation (L2)} 

Object shape and texture combinations are novel (e.g., the model has seen either red objects and squares during training, but never a red square). Failure indicates an inability learn and/or combine object-specific information, therefore unable to perform systematicity within the visual scenes.

\paragraph{Novel Object Generalisation (L3)} 
Objects shapes and textures are novel (e.g., the model has never seen blue objects or triangles during training). Failure at this level indicates difficulty in abstracting object-specific information beyond the training corpus.

\paragraph{Novel Task Generalisation (L4)} 
Tasks (including instructions and success criteria) have never been seen. Failure at this level indicates an inability to perform compositional generalisation to combine skills/movements to solve novel tasks.

\subsection{Dataset Preparation for Training}\label{app:dataset-prep}

We parse all \num{664976} instances across the \num{13} tasks used for training---as provided by \citet{JiangEtAl2023VIMAGeneralRobot}---each containing an action trajectory created by a scripted oracle. We create a validation set using stratified sampling such that a total of \num{50000} instances across all the tasks are held out.%
\footnote{Authors state that they held out \num{50000} examples for validation on their GitHub: \url{https://github.com/vimalabs/VIMA/issues/8\#issuecomment-1491255242}.}
Each instance is prepared for training in advance by tokenizing any natural language and preparing visual features for the model. 
We release all code used to prepare the dataset as well as the examples for each split, both before and after preprocessing (see \cref{app:reproducibility} for more).

\begin{table}[tb]
\centering
\footnotesize
\renewcommand{\arraystretch}{1.3}
\setlength{\tabcolsep}{4pt}
\begin{tabularx}{\linewidth}{@{}X *{13}{c}@{}}
\toprule
Task & 1 & 2 & 3 & 4 & 5 & 6 & 7 & 9 & 11 & 12 & 15 & 16 & 17 \\
\midrule
Minimum & 1 & 1 & 1 & 1 & 1 & 1 & 1 & 1 & 2  & 1  & 2  & 2  & 3  \\
Maximum & 2 & 3 & 2 & 4 & 7 & 2 & 3 & 3 & 2  & 8  & 4  & 4  & 4  \\
\bottomrule
\end{tabularx}
\caption{Minimum and maximum actions taken to solve each task across all episodes within the training data. Missing tasks (8, 10, 13, 14) do not appear in the training data as they are only seen when evaluating unseen tasks (L4).}
\label{tab:steps-training-data}
\end{table}

\subsubsection{Error Recovery Is Not Emergent Behaviour}\label{app:data-not-optimal}

We analysed the expert trajectories used to train the model from the \VIMABench dataset to determine whether models are only shown the most efficient solution. 
\cref{tab:steps-training-data} shows the minimum and maximum number of actions shown to models to solve each task from the given expert trajectories. 
The minimum number of moves required per task is dependent on the number of objects and parameters for a given episode. They are not identical for all episodes.
We found multiple observation-action pairs in several examples, showing that \VIMABench contains expert trajectories that are not always optimal, thereby suggesting that recovering from mistakes is not an emergent behaviour of the models.

\setcounter{table}{0}
\setcounter{figure}{0}
\begin{table*}[bt]
\centering
\footnotesize
\renewcommand{\arraystretch}{1.4}
\begin{tabular}{@{} llll c c@{}}
    \toprule
    Instruction-style & Instruction Modalities & Prompt-conditioning & Vision Encoder & Shuffled Objects? & Model ID \\
    \midrule
    Original & Text + Visual & Cross-Attention & Object-Centric & False & \texttt{8lkml12g} \\
    Original & Text + Visual & Cross-Attention & Object-Centric & True & \texttt{ftwoyjb1} \\
    Original & Text + Visual & Cross-Attention & Image-Patches & N/A & \texttt{ln4nrqhg} \\
    Original & Text + Visual & Concatenate & Object-Centric & False & \texttt{bhuja4vo} \\
    Original & Text + Visual & Concatenate & Object-Centric & True & \texttt{wn9jc5l8} \\
    Original & Text + Visual & Concatenate & Image-Patches & N/A & \texttt{efxugme9} \\
    Paraphrases & Text + Visual & Cross-Attention & Object-Centric & False & \texttt{2df3mwfn} \\
    Paraphrases & Text + Visual & Cross-Attention & Object-Centric & True & \texttt{0nsnkaer} \\
    Paraphrases & Text + Visual & Cross-Attention & Image-Patches & N/A & \texttt{ah5btw8w} \\
    Paraphrases & Text + Visual & Concatenate & Object-Centric & False & \texttt{fs5v61mz} \\
    Paraphrases & Text + Visual & Concatenate & Object-Centric & True & \texttt{xb3yttg9} \\
    Paraphrases & Text + Visual & Concatenate & Image-Patches & N/A & \texttt{zby6xk27} \\
    \bottomrule
\end{tabular}
\caption{Unique ID for each model checkpoint to aid with reproducibility and the conditions they were trained on.}
\label{tab:model-ids-desc}
\end{table*}

\section{Reproducibility}\label{app:reproducibility}

We are deeply committed to reproducibility in ML research. To this end, we provide a fully reproducible training and evaluation framework at \url{https://github.com/amitkparekh/CoGeLoT}.

\paragraph{License}
\VIMABench from \citet{JiangEtAl2023VIMAGeneralRobot}, including model code, pre-trained checkpoint, and the \VIMABench environment are licensed under MIT. All artefacts produced from this work will also be released under the MIT license. 

\paragraph{Codebase}
We are providing our entire codebase---the full, unabridged version we used throughout development, training, and evaluation. This includes implementations for every perturbation, including the \GDG perturbations, to encourage use in other evaluation settings and benchmarks.

\paragraph{Training Data}
We are releasing all training data, including the exact training/validation splits used, using the process outlined in \cref{app:dataset-prep}. 
Our codebase includes the methodology for generating these from the original \VIMABench dataset, which did not include pre-defined splits. 
Additionally, we provide additional datasets with paraphrased multimodal instructions, along with the commands used to create them.
For all datasets splits and variations, we provide the pre-processed instances---stripped of unnecessary metadata and with tokenised instructions with T5---that we used to accelerate model training. All datasets are hosted on our Hugging Face repository\footnote{ \url{https://huggingface.co/datasets/amitkparekh/vima}}, and we recommend using them with our provided framework. 

\paragraph{Model Checkpoints}\label{app:checkpoints}
We provide every model checkpoint used in our evaluation, including checkpoints from earlier training epochs, to facilitate further interpretability experiments and explorations.
\cref{tab:model-ids-desc} provides a list of unique IDs for each trained model, along with the architecture used. 
These IDs can be used to source model checkpoints from our Hugging Face repository\footnote{\url{https://huggingface.co/amitkparekh/cogelot}}, or using our provided framework.
As mentioned in \cref{app:hparams}, we only evaluate models after completing all 10 training epochs. However, we provide checkpoints created at the end of each epoch to support future work.

\paragraph{Reproducibility}

We trained our models using PyTorch \citep{Ansel2024PyTorchFasterMachine} and Lightning \citep{Falcon2024PyTorchLightning}, and tracked all dependencies with PDM\footnote{\url{https://pdm-project.org/}}.
We are providing all components, including a Docker image, to facilitate replication. 
Our experiments were managed using Hydra configuration files \citep{Yadan2019Hydra}, and we are sharing all configurations, commands, hyperparameters, and seeds used. 
Our codebase is designed to automatically download the required datasets and models from our Hugging Face repositories when run with the provided configurations and commands, mirroring our exact training and evaluation process.

% \VIMABench from \citet{JiangEtAl2023VIMAGeneralRobot}, including model code, pre-trained checkpoint, and the \VIMABench environment are licensed under MIT. All artefacts produced from this work will be released under the same license. 

\begin{table*}[tb]
\centering
\footnotesize
\sisetup{mode=match,tight-spacing=true,table-format=2.1,table-number-alignment=center}
\renewcommand{\arraystretch}{1.3}
\setlength{\tabcolsep}{4.5pt}
\begin{tabular}{@{} l *{18}{S} @{}}
\toprule
& {T1} & {T2} & {T3} & {T4} & {T5} & {T6} & {T7} & {T8} & {T9} & {T10} & {T11} & {T12} & {T13} & {T14} & {T15} & {T16} & {T17} & {Avg.} \\
\midrule
\multicolumn{18}{@{}l}{\textit{\textbf{Reported in \citet{JiangEtAl2023VIMAGeneralRobot}}}} \\
\addlinespace[1pt]
\textbf{L1} &100.0 & 100.0 & 99.5 & 100.0 & 56.5 & 100.0 & 100.0 & {---} & 18.0 & {---} & 77.0 & 93.0 & {---} & {---} & 97.0 & 76.5 & 43.0 & 81.6 \\
\textbf{L2} &100.0 & 100.0 & 99.5 & 100.0 & 54.5 & 100.0 & 100.0 & {---} & 17.5 & {---} & 77.0 & 93.0 & {---} & {---} & 98.5 & 75.0 & 45.0 & 81.5 \\
\textbf{L3} &99.0 & 100.0 & 100.0 & 97.0 & 54.5 & 100.0 & 99.0 & {---} & 17.5 & {---} & 90.5 & {---} & {---} & {---} & 97.5 & 46.0 & 43.5 & 70.4 \\
\textbf{L4} &{---} & {---} & {---} & {---} & {---} & {---} & {---} & 100.0 & {---} & 0.0 & {---} & {---} & 0.0 & 94.5 & {---} & {---} & {---} & 48.6 \\
\midrule[0.5pt]
\multicolumn{18}{@{}l}{\textit{\textbf{From the Provided Checkpoint}}} \\
\addlinespace[1pt]
\textbf{L1} &93.0 & 93.5 & 99.5 & 85.0 & 49.5 & 93.5 & 95.5 & {---} & 14.5 & {---} & 90.5 & 96.0 & {---} & {---} & 5.0 & 43.5 & 3.0 & 66.3 \\
\textbf{L2} &92.0 & 93.0 & 100.0 & 89.5 & 55.0 & 91.5 & 91.0 & {---} & 16.0 & {---} & 84.0 & 95.5 & {---} & {---} & 7.0 & 40.5 & 0.5 & 65.8 \\
\textbf{L3} &91.5 & 94.5 & 99.5 & 83.0 & 51.5 & 87.0 & 90.5 & {---} & 20.0 & {---} & 93.5 & {---} & {---} & {---} & 6.0 & 35.5 & 2.0 & 62.9 \\
\textbf{L4} &{---} & {---} & {---} & {---} & {---} & {---} & {---} & 80.0 & {---} & 2.0 & {---} & {---} & 0.0 & 4.5 & {---} & {---} & {---} & 21.6 \\
\bottomrule
\end{tabular}
\caption{Comparing the average success rate per task as reported by \citet{JiangEtAl2023VIMAGeneralRobot} with our results obtained from running the checkpoint provided in the environment. Each task was run for 200 samples.}
\label{tab:reprod-results}
\end{table*}

\subsection{Discrepancies in Reported Results}\label{app:cannot-reproduce}

% Note: This section is inspired by Footnote 5, Page 7 from Emergent Communication by Noukhovitch 2021. 

\citet{JiangEtAl2023VIMAGeneralRobot} only provided the code for the model and the dataset did not contain a train-test split. 
After creating a working codebase, we were unable to reproduce the results reported by \citet{JiangEtAl2023VIMAGeneralRobot} using the provided model checkpoint. 
We spent several weeks trying to reproduce the results, including consulting the original authors on their experimental setup, but were unsuccessful in doing so. 
\cref{tab:reprod-results} contains the reported results from \citet{JiangEtAl2023VIMAGeneralRobot} and our results when running the evaluation on their provided checkpoint. 
For this comparison, \textit{no new models were trained}.
Note that the provided checkpoint uses cross-attention to condition prompts and object-centric visual features. 
Across all tasks/generalisation levels (with the exception of T3), task success is significantly lower than what was reported. Possible reasons for this difference include: 

\begin{itemize}
    \item Pure randomness as only 200 episodes are sampled per task, and the exact episodes are not compared.
    \item There may be a different checkpoint provided compared to the paper.
    \item Possible misunderstandings during re-implementation.
\end{itemize}

% \subsection{Provided Model Checkpoints}\label{app:checkpoints}

% To aid with reproducibility of our results, \cref{tab:model-ids-desc} includes a list of unique IDs associated with each model that we trained and the architecture choices used. 
% These IDs can be used to source model checkpoints from \url{https://huggingface.co/amitkparekh/cogelot}, or using the provided framework.
% As mentioned in \cref{app:hparams}, we only evaluate models at the end of all 10 training epochs. However, we provide the checkpoints that were created at the end of each epoch to support extensions to this work.

\setcounter{table}{0}
\setcounter{figure}{0}
\section{Evaluation Details}\label{app:evaluation-details}

\subsection{Estimating Random Chance}\label{app:random-chance}

The model predicts actions by mapping embedded action tokens to the action space, which consists of 14 coordinates across two $\textbf{SE}(3)$ poses. Each pose has seven coordinates that predict a discrete bin. There are 50 discrete bins for each axis, except for the $y$-position which has 100. 

To correctly predict a movement, the model must accurately predict 14 coordinates. 
Assuming each axis is predicted independently, and that the likelihood of choosing each discrete bin per coordinate is equal, the probability of randomly predicting the correct action is $1 / (50 \times 12 + 100 \times 2) = 1 / 800 = 0.125\%$. 
Assuming each predicted action is i.i.d., for a task requiring $t$ time steps, the probability that a model will randomly succeed is $0.00125^t$.

\subsection{Sample Size for Computing Task Performance}
\citet{JiangEtAl2023VIMAGeneralRobot} claimed to run each task in the environment for 100 episodes.%
\footnote{While not reported in the final manuscript, it was mentioned on their public GitHub repository: \url{https://github.com/vimalabs/VIMA/issues/16\#issuecomment-1622973970}.} 
However, we assume there is some inconsistency in the statement as the reported success rates consist of multiples of ``0.5''. 
Furthermore, due to inconsistencies in the environment, the model will not view the same instantiation of each 200 episodes.
As a result, we assume that running 200 samples is large enough to fall under the law of large numbers.
\citet{LiEtAl2023MasteringRobotManipulation} also sampled 200 episodes for each task during evaluation on \VIMABench.

\subsection{When Does an Evaluation Episode End?}\label{sec:end-of-eval-episode}

During the online evaluation, the episode ends when one of two conditions are met: 

\begin{enumerate}
    \item the model has successfully completed the instruction with the previous action it took; or,
    \item the model has not successfully completed the instruction within a maximum of \num{10} actions. 
\end{enumerate}

A maximum length of \num{10} actions is longer than the default length used by \citet{JiangEtAl2023VIMAGeneralRobot}.

\subsection{\textit{Gobbledygook} Perturbations}\label{app:gdg}

We outline how \GDGWords and \GDGTokens manipulate multimodal instructions to remove all linguistic information without altering the positions of any visual referents.

% method 1: GDGWords
\paragraph{\textit{Gobbledygook Words}}
% Define the method
Let $w_i = (c_1, c_2, \dots, c_j)$ represent a word with $j$ characters, where each character is from a set $\sA$ containing all uppercase and lowercase alphabetical English characters. 
Given a multimodal prompt $\vp$ of multiple words, we transform the sequence by: first replacing each character per word with a random choice from $\sA$, then randomly swap the positions of words within the sequence without changing the position of any visual representations within the sequence.

% Method 2: GDGTokens
\paragraph{\textit{Gobbledygook Tokens}}
This method transforms the multimodal prompt by randomising each sub-word unit after tokenizing the instruction with any other token from the vocabulary such that the number of sub-word units is the same as the original instruction. See \cref{fig:lang-perturbation-example} for an example where an instruction perturbed with \GDGTokens does not contain any information in the language modality pertaining to the original task.

\begin{table}[tb]
\centering
\footnotesize
\renewcommand{\arraystretch}{1.3}
\sisetup{uncertainty-mode=separate,table-align-uncertainty=true,group-separator=\pm}
\begin{tabularx}{\linewidth}{@{} X S[table-format=2.1(2.1)] S[table-format=2.1(3.1)] @{}}
    \toprule
     & {\# Words} & {\# Tokens} \\
    \midrule
    Original Instruction & 12.9 (7.6) & 20.2 (13.6)\\ 
    \GDGTokens & 15.2 (9.3) & 20.2 (13.6) \\
    \GDGWords & 12.9 (7.6) & 49.7 (27.8)\\
    \bottomrule
\end{tabularx}
\caption{Average length of instructions (with standard deviation), both before and after transforming through a language perturbation method. A single word is defined as sequences of alphanumeric characters delimited by a whitespace character. Tokens are defined as the number of IDs returned from the tokenizer.}
\label{tab:gdg_instruction_length}
\end{table}

\paragraph{Controlling for sequence lengths}
To avoid introducing additional difficulty into the tasks, we ensure that the length of the instruction is identical to before perturbing for either natural language words or the tokenised form.
\cref{tab:gdg_instruction_length} further verifies this as the number of words in an instruction does not change for \GDGWords, and the number of tokens does not change for \GDGTokens.
It also allows for checking whether or not the length of the instruction in natural language has any impact on model performance. 

As illustrated in \cref{fig:lang-perturbation-example}, \GDGWords ensures that the number of characters and ``words'' within the multimodal prompt---and the number of words between each visual placeholder---does not change. However, the average length of the prompt after tokenizing has increased because T5 uses a SentencePiece tokenizer that was trained on natural language text \citep{RaffelEtAl2020ExploringLimitsTransfer}.

\begin{figure}[tb]
    \centering
    \includegraphics[width=\linewidth]{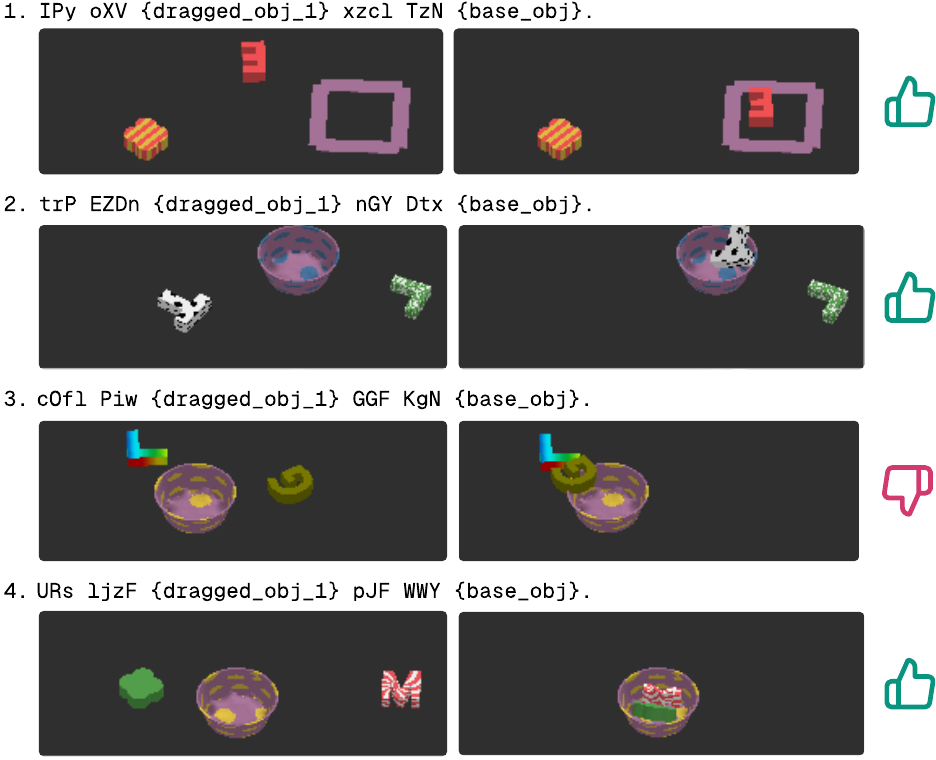}
    \caption{In-environment observations seen by the model, showing task performance when using \textit{Gobbledygook Words}. Instructions given to the model are shown on top of the images, with the images themselves showing different iterations of either success (see 1, 2, and 4) or failure (see 3).}
    \label{fig:success-with-gdg-example}
\end{figure}

\paragraph{In-environment examples after applying \textit{Gobbledygook Words}}
\cref{fig:success-with-gdg-example} contains some examples where the model still succeeds in performing the task, even when provided with perturbed language from \GDGWords. From \cref{fig:success-with-gdg-example}, Examples 1 and 2 both show that the model followed through on incomprehensible instructions and successfully performed the tasks of: identifying the task to perform with the stated object from a choice of two, picking it up, and putting it into a destination. 
Example 4 indicates interesting behaviour as the model continued to place all objects into the container to end the episode.\footnote{We outline the termination conditions for a given episode in \cref{sec:end-of-eval-episode}.} 
Such a failure is indicated in Example 3, where the model picked the object and placed it onto the receptacle in a way that resulted in a scenario it could not recover from, having chosen the wrong object to place and by balancing it on the edge of the container.

\begin{figure}[tb]
\centering
\includegraphics[width=0.8\linewidth]{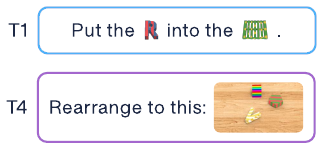}
\caption{Example instruction for T1 (pick and place) and T4 (rearrange to this scene).}
\label{fig:t1-t4}
\end{figure}

\subsection{Which Visual Referents Can Be Substituted as Text?}\label{app:textual-details}

There are two types of visual referents that appear in \VIMABench: ones that refer to a single object, and ones that represent an object \textit{within a scene}. For example, as shown in \cref{fig:t1-t4}, T1 directly refers to an object whereas T4 directly includes a frame of a scene. 
As a result, it does not make sense to convert tasks that include frames or scenes in their instruction as the textual description can refer to more than necessary. 
In total, 9 of the 17 tasks (across all 4 generalisation levels) use instructions that do not use frames.

\setcounter{table}{0}
\setcounter{figure}{0}
\section{Extensions to VIMA-Bench}

In this work, we propose multiple extensions to \VIMABench. In this section, we provide further analysis and details for each. 

\subsection{Increasing Difficulty Across All Tasks}\label{app:difficulty-levels}

\begin{figure}[tbh]
    \centering
    \includegraphics[width=\linewidth]{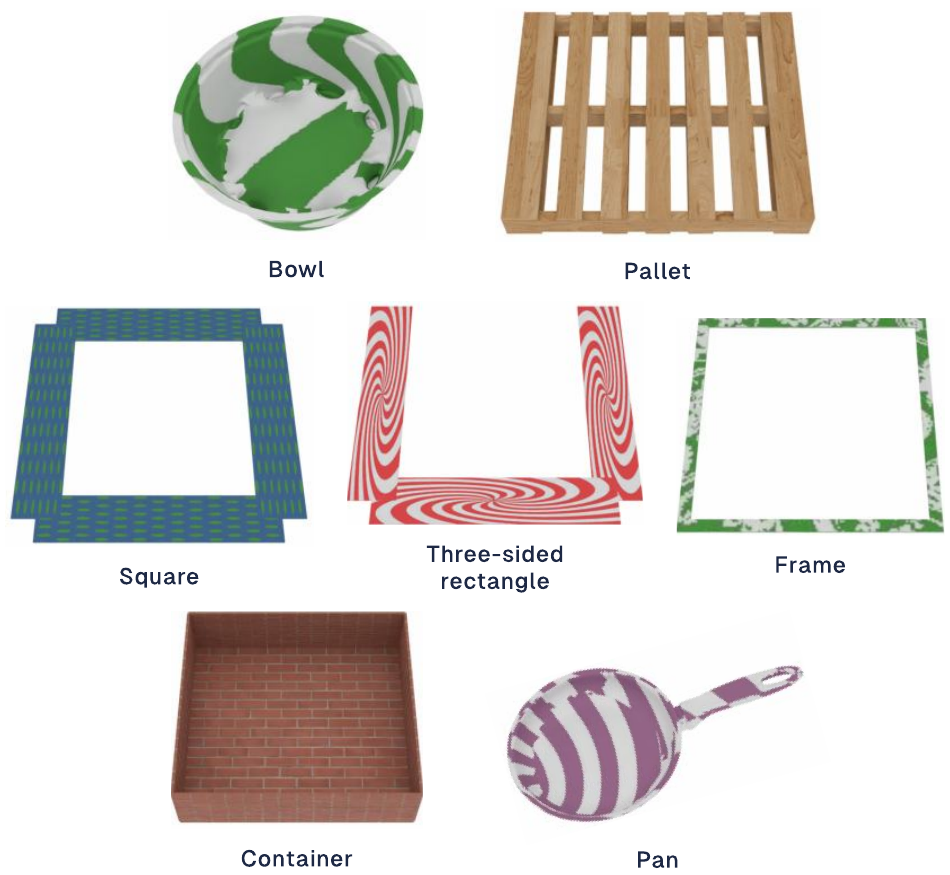}
    \caption{Objects within \VIMABench that are often regarded as ``containers''; i.e., other objects are always placed within these.}
    \label{fig:container-subset}
\end{figure}

\cref{tab:meaning-of-difficulty} outlines the changes made for each difficulty level for each task. The \textit{Distracting} difficulty level focuses on drastically increasing the number of distractors in the scene to try and confuse the model, whereas the \textit{Extreme} difficulty level alters the parameters of the task to check whether a model is over-reliant on the parameters seen during training. 
Additionally, a subset of objects in \VIMABench is always used as ``containers'' (\cref{fig:container-subset}): objects are always put into/onto them across all tasks. Therefore, as part of the \textit{Extreme} difficulty, the container/destination object is just any other acceptable object (within the generalisation level constraints) that is not one of these.

\begin{table*}[tbhp]
\centering
\scriptsize
\renewcommand{\arraystretch}{1.4}
\begin{tabularx}{\textwidth}{@{}p{0.2cm} X X X @{}}
\toprule
 & Description & Distracting & Extreme \\
\midrule
T1 
& Put specified objects into specified container.
& Distractors: 1 $\rightarrow$ 6 
& Containers are now one of the draggable objects instead of the designated container shapes \\
T2 
& Place objects with specified texture \textit{from the given frame} into container with specified colour.
& Distractors in frame: 1 $\rightarrow$ 3. Distractors in workspace: 1 $\rightarrow$ 3 
& Containers are now one of the draggable objects instead of the designated container shapes \\
T3 
& Rotate the specified object by the given number of degrees.
& Distractors: 1 $\rightarrow$ 8
& Possible Angles of rotation: From [30, 60, 90, 120, 150] to [20, 40, 60, 80, 100, 120, 140, 160] \\
T4 
& Look at the objects within the frame and move the objects in the workspace to those positions. Distractors in the workspace may need to be moved out the way. Not all objects in the workspace are visible in the frame. 
& Distractors in workspace: 2 $\rightarrow$ 3
& Distractors in workspace will ALWAYS be in the way (therefore the model must move them out the way to complete the task) \\
T5 
& Perform T4, and then put all the objects back to the start
& Distractors in workspace: 2 $\rightarrow$ 3
& Distractors in workspace will ALWAYS be in the way (therefore the model must move them out the way to complete the task) \\
T6 
& Compare the size or texture saturation of objects and make adjustments to the specified object(s) accordingly.
& Distractors: 1 $\rightarrow$ 3 
& All container shapes are replaced with other shapes. Adjective word choices are now: ``xachup'', ``feplicat'', ``gazip'', or ``duchat''. \\
T7 
& Apply novel words to two objects (one is a container class), and put one object into the container.
& Distractors: 1 $\rightarrow$ 3 
& All container shapes are replaced with other shapes. Noun word choices are now: \\
T8 
& Combination of T6 and T7
& Combination of T6 and T7.
& All container shapes are replaced with other shapes. \\
T9 
& Determine the degrees to rotate an object from three before/after demonstrations (i.e., 3-shot demonstration to learn the task)
& Total number of objects: 3 $\rightarrow$ 8
& Possible Angles of rotation: [30, 60, 90, 120, 150, 180, 210, 240, 270, 300, 330] $\rightarrow$ [20, 40, 60, 80, 100, 120, 140, 160] \\
T10
& Follow motions for specific objects from demonstrations of frames
& Distractors in workspace: 1 $\rightarrow$ 3. Distractors in frames: 1 $\rightarrow$ 3
& Possible motion points: 5 $\rightarrow$ 10 \\
T11 
& Stack objects with the order illustrated in given frames.
& Distractors in workspace: 1 $\rightarrow$ 3
& Objects in workspace: 3 $\rightarrow$ 5 \\
T12 
& Sweep the objects into the region \textbf{without exceeding} the boundary
& Objects in the scene 1--5 $\rightarrow$ 6--10 
& Sweepable objects are now any dragged object  \\
T13 
& Sweep the objects into a region \textbf{without touching} the constraint.
& Objects in the scene 1--5 $\rightarrow$ 6--10
& Sweepable objects are now any dragged object \\
T14 
& Pick all objects in the workspace with the same texture as the container object specified in the prompt, into it.
& Distractors: 1 $\rightarrow$ 5
& All container shapes are replaced with other shapes. \\
T15 
& Put all objects in the workspace with the same top-down profile goal container into it.
& Distractors: 1 $\rightarrow$ 5
& All container shapes are replaced with other shapes. \\
T16 
& Put the target object into the container, and then put one of its old neighbours into the same container
& Distractors: 1 $\rightarrow$ 3
& Density grid of objects: 3$\times$3 $\rightarrow$ 4$\times$4 \\
T17 
& Pick and place the object into different containers in order then restore to the initial container.
& Distractors: 0 $\rightarrow$ 4
& All containers can be different types of shapes \\
\bottomrule
\end{tabularx}
\caption{Descriptions of each task, number of distractors added to increase difficulty, and description of the extreme difficulty for each.}
\label{tab:meaning-of-difficulty}
\end{table*}

\subsection{Paraphrasing Multimodal Instructions}\label{app:paraphases}

\begin{table*}[tbhp]
\centering
\scriptsize
\renewcommand{\arraystretch}{1.3}
\begin{tabularx}{\textwidth}{@{}cXX@{}}
    \toprule
    Task & Original & Alternative \\
    \midrule
    1 & Put the \texttt{blue spiral} object in \texttt{\{scene\}} into the \texttt{wooden} object. & From the \texttt{\{scene\}} stack the \texttt{blue spiral} object on the \texttt{wooden} thing. \\
    2 & Put the \texttt{\{dragged\_texture\}} object in \texttt{\{scene\}} into the \texttt{\{base\_texture\}} object. & Move objects in the \texttt{\{scene\}} so that the \texttt{\{dragged\_texture\}} item is on one \texttt{\{base\_texture\}} item. \\
    3 & Rotate the \texttt{\{dragged\_obj\}} \texttt{\{angle\_in\_degree\}} degrees. & Turn the \texttt{\{dragged\_obj\}} precisely \texttt{\{angle\_in\_degree\}} degrees. \\
    4 & Rearrange to this \texttt{\{scene\}}. & Rearrange things into this setup \texttt{\{scene\}}. \\
    5 & Rearrange objects to this setup \texttt{\{scene\}} and then restore. & Rearrange objects into this configuration \texttt{\{scene\}} and put it back. \\
    6 & \texttt{\{demo\_blicker\_obj\_1\}} is kobar than \texttt{\{demo\_blicker\_obj\_2\}}. \texttt{\{demo\_blicker\_obj\_3\}} is kobar than \texttt{\{demo\_blicker\_obj\_4\}}. Put the kobar \texttt{\{dragged\_obj\}} into the \texttt{\{base\_obj\}}. & \texttt{\{object1\}} \texttt{\{object3\}} and \texttt{\{object5\}} are all kobar than objects \texttt{\{object2\}} \texttt{\{object4\}} and \texttt{\{object6\}} respectively. move the kobar \texttt{\{dragged\_obj\}} inside of the \texttt{\{base\_obj\}}. \\
    7 & This is a blinket \texttt{\{dragged\_obj\}}. This is a zup \texttt{\{base\_obj\}}. Put a zup into a blinket. & This is a blinket \texttt{\{object2\}}. this is a zup \texttt{\{object1\}}. drop the zup inside of the blinket.\\
    11 & Stack objects in this order: \texttt{\{frame1\}} \texttt{\{frame2\}} \texttt{\{frame3\}}. & Move objects like this: \texttt{\{frame1\}} \texttt{\{frame2\}} \texttt{\{frame3\}}. \\
    16 & First put \texttt{\{object1\}} into \texttt{\{object2\}} then put the object that was previously at its \texttt{\{direction\}} into the same \texttt{\{object2\}}. & Set \texttt{\{object1\}} in \texttt{\{object2\}} then place the item that was at its \texttt{\{direction\}} before you placed it into the same place.\\
    17 & Put \texttt{\{object1\}} into \texttt{\{object2\}}. Finally restore it into its original container. & Set \texttt{\{object1\}} within \texttt{\{object2\}} then restore it to its original place. \\
    \bottomrule
\end{tabularx}
\caption{Examples of how each original instruction was converted into an alternative paraphrase using the meta-templates.}
\label{fig:paraphrase-example}
\end{table*}

We created paraphrases by manually inspecting the instructions and using meta-templates to construct variations.
Notably, we were careful to avoid introducing ambiguity that could introduce any misunderstanding into the semantic meaning of the instruction.
As a result, only the natural language words are altered; any novel words (as in T6--8) remained unchanged. The observations seen, the actions the model must perform, and the instances for each train-valid-test split are unchanged.
We provide examples of some paraphrased alternatives of the original instruction in \cref{fig:paraphrase-example}. All meta-templates used for each task are included within the provided source code.

\setcounter{table}{0}
\setcounter{figure}{0}

% \clearpage

\FloatBarrier

\section{Further Experimental Results}\label{app:further-experimental-results}

\subsection{Per-Task Results}

We report the per-task results for each table reported in the main paper. \cref{tab:per-task-table} contains a mapping from each table in the paper to the one with the per-task results.
Some tasks only exist for certain generalisation levels and therefore are left blank for other levels.

\begin{table}[tbh]
\centering
\footnotesize
\renewcommand{\arraystretch}{1.3}
\begin{tabular}{@{} l l @{}}
\toprule
Per-Level & Per-Task \\
\midrule
\cref{tab:perf-paraphrases}  & \cref{tab:trained-on-orig-full} and \cref{tab:trained-on-para-full} \\
\cref{tab:perf-obj-text-para} & \cref{tab:full-obj-as-text} \\
\cref{tab:perf-gdg-on-para} & \cref{tab:full-gdg} \\
\cref{tab:perf-mask-modalities-on-para} & \cref{tab:mask-modality-full} \\
\cref{tab:perf-no-mistakes-on-para} & \cref{tab:strict-time-full} \\
\cref{tab:perf-no-instructions-trained-para} & \cref{tab:no-instruct-full} \\
\cref{tab:perf-difficulty-on-para} & \cref{tab:difficulty-full} and \cref{tab:difficulty-no-prompt-full} \\
\cref{tab:perf-eval-shuffle-on-para} & \cref{tab:shuffle-para-full}, \cref{tab:shuffle-difficulty-full}, \cref{tab:trained-shuffle-full}, and \cref{tab:trained-shuffle-full-strict} \\
\bottomrule
\end{tabular}
\captionof{table}{Mapping of per-task results for each table listed in the main paper.}
\label{tab:per-task-table}
\end{table}

% \clearpage

\begin{table*}[tbhp]
\centering
\scriptsize
\sisetup{mode=match,tight-spacing=true,table-format=2.1,table-number-alignment=center}
\renewcommand{\arraystretch}{1.2}
\setlength{\tabcolsep}{4.5pt}
\begin{tabular}{@{} l *{18}{S} @{}}
\toprule
& {T1} & {T2} & {T3} & {T4} & {T5} & {T6} & {T7} & {T8} & {T9} & {T10} & {T11} & {T12} & {T13} & {T14} & {T15} & {T16} & {T17} & {Avg.} \\
\midrule
\multicolumn{18}{@{}l}{\textit{\textbf{Trained and Evaluated on Original Instructions}}} \\
\multicolumn{18}{@{}l}{\textit{Cross-Attn + Obj-Centric}} \\
\bfseries L1 & 100.0 & 99.5 & 99.5 & 97.0 & 8.5 & 100.0 & 100.0 & {---} & 19.0 & {---} & 91.5 & 96.0 & {---} & {---} & 96.5 & 49.5 & 73.5 & 79.3 \\
\bfseries L2 & 99.5 & 99.5 & 100.0 & 98.0 & 9.5 & 99.0 & 99.5 & {---} & 18.0 & {---} & 95.0 & 96.5 & {---} & {---} & 92.0 & 47.5 & 71.0 & 78.8 \\
\bfseries L3 & 100.0 & 99.0 & 100.0 & 99.0 & 10.0 & 98.5 & 99.5 & {---} & 14.0 & {---} & 90.5 & {---} & {---} & {---} & 93.0 & 43.0 & 21.5 & 72.3 \\
\bfseries L4 & {---} & {---} & {---} & {---} & {---} & {---} & {---} & 96.5 & {---} & 0.5 & {---} & {---} & 0.0 & 97.5 & {---} & {---} & {---} & 48.6 \\
\multicolumn{18}{@{}l}{\textit{Cross-Attn + Patches}} \\
\bfseries L1 & 91.5 & 75.0 & 97.5 & 12.0 & 1.0 & 76.5 & 95.0 & {---} & 9.0 & {---} & 90.5 & 93.0 & {---} & {---} & 79.5 & 95.5 & 2.5 & 63.0 \\
\bfseries L2 & 94.0 & 73.5 & 96.5 & 9.5 & 2.5 & 78.5 & 92.0 & {---} & 14.0 & {---} & 87.5 & 91.5 & {---} & {---} & 71.5 & 94.5 & 0.0 & 62.0 \\
\bfseries L3 & 57.0 & 70.0 & 68.0 & 9.0 & 0.5 & 72.5 & 57.5 & {---} & 11.5 & {---} & 85.0 & {---} & {---} & {---} & 62.0 & 44.0 & 2.0 & 44.9 \\
\bfseries L4 & {---} & {---} & {---} & {---} & {---} & {---} & {---} & 25.5 & {---} & 1.0 & {---} & {---} & 0.0 & 29.0 & {---} & {---} & {---} & 13.9 \\
\multicolumn{18}{@{}l}{\textit{Concatenate + Obj-Centric}} \\
\bfseries L1 & 100.0 & 100.0 & 99.5 & 97.0 & 19.0 & 100.0 & 100.0 & {---} & 13.5 & {---} & 88.5 & 95.0 & {---} & {---} & 96.0 & 45.5 & 75.5 & 79.2 \\
\bfseries L2 & 99.5 & 100.0 & 99.5 & 99.0 & 19.0 & 100.0 & 100.0 & {---} & 16.0 & {---} & 91.0 & 95.5 & {---} & {---} & 92.0 & 39.0 & 74.5 & 78.8 \\
\bfseries L3 & 98.0 & 97.0 & 100.0 & 99.0 & 21.0 & 92.0 & 96.5 & {---} & 18.5 & {---} & 95.0 & {---} & {---} & {---} & 96.5 & 43.0 & 68.5 & 77.1 \\
\bfseries L4 & {---} & {---} & {---} & {---} & {---} & {---} & {---} & 97.0 & {---} & 2.5 & {---} & {---} & 0.0 & 97.5 & {---} & {---} & {---} & 49.2 \\
\multicolumn{18}{@{}l}{\textit{Concatenate + Patches}} \\
\bfseries L1 & 96.0 & 84.5 & 97.5 & 13.0 & 2.5 & 87.0 & 95.0 & {---} & 42.5 & {---} & 96.0 & 96.5 & {---} & {---} & 75.0 & 94.0 & 4.0 & 68.0 \\
\bfseries L2 & 92.5 & 73.5 & 97.5 & 17.0 & 3.5 & 93.5 & 91.0 & {---} & 31.0 & {---} & 95.5 & 88.0 & {---} & {---} & 75.0 & 96.5 & 7.0 & 66.3 \\
\bfseries L3 & 71.5 & 66.5 & 91.0 & 12.5 & 3.5 & 93.0 & 58.0 & {---} & 30.5 & {---} & 87.0 & {---} & {---} & {---} & 58.5 & 61.0 & 2.0 & 52.9 \\
\bfseries L4 & {---} & {---} & {---} & {---} & {---} & {---} & {---} & 44.0 & {---} & 11.0 & {---} & {---} & 0.0 & 38.5 & {---} & {---} & {---} & 23.4 \\
\midrule[0.2pt]
\addlinespace[2pt]
\multicolumn{18}{@{}l}{\textit{\textbf{Trained on Original Instructions; Evaluated on Paraphrases}}} \\
\multicolumn{18}{@{}l}{\textit{Cross-Attn + Obj-Centric}} \\
\bfseries L1 & 99.5 & 100.0 & 99.0 & 86.5 & 57.0 & 100.0 & 100.0 & {---} & 15.0 & {---} & 60.5 & 94.0 & {---} & {---} & 99.5 & 47.0 & 64.0 & 78.6 \\
\bfseries L2 & 97.5 & 100.0 & 99.5 & 85.5 & 52.5 & 100.0 & 100.0 & {---} & 14.5 & {---} & 59.0 & 96.5 & {---} & {---} & 97.5 & 49.5 & 57.0 & 77.6 \\
\bfseries L3 & 92.0 & 96.5 & 99.5 & 87.0 & 58.5 & 99.0 & 99.0 & {---} & 12.0 & {---} & 51.0 & {---} & {---} & {---} & 97.0 & 39.5 & 6.5 & 69.8 \\
\bfseries L4 & {---} & {---} & {---} & {---} & {---} & {---} & {---} & 90.5 & {---} & 0.0 & {---} & {---} & 0.0 & 98.0 & {---} & {---} & {---} & 47.1 \\
\multicolumn{18}{@{}l}{\textit{Cross-Attn + Patches}} \\
\bfseries L1 & 88.5 & 72.5 & 96.0 & 11.5 & 0.5 & 66.5 & 95.0 & {---} & 13.5 & {---} & 91.5 & 92.5 & {---} & {---} & 76.5 & 86.5 & 3.0 & 61.1 \\
\bfseries L2 & 81.5 & 52.0 & 93.5 & 8.0 & 2.0 & 66.0 & 93.5 & {---} & 12.5 & {---} & 94.5 & 89.5 & {---} & {---} & 68.5 & 93.5 & 6.0 & 58.5 \\
\bfseries L3 & 57.0 & 64.0 & 81.0 & 8.5 & 2.0 & 64.0 & 65.0 & {---} & 14.5 & {---} & 90.0 & {---} & {---} & {---} & 51.0 & 45.5 & 1.0 & 45.3 \\
\bfseries L4 & {---} & {---} & {---} & {---} & {---} & {---} & {---} & 27.5 & {---} & 1.5 & {---} & {---} & 0.0 & 38.0 & {---} & {---} & {---} & 16.8 \\
\multicolumn{18}{@{}l}{\textit{Concatenate + Obj-Centric}} \\
\bfseries L1 & 100.0 & 100.0 & 96.5 & 73.5 & 4.0 & 100.0 & 100.0 & {---} & 16.5 & {---} & 80.0 & 89.5 & {---} & {---} & 88.5 & 42.0 & 38.5 & 71.5 \\
\bfseries L2 & 99.5 & 99.0 & 96.5 & 79.0 & 10.0 & 99.5 & 100.0 & {---} & 18.0 & {---} & 79.0 & 94.5 & {---} & {---} & 83.5 & 46.5 & 33.0 & 72.2 \\
\bfseries L3 & 87.0 & 81.5 & 97.0 & 76.0 & 6.0 & 85.0 & 94.5 & {---} & 17.5 & {---} & 71.0 & {---} & {---} & {---} & 79.0 & 47.5 & 10.0 & 62.7 \\
\bfseries L4 & {---} & {---} & {---} & {---} & {---} & {---} & {---} & 90.5 & {---} & 0.5 & {---} & {---} & 0.5 & 80.5 & {---} & {---} & {---} & 43.0 \\
\multicolumn{18}{@{}l}{\textit{Concatenate + Patches}} \\
\bfseries L1 & 95.0 & 80.5 & 47.5 & 11.5 & 3.0 & 76.0 & 93.5 & {---} & 34.0 & {---} & 91.5 & 94.0 & {---} & {---} & 78.5 & 85.5 & 7.0 & 61.3 \\
\bfseries L2 & 90.0 & 66.5 & 42.5 & 12.5 & 4.0 & 74.5 & 89.5 & {---} & 28.5 & {---} & 86.5 & 89.0 & {---} & {---} & 77.0 & 77.5 & 3.0 & 57.0 \\
\bfseries L3 & 70.5 & 66.5 & 45.0 & 13.0 & 1.0 & 69.0 & 59.0 & {---} & 33.0 & {---} & 89.0 & {---} & {---} & {---} & 47.0 & 56.0 & 3.0 & 46.0 \\
\bfseries L4 & {---} & {---} & {---} & {---} & {---} & {---} & {---} & 29.0 & {---} & 17.0 & {---} & {---} & 0.0 & 36.0 & {---} & {---} & {---} & 20.5 \\
\bottomrule
\end{tabular}
\caption{Per-task average success rate when evaluating performance on either \textbf{original instructions or paraphrases} during inference, corresponding to \crefsubtable{tab:perf-paraphrases}{a} and \crefsubtable{tab:perf-paraphrases}{b} respectively. All models are trained \textbf{on original instructions.}}
\label{tab:trained-on-orig-full}
\end{table*}

\begin{table*}[tbhp]
\centering
\scriptsize
\sisetup{mode=match,tight-spacing=true,table-format=2.1,table-number-alignment=center}
\renewcommand{\arraystretch}{1.2}
\setlength{\tabcolsep}{4.5pt}
\begin{tabular}{@{} l *{18}{S} @{}}
\toprule
& {T1} & {T2} & {T3} & {T4} & {T5} & {T6} & {T7} & {T8} & {T9} & {T10} & {T11} & {T12} & {T13} & {T14} & {T15} & {T16} & {T17} & {Avg.} \\
\midrule
\multicolumn{18}{@{}l}{\textit{\textbf{Trained on Paraphrases; Evaluated on the Original Instructions}}} \\
\multicolumn{18}{@{}l}{\textit{Cross-Attn + Obj-Centric}} \\
\bfseries L1 & 99.5 & 100.0 & 99.5 & 98.5 & 62.5 & 100.0 & 100.0 & {---} & 11.5 & {---} & 92.0 & 97.5 & {---} & {---} & 99.0 & 43.5 & 72.0 & 82.7 \\
\bfseries L2 & 99.0 & 100.0 & 99.5 & 98.0 & 55.5 & 100.0 & 100.0 & {---} & 13.5 & {---} & 91.5 & 92.5 & {---} & {---} & 97.0 & 48.0 & 69.0 & 81.8 \\
\bfseries L3 & 99.0 & 99.0 & 99.5 & 99.0 & 68.5 & 99.0 & 99.0 & {---} & 15.5 & {---} & 93.0 & {---} & {---} & {---} & 99.0 & 48.5 & 10.0 & 77.4 \\
\bfseries L4 & {---} & {---} & {---} & {---} & {---} & {---} & {---} & 93.0 & {---} & 0.5 & {---} & {---} & 0.0 & 98.5 & {---} & {---} & {---} & 48.0 \\
\multicolumn{18}{@{}l}{\textit{Cross-Attn + Patches}} \\
\bfseries L1 & 92.0 & 77.0 & 96.0 & 12.5 & 0.5 & 83.0 & 97.0 & {---} & 16.5 & {---} & 93.0 & 93.0 & {---} & {---} & 74.5 & 92.0 & 3.5 & 63.9 \\
\bfseries L2 & 90.0 & 66.5 & 97.0 & 9.5 & 1.0 & 93.5 & 94.5 & {---} & 13.0 & {---} & 93.0 & 87.5 & {---} & {---} & 79.0 & 91.0 & 3.5 & 63.0 \\
\bfseries L3 & 66.5 & 65.0 & 78.0 & 12.5 & 0.5 & 88.0 & 58.5 & {---} & 10.5 & {---} & 89.5 & {---} & {---} & {---} & 60.5 & 61.5 & 2.5 & 49.5 \\
\bfseries L4 & {---} & {---} & {---} & {---} & {---} & {---} & {---} & 45.5 & {---} & 0.5 & {---} & {---} & 0.0 & 35.5 & {---} & {---} & {---} & 20.4 \\
\multicolumn{18}{@{}l}{\textit{Concatenate + Obj-Centric}} \\
\bfseries L1 & 100.0 & 100.0 & 99.0 & 99.0 & 18.0 & 100.0 & 100.0 & {---} & 13.0 & {---} & 93.0 & 98.0 & {---} & {---} & 96.5 & 51.0 & 77.5 & 80.4 \\
\bfseries L2 & 100.0 & 100.0 & 100.0 & 98.0 & 8.5 & 100.0 & 100.0 & {---} & 13.5 & {---} & 92.0 & 92.5 & {---} & {---} & 92.0 & 46.5 & 73.5 & 78.2 \\
\bfseries L3 & 98.0 & 94.0 & 100.0 & 99.5 & 14.0 & 94.5 & 93.0 & {---} & 12.5 & {---} & 96.5 & {---} & {---} & {---} & 98.0 & 42.0 & 56.0 & 74.8 \\
\bfseries L4 & {---} & {---} & {---} & {---} & {---} & {---} & {---} & 96.5 & {---} & 2.5 & {---} & {---} & 0.0 & 97.0 & {---} & {---} & {---} & 49.0 \\
\multicolumn{18}{@{}l}{\textit{Concatenate + Patches}} \\
\bfseries L1 & 97.0 & 81.5 & 98.5 & 13.0 & 1.5 & 94.5 & 96.0 & {---} & 33.0 & {---} & 89.0 & 92.5 & {---} & {---} & 73.5 & 98.0 & 4.0 & 67.1 \\
\bfseries L2 & 89.5 & 69.5 & 96.0 & 11.5 & 2.0 & 93.5 & 87.5 & {---} & 23.0 & {---} & 91.5 & 92.5 & {---} & {---} & 67.0 & 92.5 & 1.0 & 62.8 \\
\bfseries L3 & 65.0 & 74.5 & 87.5 & 14.0 & 3.0 & 88.5 & 60.5 & {---} & 29.0 & {---} & 85.5 & {---} & {---} & {---} & 50.5 & 65.0 & 1.5 & 52.0 \\
\bfseries L4 & {---} & {---} & {---} & {---} & {---} & {---} & {---} & 38.5 & {---} & 10.0 & {---} & {---} & 0.0 & 30.5 & {---} & {---} & {---} & 19.8 \\
\midrule[0.2pt]
\addlinespace[2pt]
\multicolumn{18}{@{}l}{\textit{\textbf{Trained and Evaluated on Paraphrases}}} \\
\multicolumn{18}{@{}l}{\textit{Cross-Attn + Obj-Centric}} \\
\bfseries L1 & 98.5 & 100.0 & 97.5 & 85.5 & 56.0 & 99.5 & 100.0 & {---} & 12.0 & {---} & 55.5 & 96.0 & {---} & {---} & 99.0 & 44.0 & 62.5 & 77.4 \\
\bfseries L2 & 98.0 & 99.5 & 99.5 & 90.5 & 63.5 & 99.0 & 100.0 & {---} & 10.5 & {---} & 55.0 & 92.0 & {---} & {---} & 98.0 & 45.0 & 57.5 & 77.5 \\
\bfseries L3 & 91.5 & 98.0 & 98.5 & 88.0 & 64.0 & 97.5 & 98.5 & {---} & 14.0 & {---} & 48.0 & {---} & {---} & {---} & 95.5 & 53.0 & 2.5 & 70.8 \\
\bfseries L4 & {---} & {---} & {---} & {---} & {---} & {---} & {---} & 94.0 & {---} & 2.5 & {---} & {---} & 0.0 & 98.0 & {---} & {---} & {---} & 48.6 \\
\multicolumn{18}{@{}l}{\textit{Cross-Attn + Patches}} \\
\bfseries L1 & 92.0 & 66.5 & 97.5 & 8.0 & 0.5 & 73.0 & 95.5 & {---} & 15.0 & {---} & 91.5 & 92.0 & {---} & {---} & 82.0 & 92.0 & 3.5 & 62.2 \\
\bfseries L2 & 92.5 & 53.5 & 96.0 & 13.5 & 0.5 & 72.5 & 94.0 & {---} & 17.5 & {---} & 93.0 & 90.0 & {---} & {---} & 75.0 & 93.5 & 1.5 & 61.0 \\
\bfseries L3 & 63.5 & 50.0 & 80.0 & 1.5 & 0.5 & 69.0 & 62.0 & {---} & 16.0 & {---} & 87.5 & {---} & {---} & {---} & 64.5 & 50.5 & 3.0 & 45.7 \\
\bfseries L4 & {---} & {---} & {---} & {---} & {---} & {---} & {---} & 24.5 & {---} & 3.5 & {---} & {---} & 0.0 & 36.5 & {---} & {---} & {---} & 16.1 \\
\multicolumn{18}{@{}l}{\textit{Concatenate + Obj-Centric}} \\
\bfseries L1 & 100.0 & 99.5 & 99.5 & 56.5 & 2.5 & 100.0 & 100.0 & {---} & 15.5 & {---} & 69.0 & 95.5 & {---} & {---} & 94.0 & 45.5 & 17.5 & 68.8 \\
\bfseries L2 & 100.0 & 97.5 & 99.5 & 54.5 & 7.0 & 100.0 & 99.5 & {---} & 15.5 & {---} & 60.0 & 94.0 & {---} & {---} & 89.5 & 43.5 & 13.0 & 67.2 \\
\bfseries L3 & 88.5 & 84.0 & 100.0 & 52.5 & 5.0 & 94.5 & 94.0 & {---} & 14.5 & {---} & 54.0 & {---} & {---} & {---} & 86.0 & 39.0 & 3.5 & 59.6 \\
\bfseries L4 & {---} & {---} & {---} & {---} & {---} & {---} & {---} & 97.5 & {---} & 2.0 & {---} & {---} & 0.0 & 84.5 & {---} & {---} & {---} & 46.0 \\
\multicolumn{18}{@{}l}{\textit{Concatenate + Patches}} \\
\bfseries L1 & 100.0 & 99.0 & 99.5 & 53.5 & 5.5 & 99.5 & 100.0 & {---} & 11.0 & {---} & 61.0 & 94.5 & {---} & {---} & 92.5 & 42.5 & 15.5 & 67.2 \\
\bfseries L2 & 99.5 & 99.5 & 100.0 & 62.0 & 6.5 & 100.0 & 100.0 & {---} & 14.0 & {---} & 60.5 & 95.0 & {---} & {---} & 90.0 & 43.0 & 12.0 & 67.8 \\
\bfseries L3 & 93.0 & 79.5 & 99.5 & 59.5 & 4.0 & 95.5 & 94.5 & {---} & 13.0 & {---} & 57.5 & {---} & {---} & {---} & 88.5 & 38.0 & 4.0 & 60.5 \\
\bfseries L4 & {---} & {---} & {---} & {---} & {---} & {---} & {---} & 97.0 & {---} & 4.0 & {---} & {---} & 0.0 & 86.5 & {---} & {---} & {---} & 46.9 \\
\bottomrule
\end{tabular}
\caption{Per-task average success rate when evaluating performance on either \textbf{original instructions or paraphrases} during inference, corresponding to \cref{tab:perf-paraphrases}\hyperref[tab:perf-paraphrases]{c} and \cref{tab:perf-paraphrases}\hyperref[tab:perf-paraphrases]{d} respectively. All models are trained \textbf{on paraphrased instructions.}}
\label{tab:trained-on-para-full}
\end{table*}

\begin{table*}[tbhp]
\centering
\scriptsize
\sisetup{table-format=2.1}
\renewcommand{\arraystretch}{1.2}
\setlength{\tabcolsep}{4.5pt}
\begin{tabular}{@{} l *{18}{S} @{}}
\toprule
& {T1} & {T2} & {T3} & {T4} & {T5} & {T6} & {T7} & {T8} & {T9} & {T10} & {T11} & {T12} & {T13} & {T14} & {T15} & {T16} & {T17} & {Avg.} \\
\midrule
\multicolumn{18}{@{}l}{\textit{\textbf{With Visual Referents*}}} \\
\multicolumn{18}{@{}l}{\textit{Cross-Attn + Obj-Centric}} \\
\bfseries L1 & 99.5 & {---} & 99.5 & {---} & {---} & {---} & 100.0 & {---} & {---} & {---} & {---} & 97.5 & {---} & {---} & 99.0 & 43.5 & 72.0 & 87.3 \\
\bfseries L2 & 99.0 & {---} & 99.5 & {---} & {---} & {---} & 100.0 & {---} & {---} & {---} & {---} & 92.5 & {---} & {---} & 97.0 & 48.0 & 69.0 & 86.4 \\
\bfseries L3 & 99.0 & {---} & 99.5 & {---} & {---} & {---} & 99.0 & {---} & {---} & {---} & {---} & {---} & {---} & {---} & 99.0 & 48.5 & 10.0 & 75.8 \\
\bfseries L4 & {---} & {---} & {---} & {---} & {---} & {---} & {---} & {---} & {---} & {---} & {---} & {---} & 0.0 & 98.5 & {---} & {---} & {---} & 49.2 \\
\multicolumn{18}{@{}l}{\textit{Cross-attention + Patches}}\\
\bfseries L1 & 92.0 & {---} & 96.0 & {---} & {---} & {---} & 97.0 & {---} & {---} & {---} & {---} & 93.0 & {---} & {---} & 74.5 & 92.0 & 3.5 & 78.3 \\
\bfseries L2 & 90.0 & {---} & 97.0 & {---} & {---} & {---} & 94.5 & {---} & {---} & {---} & {---} & 87.5 & {---} & {---} & 79.0 & 91.0 & 3.5 & 77.5 \\
\bfseries L3 & 66.5 & {---} & 78.0 & {---} & {---} & {---} & 58.5 & {---} & {---} & {---} & {---} & {---} & {---} & {---} & 60.5 & 61.5 & 2.5 & 54.6 \\
\bfseries L4 & {---} & {---} & {---} & {---} & {---} & {---} & {---} & {---} & {---} & {---} & {---} & {---} & 0.0 & 35.5 & {---} & {---} & {---} & 17.8 \\
\multicolumn{18}{@{}l}{\textit{Concatenate + Obj-Centric}} \\
\bfseries L1 & 100.0 & {---} & 99.0 & {---} & {---} & {---} & 100.0 & {---} & {---} & {---} & {---} & 98.0 & {---} & {---} & 96.5 & 51.0 & 77.5 & 88.9 \\
\bfseries L2 & 100.0 & {---} & 100.0 & {---} & {---} & {---} & 100.0 & {---} & {---} & {---} & {---} & 92.5 & {---} & {---} & 92.0 & 46.5 & 73.5 & 86.4 \\
\bfseries L3 & 98.0 & {---} & 100.0 & {---} & {---} & {---} & 93.0 & {---} & {---} & {---} & {---} & {---} & {---} & {---} & 98.0 & 42.0 & 56.0 & 81.2 \\
\bfseries L4 & {---} & {---} & {---} & {---} & {---} & {---} & {---} & {---} & {---} & {---} & {---} & {---} & 0.0 & 97.0 & {---} & {---} & {---} & 48.5 \\
\multicolumn{18}{@{}l}{\textit{Concatenate + Patches}} \\
\bfseries L1 & 97.0 & {---} & 98.5 & {---} & {---} & {---} & 96.0 & {---} & {---} & {---} & {---} & 92.5 & {---} & {---} & 73.5 & 98.0 & 4.0 & 79.9 \\
\bfseries L2 & 89.5 & {---} & 96.0 & {---} & {---} & {---} & 87.5 & {---} & {---} & {---} & {---} & 92.5 & {---} & {---} & 67.0 & 92.5 & 1.0 & 75.1 \\
\bfseries L3 & 65.0 & {---} & 87.5 & {---} & {---} & {---} & 60.5 & {---} & {---} & {---} & {---} & {---} & {---} & {---} & 50.5 & 65.0 & 1.5 & 55.0 \\
\bfseries L4 & {---} & {---} & {---} & {---} & {---} & {---} & {---} & {---} & {---} & {---} & {---} & {---} & 0.0 & 30.5 & {---} & {---} & {---} & 15.2 \\
\midrule[0.2pt]
\addlinespace[2pt]
\multicolumn{18}{@{}l}{\textit{\textbf{Replace Visual Referents with Descriptors*}}} \\
\multicolumn{18}{@{}l}{\textit{Cross-Attn + Obj-Centric}} \\
\bfseries L1 & 100.0 & {---} & 100.0 & {---} & {---} & {---} & 100.0 & {---} & {---} & {---} & {---} & 97.5 & {---} & {---} & 97.5 & 47.5 & 72.5 & 87.9 \\
\bfseries L2 & 100.0 & {---} & 99.0 & {---} & {---} & {---} & 99.5 & {---} & {---} & {---} & {---} & 94.5 & {---} & {---} & 98.5 & 47.0 & 72.0 & 87.2 \\
\bfseries L3 & 99.0 & {---} & 99.5 & {---} & {---} & {---} & 96.5 & {---} & {---} & {---} & {---} & {---} & {---} & {---} & 96.5 & 48.5 & 0.0 & 73.3 \\
\bfseries L4 & {---} & {---} & {---} & {---} & {---} & {---} & {---} & {---} & {---} & {---} & {---} & {---} & 0.0 & 98.0 & {---} & {---} & {---} & 49.0 \\
\multicolumn{18}{@{}l}{\textit{Cross-attention + Patches}}\\
\bfseries L1 & 69.5 & {---} & 41.5 & {---} & {---} & {---} & 64.0 & {---} & {---} & {---} & {---} & 72.5 & {---} & {---} & 44.0 & 33.0 & 3.0 & 46.8 \\
\bfseries L2 & 61.5 & {---} & 31.0 & {---} & {---} & {---} & 64.5 & {---} & {---} & {---} & {---} & 81.0 & {---} & {---} & 38.5 & 35.5 & 1.0 & 44.7 \\
\bfseries L3 & 56.0 & {---} & 42.5 & {---} & {---} & {---} & 50.0 & {---} & {---} & {---} & {---} & {---} & {---} & {---} & 42.0 & 37.0 & 1.5 & 38.2 \\
\bfseries L4 & {---} & {---} & {---} & {---} & {---} & {---} & {---} & {---} & {---} & {---} & {---} & {---} & 0.0 & 51.0 & {---} & {---} & {---} & 25.5 \\
\multicolumn{18}{@{}l}{\textit{Concatenate + Obj-Centric}} \\
\bfseries L1 & 100.0 & {---} & 99.5 & {---} & {---} & {---} & 100.0 & {---} & {---} & {---} & {---} & 97.0 & {---} & {---} & 98.5 & 44.0 & 17.0 & 79.4 \\
\bfseries L2 & 99.0 & {---} & 100.0 & {---} & {---} & {---} & 99.5 & {---} & {---} & {---} & {---} & 93.0 & {---} & {---} & 92.5 & 47.0 & 16.0 & 78.1 \\
\bfseries L3 & 94.5 & {---} & 99.5 & {---} & {---} & {---} & 94.5 & {---} & {---} & {---} & {---} & {---} & {---} & {---} & 89.0 & 41.0 & 1.5 & 70.0 \\
\bfseries L4 & {---} & {---} & {---} & {---} & {---} & {---} & {---} & {---} & {---} & {---} & {---} & {---} & 0.0 & 77.0 & {---} & {---} & {---} & 38.5 \\
\multicolumn{18}{@{}l}{\textit{Concatenate + Patches}} \\
\bfseries L1 & 82.5 & {---} & 92.5 & {---} & {---} & {---} & 61.5 & {---} & {---} & {---} & {---} & 85.5 & {---} & {---} & 25.5 & 45.0 & 2.5 & 56.4 \\
\bfseries L2 & 65.5 & {---} & 80.5 & {---} & {---} & {---} & 58.5 & {---} & {---} & {---} & {---} & 81.0 & {---} & {---} & 29.0 & 39.0 & 0.5 & 50.6 \\
\bfseries L3 & 72.0 & {---} & 88.0 & {---} & {---} & {---} & 55.5 & {---} & {---} & {---} & {---} & {---} & {---} & {---} & 55.5 & 41.0 & 0.0 & 52.0 \\
\bfseries L4 & {---} & {---} & {---} & {---} & {---} & {---} & {---} & {---} & {---} & {---} & {---} & {---} & 0.0 & 51.5 & {---} & {---} & {---} & 25.8 \\
\bottomrule
\end{tabular}
\caption{Per-task average success rate when evaluating performance either with \textbf{visual referents (top) and when visual referents are replaced with descriptors (bottom),} corresponding to \cref{tab:perf-obj-text-para}. All \textbf{models are trained on paraphrases}. As mentioned in \cref{app:textual-details}, not all instructions contain visual referents that can be directly substituted for language. For ease of comparison, only tasks with instructions that support substitutions are included in the top section, with the average for the level calculated with only these tasks.}
\label{tab:full-obj-as-text}
\end{table*}

\begin{table*}[tbhp]
\centering
\scriptsize
\sisetup{table-format=2.1}
\renewcommand{\arraystretch}{1.2}
\setlength{\tabcolsep}{4.5pt}
\begin{tabular}{@{} l *{18}{S} @{}}
\toprule
& {T1} & {T2} & {T3} & {T4} & {T5} & {T6} & {T7} & {T8} & {T9} & {T10} & {T11} & {T12} & {T13} & {T14} & {T15} & {T16} & {T17} & {Avg.} \\
\midrule
\multicolumn{18}{@{}l}{\textit{\textbf{Gobbledygook Tokens}}} \\
\multicolumn{18}{@{}l}{\textit{Cross-Attn + Obj-Centric}} \\
\bfseries L1 & 89.5 & 74.5 & 5.5 & 73.5 & 8.5 & 92.5 & 90.0 & {---} & 1.0 & {---} & 92.5 & 94.0 & {---} & {---} & 29.0 & 19.5 & 67.5 & 56.7 \\
\bfseries L2 & 89.5 & 67.5 & 5.0 & 71.5 & 12.0 & 86.5 & 89.5 & {---} & 0.5 & {---} & 92.0 & 93.5 & {---} & {---} & 25.0 & 19.0 & 56.5 & 54.5 \\
\bfseries L3 & 63.0 & 48.0 & 7.5 & 75.0 & 11.5 & 59.5 & 59.5 & {---} & 0.0 & {---} & 93.5 & {---} & {---} & {---} & 10.5 & 11.0 & 0.5 & 36.6 \\
\bfseries L4 & {---} & {---} & {---} & {---} & {---} & {---} & {---} & 72.5 & {---} & 0.0 & {---} & {---} & 0.0 & 19.0 & {---} & {---} & {---} & 22.9 \\
\multicolumn{18}{@{}l}{\textit{Cross-Attn + Patches}} \\
\bfseries L1 & 88.5 & 32.5 & 23.0 & 7.0 & 0.5 & 53.5 & 90.0 & {---} & 13.0 & {---} & 92.0 & 88.5 & {---} & {---} & 66.5 & 28.0 & 4.5 & 45.2 \\
\bfseries L2 & 87.5 & 33.0 & 24.5 & 8.5 & 0.5 & 53.5 & 92.5 & {---} & 12.0 & {---} & 94.0 & 91.5 & {---} & {---} & 67.0 & 30.0 & 2.0 & 45.9 \\
\bfseries L3 & 63.0 & 43.5 & 17.5 & 7.0 & 0.5 & 48.5 & 61.5 & {---} & 13.5 & {---} & 88.0 & {---} & {---} & {---} & 49.5 & 14.5 & 1.5 & 34.0 \\
\bfseries L4 & {---} & {---} & {---} & {---} & {---} & {---} & {---} & 22.5 & {---} & 1.5 & {---} & {---} & 0.0 & 36.5 & {---} & {---} & {---} & 15.1 \\
\multicolumn{18}{@{}l}{\textit{Concatenate + Obj-Centric}} \\
\bfseries L1 & 100.0 & 97.0 & 15.0 & 81.5 & 9.5 & 100.0 & 99.5 & {---} & 0.5 & {---} & 93.0 & 95.0 & {---} & {---} & 17.5 & 21.5 & 7.5 & 56.7 \\
\bfseries L2 & 99.0 & 95.0 & 9.5 & 86.0 & 5.5 & 99.0 & 99.5 & {---} & 1.0 & {---} & 94.0 & 96.5 & {---} & {---} & 12.5 & 16.0 & 5.5 & 55.3 \\
\bfseries L3 & 87.0 & 68.0 & 10.0 & 93.5 & 7.0 & 86.0 & 78.0 & {---} & 0.5 & {---} & 90.5 & {---} & {---} & {---} & 15.5 & 11.5 & 2.0 & 45.8 \\
\bfseries L4 & {---} & {---} & {---} & {---} & {---} & {---} & {---} & 89.5 & {---} & 4.5 & {---} & {---} & 0.0 & 11.5 & {---} & {---} & {---} & 26.4 \\
\multicolumn{18}{@{}l}{\textit{Concatenate + Patches}} \\
\bfseries L1 & 89.5 & 56.5 & 16.0 & 11.5 & 0.5 & 52.0 & 85.0 & {---} & 11.5 & {---} & 76.0 & 92.0 & {---} & {---} & 66.5 & 35.5 & 4.0 & 45.9 \\
\bfseries L2 & 89.5 & 46.5 & 15.0 & 7.5 & 2.0 & 53.0 & 85.0 & {---} & 12.0 & {---} & 78.5 & 87.0 & {---} & {---} & 64.0 & 33.0 & 3.0 & 44.3 \\
\bfseries L3 & 66.5 & 53.0 & 21.0 & 8.5 & 1.5 & 47.5 & 45.5 & {---} & 11.0 & {---} & 80.5 & {---} & {---} & {---} & 37.5 & 21.0 & 1.0 & 32.9 \\
\bfseries L4 & {---} & {---} & {---} & {---} & {---} & {---} & {---} & 25.5 & {---} & 8.0 & {---} & {---} & 0.0 & 46.5 & {---} & {---} & {---} & 20.0 \\
\midrule[0.2pt]
\addlinespace[2pt]
\multicolumn{18}{@{}l}{\textit{\textbf{Gobbledygook Words}}} \\
\multicolumn{18}{@{}l}{\textit{Cross-Attn + Obj-Centric}} \\
\bfseries L1 & 95.0 & 98.5 & 9.5 & 17.0 & 4.5 & 98.0 & 99.5 & {---} & 0.5 & {---} & 56.0 & 90.5 & {---} & {---} & 58.5 & 17.0 & 16.5 & 50.8 \\
\bfseries L2 & 95.5 & 97.0 & 7.0 & 23.0 & 2.5 & 99.0 & 100.0 & {---} & 1.5 & {---} & 52.5 & 93.0 & {---} & {---} & 64.5 & 25.0 & 13.5 & 51.8 \\
\bfseries L3 & 74.5 & 86.5 & 6.0 & 23.5 & 3.0 & 93.5 & 87.5 & {---} & 0.0 & {---} & 56.5 & {---} & {---} & {---} & 37.0 & 11.0 & 0.0 & 39.9 \\
\bfseries L4 & {---} & {---} & {---} & {---} & {---} & {---} & {---} & 85.5 & {---} & 0.0 & {---} & {---} & 0.0 & 49.5 & {---} & {---} & {---} & 33.8 \\
\multicolumn{18}{@{}l}{\textit{Cross-Attn + Patches}} \\
\bfseries L1 & 95.5 & 47.5 & 21.5 & 8.0 & 0.5 & 50.5 & 92.5 & {---} & 11.0 & {---} & 89.0 & 87.5 & {---} & {---} & 72.5 & 28.0 & 3.5 & 46.7 \\
\bfseries L2 & 91.5 & 55.5 & 21.5 & 6.5 & 1.0 & 63.0 & 93.0 & {---} & 12.5 & {---} & 91.5 & 91.5 & {---} & {---} & 75.0 & 24.5 & 2.5 & 48.4 \\
\bfseries L3 & 61.0 & 52.0 & 13.5 & 7.0 & 0.5 & 52.5 & 57.0 & {---} & 12.5 & {---} & 81.5 & {---} & {---} & {---} & 47.5 & 22.0 & 0.0 & 33.9 \\
\bfseries L4 & {---} & {---} & {---} & {---} & {---} & {---} & {---} & 26.5 & {---} & 3.5 & {---} & {---} & 0.0 & 44.5 & {---} & {---} & {---} & 18.6 \\
\multicolumn{18}{@{}l}{\textit{Concatenate + Obj-Centric}} \\
\bfseries L1 & 99.5 & 99.0 & 15.5 & 10.0 & 0.0 & 100.0 & 100.0 & {---} & 0.5 & {---} & 46.0 & 90.5 & {---} & {---} & 4.0 & 13.0 & 5.0 & 44.8 \\
\bfseries L2 & 99.5 & 94.5 & 17.5 & 11.0 & 1.0 & 98.5 & 98.5 & {---} & 1.5 & {---} & 47.0 & 87.5 & {---} & {---} & 5.0 & 12.0 & 4.5 & 44.5 \\
\bfseries L3 & 84.0 & 74.5 & 19.5 & 10.0 & 0.0 & 88.5 & 84.0 & {---} & 2.5 & {---} & 51.5 & {---} & {---} & {---} & 7.0 & 3.0 & 0.0 & 35.4 \\
\bfseries L4 & {---} & {---} & {---} & {---} & {---} & {---} & {---} & 87.0 & {---} & 3.0 & {---} & {---} & 0.0 & 5.5 & {---} & {---} & {---} & 23.9 \\
\multicolumn{18}{@{}l}{\textit{Concatenate + Patches}} \\
\bfseries L1 & 91.0 & 60.5 & 15.0 & 7.0 & 1.0 & 48.5 & 86.0 & {---} & 10.0 & {---} & 70.5 & 91.0 & {---} & {---} & 69.0 & 23.5 & 2.5 & 44.3 \\
\bfseries L2 & 87.0 & 50.5 & 22.0 & 9.5 & 0.5 & 51.5 & 82.0 & {---} & 5.5 & {---} & 69.5 & 90.5 & {---} & {---} & 63.0 & 22.0 & 1.5 & 42.7 \\
\bfseries L3 & 60.5 & 51.0 & 17.0 & 6.0 & 1.5 & 46.0 & 56.5 & {---} & 6.0 & {---} & 70.0 & {---} & {---} & {---} & 40.5 & 16.0 & 0.5 & 31.0 \\
\bfseries L4 & {---} & {---} & {---} & {---} & {---} & {---} & {---} & 21.0 & {---} & 9.0 & {---} & {---} & 0.0 & 46.0 & {---} & {---} & {---} & 19.0 \\
\bottomrule
\end{tabular}
\caption{Per-task average success rate when evaluating performance with either \textbf{\GDGTokens (top) and \GDGWords (bottom).} All models are trained \textbf{on paraphrased instructions}. This table corresponds to \cref{tab:perf-gdg-on-para}.}
\label{tab:full-gdg}
\end{table*}

\begin{table*}[tbhp]
\centering
\scriptsize
\sisetup{table-format=2.1}
\renewcommand{\arraystretch}{1.2}
\setlength{\tabcolsep}{4.5pt}
\begin{tabular}{@{} l *{18}{S} @{}}
\toprule
& {T1} & {T2} & {T3} & {T4} & {T5} & {T6} & {T7} & {T8} & {T9} & {T10} & {T11} & {T12} & {T13} & {T14} & {T15} & {T16} & {T17} & {Avg.} \\
\midrule
\multicolumn{18}{@{}l}{\textit{\textbf{Mask Language Tokens}}} \\
\multicolumn{18}{@{}l}{\textit{Cross-Attn + Obj-Centric}} \\
\bfseries L1 & 74.5 & 49.0 & 0.5 & 2.0 & 0.0 & 76.0 & 71.0 & {---} & 0.0 & {---} & 23.5 & 96.0 & {---} & {---} & 20.5 & 15.5 & 44.0 & 36.3 \\
\bfseries L2 & 72.5 & 53.0 & 0.0 & 4.0 & 0.0 & 66.0 & 68.0 & {---} & 0.0 & {---} & 25.5 & 94.0 & {---} & {---} & 20.0 & 17.5 & 35.5 & 35.1 \\
\bfseries L3 & 45.0 & 39.5 & 0.5 & 3.5 & 0.0 & 51.0 & 48.0 & {---} & 0.0 & {---} & 30.0 & {---} & {---} & {---} & 6.0 & 5.5 & 0.0 & 19.1 \\
\bfseries L4 & {---} & {---} & {---} & {---} & {---} & {---} & {---} & 43.0 & {---} & 0.5 & {---} & {---} & 0.0 & 15.5 & {---} & {---} & {---} & 14.8 \\
\multicolumn{18}{@{}l}{\textit{Cross-Attn + Patches}} \\
\bfseries L1 & 58.5 & 35.5 & 8.0 & 0.0 & 0.0 & 46.5 & 56.0 & {---} & 6.0 & {---} & 18.0 & 87.0 & {---} & {---} & 6.5 & 17.0 & 3.5 & 26.3 \\
\bfseries L2 & 58.0 & 24.0 & 10.0 & 0.0 & 0.0 & 54.0 & 60.0 & {---} & 6.0 & {---} & 17.0 & 83.5 & {---} & {---} & 6.0 & 23.5 & 2.5 & 26.5 \\
\bfseries L3 & 50.5 & 23.5 & 14.0 & 0.0 & 0.0 & 43.5 & 50.0 & {---} & 11.5 & {---} & 21.5 & {---} & {---} & {---} & 21.0 & 11.0 & 1.5 & 20.7 \\
\bfseries L4 & {---} & {---} & {---} & {---} & {---} & {---} & {---} & 19.5 & {---} & 2.5 & {---} & {---} & 0.0 & 23.0 & {---} & {---} & {---} & 11.2 \\
\multicolumn{18}{@{}l}{\textit{Concatenate + Obj-Centric}} \\
\bfseries L1 & 97.0 & 95.5 & 8.0 & 0.0 & 0.0 & 99.5 & 99.0 & {---} & 0.0 & {---} & 5.0 & 92.5 & {---} & {---} & 2.5 & 7.0 & 0.5 & 39.0 \\
\bfseries L2 & 96.0 & 96.0 & 6.0 & 0.0 & 0.0 & 99.0 & 98.5 & {---} & 1.5 & {---} & 5.0 & 91.5 & {---} & {---} & 1.5 & 11.0 & 1.5 & 39.0 \\
\bfseries L3 & 80.0 & 79.0 & 5.0 & 0.0 & 0.0 & 87.0 & 79.5 & {---} & 1.5 & {---} & 3.0 & {---} & {---} & {---} & 1.0 & 9.0 & 0.0 & 28.8 \\
\bfseries L4 & {---} & {---} & {---} & {---} & {---} & {---} & {---} & 94.5 & {---} & 4.0 & {---} & {---} & 0.0 & 5.0 & {---} & {---} & {---} & 25.9 \\
\multicolumn{18}{@{}l}{\textit{Concatenate + Patches}} \\
\bfseries L1 & 71.5 & 58.5 & 7.0 & 0.0 & 0.0 & 56.0 & 60.5 & {---} & 1.5 & {---} & 9.0 & 78.5 & {---} & {---} & 33.0 & 17.5 & 0.0 & 30.2 \\
\bfseries L2 & 66.5 & 60.0 & 5.0 & 0.0 & 0.0 & 49.0 & 61.5 & {---} & 1.0 & {---} & 11.0 & 77.5 & {---} & {---} & 28.5 & 16.0 & 1.0 & 29.0 \\
\bfseries L3 & 60.0 & 55.5 & 8.5 & 0.0 & 0.0 & 57.0 & 57.0 & {---} & 1.0 & {---} & 9.0 & {---} & {---} & {---} & 37.0 & 9.5 & 0.0 & 24.5 \\
\bfseries L4 & {---} & {---} & {---} & {---} & {---} & {---} & {---} & 23.0 & {---} & 3.0 & {---} & {---} & 0.0 & 39.0 & {---} & {---} & {---} & 16.2 \\
\midrule[0.2pt]
\addlinespace[2pt]
\multicolumn{18}{@{}l}{\textit{\textbf{Mask Visual Referents}}} \\
\multicolumn{18}{@{}l}{\textit{Cross-Attn + Obj-Centric}} \\
\bfseries L1 & 100.0 & 100.0 & 100.0 & 0.0 & 0.0 & 100.0 & 100.0 & {---} & 13.5 & {---} & 6.5 & 95.0 & {---} & {---} & 94.5 & 48.5 & 69.0 & 63.6 \\
\bfseries L2 & 100.0 & 99.5 & 100.0 & 0.0 & 0.0 & 98.0 & 99.5 & {---} & 12.0 & {---} & 6.0 & 94.0 & {---} & {---} & 88.0 & 49.5 & 67.0 & 62.6 \\
\bfseries L3 & 99.0 & 100.0 & 100.0 & 0.0 & 0.0 & 98.5 & 99.0 & {---} & 9.0 & {---} & 4.0 & {---} & {---} & {---} & 98.0 & 50.0 & 19.5 & 56.4 \\
\bfseries L4 & {---} & {---} & {---} & {---} & {---} & {---} & {---} & 92.0 & {---} & 0.5 & {---} & {---} & 0.0 & 99.0 & {---} & {---} & {---} & 47.9 \\
\multicolumn{18}{@{}l}{\textit{Cross-Attn + Patches}} \\
\bfseries L1 & 96.0 & 82.0 & 98.5 & 10.0 & 0.5 & 89.0 & 95.5 & {---} & 16.0 & {---} & 91.5 & 93.5 & {---} & {---} & 72.5 & 94.5 & 2.5 & 64.8 \\
\bfseries L2 & 92.0 & 64.5 & 97.5 & 12.5 & 1.5 & 92.5 & 95.0 & {---} & 11.5 & {---} & 89.5 & 93.5 & {---} & {---} & 70.0 & 94.0 & 5.0 & 63.0 \\
\bfseries L3 & 64.0 & 70.0 & 82.0 & 9.5 & 1.5 & 83.5 & 58.5 & {---} & 14.5 & {---} & 90.0 & {---} & {---} & {---} & 63.0 & 57.0 & 2.0 & 49.6 \\
\bfseries L4 & {---} & {---} & {---} & {---} & {---} & {---} & {---} & 41.5 & {---} & 5.5 & {---} & {---} & 0.0 & 35.5 & {---} & {---} & {---} & 20.6 \\
\multicolumn{18}{@{}l}{\textit{Concatenate + Obj-Centric}} \\
\bfseries L1 & 100.0 & 100.0 & 99.5 & 0.0 & 0.0 & 100.0 & 100.0 & {---} & 15.5 & {---} & 1.0 & 98.5 & {---} & {---} & 96.5 & 47.0 & 19.0 & 59.8 \\
\bfseries L2 & 100.0 & 100.0 & 99.5 & 0.0 & 0.0 & 100.0 & 100.0 & {---} & 15.5 & {---} & 0.5 & 97.0 & {---} & {---} & 91.0 & 43.5 & 19.0 & 58.9 \\
\bfseries L3 & 94.5 & 92.5 & 100.0 & 0.0 & 0.0 & 95.5 & 94.5 & {---} & 14.5 & {---} & 3.0 & {---} & {---} & {---} & 94.0 & 47.0 & 2.5 & 53.2 \\
\bfseries L4 & {---} & {---} & {---} & {---} & {---} & {---} & {---} & 95.5 & {---} & 2.5 & {---} & {---} & 0.0 & 93.0 & {---} & {---} & {---} & 47.8 \\
\multicolumn{18}{@{}l}{\textit{Concatenate + Patches}} \\
\bfseries L1 & 97.5 & 86.0 & 98.5 & 13.5 & 5.5 & 88.5 & 94.0 & {---} & 35.0 & {---} & 90.0 & 92.5 & {---} & {---} & 71.5 & 97.0 & 3.0 & 67.1 \\
\bfseries L2 & 89.5 & 67.0 & 95.5 & 15.5 & 2.0 & 93.5 & 90.0 & {---} & 29.0 & {---} & 89.0 & 90.5 & {---} & {---} & 71.5 & 92.5 & 2.0 & 63.7 \\
\bfseries L3 & 69.0 & 73.5 & 89.0 & 15.5 & 2.0 & 85.5 & 63.5 & {---} & 30.5 & {---} & 86.5 & {---} & {---} & {---} & 52.5 & 65.5 & 0.5 & 52.8 \\
\bfseries L4 & {---} & {---} & {---} & {---} & {---} & {---} & {---} & 44.0 & {---} & 10.0 & {---} & {---} & 0.0 & 38.0 & {---} & {---} & {---} & 23.0 \\
\bottomrule
\end{tabular}
\caption{Per-task average success rate when \textbf{evaluating performance after masking language tokens (top) or visual referents (bottom)} within each multimodal instruction. All models are trained \textbf{on paraphrased instructions}. This table corresponds to \cref{tab:perf-mask-modalities-on-para}.}
\label{tab:mask-modality-full}
\end{table*}

\begin{table*}[tbhp]
\centering
\scriptsize
\sisetup{table-format=2.1}
\renewcommand{\arraystretch}{1.2}
\setlength{\tabcolsep}{4.5pt}
\begin{tabular}{@{} l *{18}{S} @{}}
\toprule
& {T1} & {T2} & {T3} & {T4} & {T5} & {T6} & {T7} & {T8} & {T9} & {T10} & {T11} & {T12} & {T13} & {T14} & {T15} & {T16} & {T17} & {Avg.} \\
\midrule
\multicolumn{18}{@{}l}{\textit{Cross-Attn + Obj-Centric}} \\
\bfseries L1 & 99.0 & 99.0 & 99.0 & 83.5 & 0.0 & 97.5 & 98.0 & {---} & 11.5 & {---} & 92.0 & 97.5 & {---} & {---} & 96.0 & 41.5 & 0.0 & 70.3 \\
\bfseries L2 & 97.5 & 98.0 & 99.5 & 78.0 & 0.0 & 98.0 & 99.0 & {---} & 13.5 & {---} & 91.0 & 91.5 & {---} & {---} & 94.5 & 46.0 & 0.0 & 69.7 \\
\bfseries L3 & 98.0 & 97.0 & 99.5 & 77.5 & 0.0 & 97.5 & 95.5 & {---} & 15.5 & {---} & 92.5 & {---} & {---} & {---} & 94.5 & 47.5 & 0.0 & 67.9 \\
\bfseries L4 & {---} & {---} & {---} & {---} & {---} & {---} & {---} & 92.0 & {---} & 0.0 & {---} & {---} & 0.0 & 95.0 & {---} & {---} & {---} & 46.8 \\
\multicolumn{18}{@{}l}{\textit{Cross-Attn + Patches}} \\
\bfseries L1 & 88.5 & 66.0 & 92.0 & 8.0 & 0.0 & 79.0 & 96.5 & {---} & 10.5 & {---} & 93.0 & 75.5 & {---} & {---} & 58.5 & 89.5 & 0.0 & 58.2 \\
\bfseries L2 & 87.5 & 56.0 & 92.0 & 6.5 & 0.5 & 87.0 & 91.5 & {---} & 4.5 & {---} & 92.0 & 82.0 & {---} & {---} & 58.5 & 87.0 & 0.0 & 57.3 \\
\bfseries L3 & 61.0 & 59.5 & 69.0 & 7.0 & 0.0 & 84.5 & 50.5 & {---} & 5.0 & {---} & 87.5 & {---} & {---} & {---} & 46.5 & 60.5 & 0.0 & 44.2 \\
\bfseries L4 & {---} & {---} & {---} & {---} & {---} & {---} & {---} & 38.0 & {---} & 0.0 & {---} & {---} & 0.0 & 25.5 & {---} & {---} & {---} & 15.9 \\
\multicolumn{18}{@{}l}{\textit{Concatenate + Obj-Centric}} \\
\bfseries L1 & 99.5 & 99.5 & 98.5 & 98.0 & 6.5 & 99.5 & 99.0 & {---} & 13.0 & {---} & 90.0 & 92.0 & {---} & {---} & 95.5 & 47.5 & 0.0 & 72.2 \\
\bfseries L2 & 99.0 & 100.0 & 99.5 & 97.5 & 4.5 & 98.5 & 99.5 & {---} & 13.5 & {---} & 90.5 & 91.0 & {---} & {---} & 89.5 & 45.0 & 0.0 & 71.4 \\
\bfseries L3 & 91.5 & 90.0 & 100.0 & 99.5 & 9.5 & 72.0 & 84.0 & {---} & 12.5 & {---} & 93.0 & {---} & {---} & {---} & 94.0 & 42.0 & 0.0 & 65.7 \\
\bfseries L4 & {---} & {---} & {---} & {---} & {---} & {---} & {---} & 90.5 & {---} & 0.0 & {---} & {---} & 0.0 & 91.5 & {---} & {---} & {---} & 45.5 \\
\multicolumn{18}{@{}l}{\textit{Concatenate + Patches}} \\
\bfseries L1 & 94.0 & 72.5 & 97.5 & 6.5 & 1.5 & 92.5 & 90.5 & {---} & 26.0 & {---} & 87.0 & 74.0 & {---} & {---} & 59.0 & 95.0 & 0.0 & 61.2 \\
\bfseries L2 & 83.5 & 62.5 & 90.5 & 9.5 & 2.0 & 91.5 & 82.5 & {---} & 14.0 & {---} & 90.0 & 79.0 & {---} & {---} & 52.0 & 91.5 & 0.0 & 57.6 \\
\bfseries L3 & 57.5 & 68.5 & 81.5 & 11.0 & 1.5 & 86.5 & 49.5 & {---} & 18.0 & {---} & 81.5 & {---} & {---} & {---} & 32.0 & 64.0 & 0.0 & 46.0 \\
\bfseries L4 & {---} & {---} & {---} & {---} & {---} & {---} & {---} & 34.0 & {---} & 2.5 & {---} & {---} & 0.0 & 15.0 & {---} & {---} & {---} & 12.9 \\
\bottomrule
\end{tabular}
\caption{Per-task average success rate when evaluating performance \textbf{without allowing models to recover from mistakes.} This table corresponds to \cref{tab:perf-no-mistakes-on-para}. All models are trained \textbf{on paraphrased instructions}.}
\label{tab:strict-time-full}
\end{table*}

\begin{table*}[tbph]
\centering
\scriptsize
\sisetup{table-format=2.1}
\renewcommand{\arraystretch}{1.2}
\setlength{\tabcolsep}{4.5pt}
\begin{tabular}{@{} l *{18}{S} @{}}
\toprule
& {T1} & {T2} & {T3} & {T4} & {T5} & {T6} & {T7} & {T8} & {T9} & {T10} & {T11} & {T12} & {T13} & {T14} & {T15} & {T16} & {T17} & {Avg.} \\
\midrule
\multicolumn{18}{@{}l}{\textit{\textbf{Mask Instructions; Mistakes Allowed}}} \\
\multicolumn{18}{@{}l}{\textit{Cross-Attn + Obj-Centric}} \\
\bfseries L1 & 99.5 & 99.0 & 22.0 & 0.0 & 0.0 & 96.0 & 100.0 & {---} & 2.0 & {---} & 30.5 & 97.5 & {---} & {---} & 73.5 & 23.0 & 38.0 & 52.4 \\
\bfseries L2 & 99.0 & 98.5 & 15.5 & 0.0 & 0.0 & 96.5 & 100.0 & {---} & 1.0 & {---} & 26.0 & 93.5 & {---} & {---} & 67.0 & 29.0 & 30.0 & 50.5 \\
\bfseries L3 & 97.0 & 92.5 & 17.5 & 0.0 & 0.0 & 84.0 & 94.5 & {---} & 1.5 & {---} & 30.0 & {---} & {---} & {---} & 48.5 & 22.0 & 0.0 & 40.6 \\
\bfseries L4 & {---} & {---} & {---} & {---} & {---} & {---} & {---} & 84.0 & {---} & 0.0 & {---} & {---} & 0.0 & 48.0 & {---} & {---} & {---} & 33.0 \\
\multicolumn{18}{@{}l}{\textit{Cross-Attn + Patches}} \\
\bfseries L1 & 62.0 & 24.0 & 11.5 & 0.0 & 0.0 & 46.5 & 57.0 & {---} & 8.0 & {---} & 13.5 & 89.0 & {---} & {---} & 7.5 & 19.5 & 5.5 & 26.5 \\
\bfseries L2 & 54.0 & 26.5 & 15.0 & 0.0 & 0.0 & 53.0 & 53.0 & {---} & 6.5 & {---} & 17.0 & 86.0 & {---} & {---} & 6.0 & 16.5 & 2.5 & 25.8 \\
\bfseries L3 & 46.0 & 30.0 & 12.5 & 0.0 & 0.0 & 44.5 & 50.0 & {---} & 5.5 & {---} & 18.0 & {---} & {---} & {---} & 19.0 & 14.5 & 2.0 & 20.2 \\
\bfseries L4 & {---} & {---} & {---} & {---} & {---} & {---} & {---} & 22.5 & {---} & 1.0 & {---} & {---} & 0.0 & 24.0 & {---} & {---} & {---} & 11.9 \\
\multicolumn{18}{@{}l}{\textit{Concatenate + Obj-Centric}} \\
\bfseries L1 & 73.0 & 71.5 & 8.0 & 0.0 & 0.0 & 71.5 & 75.0 & {---} & 1.0 & {---} & 1.5 & 88.5 & {---} & {---} & 1.0 & 11.0 & 0.0 & 30.9 \\
\bfseries L2 & 67.5 & 68.5 & 11.0 & 0.0 & 0.0 & 72.0 & 76.5 & {---} & 1.0 & {---} & 1.5 & 85.5 & {---} & {---} & 1.0 & 11.0 & 0.5 & 30.5 \\
\bfseries L3 & 58.0 & 63.5 & 9.0 & 0.0 & 0.0 & 59.5 & 60.0 & {---} & 1.5 & {---} & 3.0 & {---} & {---} & {---} & 4.0 & 6.5 & 0.0 & 22.1 \\
\bfseries L4 & {---} & {---} & {---} & {---} & {---} & {---} & {---} & 53.0 & {---} & 0.0 & {---} & {---} & 0.0 & 3.5 & {---} & {---} & {---} & 14.1 \\
\multicolumn{18}{@{}l}{\textit{Concatenate + Patches}} \\
\bfseries L1 & 70.0 & 65.0 & 4.0 & 0.0 & 0.0 & 56.0 & 62.5 & {---} & 1.5 & {---} & 13.5 & 74.5 & {---} & {---} & 28.5 & 17.5 & 2.0 & 30.4 \\
\bfseries L2 & 71.5 & 59.5 & 8.5 & 0.0 & 0.0 & 51.0 & 61.0 & {---} & 0.5 & {---} & 6.5 & 80.5 & {---} & {---} & 27.5 & 14.0 & 1.5 & 29.4 \\
\bfseries L3 & 62.5 & 53.0 & 7.0 & 0.0 & 0.0 & 53.0 & 55.5 & {---} & 0.0 & {---} & 16.0 & {---} & {---} & {---} & 38.0 & 15.0 & 0.0 & 25.0 \\
\bfseries L4 & {---} & {---} & {---} & {---} & {---} & {---} & {---} & 25.0 & {---} & 4.0 & {---} & {---} & 0.0 & 33.5 & {---} & {---} & {---} & 15.6 \\
\midrule[0.2pt]
\addlinespace[2pt]
\multicolumn{18}{@{}l}{\textit{\textbf{Mask Instructions; No Mistakes Allowed}}} \\
\multicolumn{18}{@{}l}{\textit{Cross-Attn + Obj-Centric}} \\
\bfseries L1 & 99.0 & 97.0 & 10.0 & 0.0 & 0.0 & 83.0 & 99.5 & {---} & 1.5 & {---} & 22.0 & 95.0 & {---} & {---} & 61.0 & 19.5 & 0.0 & 45.2 \\
\bfseries L2 & 98.0 & 95.5 & 8.5 & 0.0 & 0.0 & 86.5 & 98.0 & {---} & 0.5 & {---} & 13.0 & 89.5 & {---} & {---} & 55.5 & 26.0 & 0.0 & 43.9 \\
\bfseries L3 & 96.0 & 78.0 & 6.0 & 0.0 & 0.0 & 62.0 & 86.5 & {---} & 1.0 & {---} & 18.0 & {---} & {---} & {---} & 32.0 & 18.0 & 0.0 & 33.1 \\
\bfseries L4 & {---} & {---} & {---} & {---} & {---} & {---} & {---} & 76.0 & {---} & 0.0 & {---} & {---} & 0.0 & 33.5 & {---} & {---} & {---} & 27.4 \\
\multicolumn{18}{@{}l}{\textit{Cross-Attn + Patches}} \\
\bfseries L1 & 52.0 & 13.0 & 2.5 & 0.0 & 0.0 & 37.0 & 45.0 & {---} & 1.5 & {---} & 4.5 & 79.5 & {---} & {---} & 3.0 & 16.0 & 0.0 & 19.5 \\
\bfseries L2 & 44.0 & 15.5 & 4.5 & 0.0 & 0.0 & 41.0 & 41.0 & {---} & 1.0 & {---} & 5.0 & 75.5 & {---} & {---} & 1.0 & 14.0 & 0.0 & 18.7 \\
\bfseries L3 & 37.5 & 17.5 & 3.5 & 0.0 & 0.0 & 35.5 & 44.5 & {---} & 1.0 & {---} & 5.0 & {---} & {---} & {---} & 15.0 & 14.0 & 0.0 & 14.5 \\
\bfseries L4 & {---} & {---} & {---} & {---} & {---} & {---} & {---} & 12.5 & {---} & 0.0 & {---} & {---} & 0.0 & 18.0 & {---} & {---} & {---} & 7.6 \\
\multicolumn{18}{@{}l}{\textit{Concatenate + Obj-Centric}} \\
\bfseries L1 & 12.5 & 14.5 & 0.0 & 0.0 & 0.0 & 12.0 & 15.0 & {---} & 0.0 & {---} & 0.0 & 33.0 & {---} & {---} & 0.0 & 4.5 & 0.0 & 7.0 \\
\bfseries L2 & 16.5 & 15.5 & 0.0 & 0.0 & 0.0 & 10.0 & 15.5 & {---} & 0.0 & {---} & 0.5 & 32.5 & {---} & {---} & 0.5 & 2.5 & 0.0 & 7.2 \\
\bfseries L3 & 10.0 & 11.0 & 0.0 & 0.0 & 0.0 & 10.0 & 10.0 & {---} & 0.0 & {---} & 1.0 & {---} & {---} & {---} & 0.0 & 0.0 & 0.0 & 3.5 \\
\bfseries L4 & {---} & {---} & {---} & {---} & {---} & {---} & {---} & 9.0 & {---} & 0.0 & {---} & {---} & 0.0 & 0.0 & {---} & {---} & {---} & 2.2 \\
\multicolumn{18}{@{}l}{\textit{Concatenate + Patches}} \\
\bfseries L1 & 56.5 & 50.0 & 0.5 & 0.0 & 0.0 & 45.0 & 46.0 & {---} & 0.0 & {---} & 6.0 & 63.0 & {---} & {---} & 11.5 & 13.5 & 0.0 & 22.5 \\
\bfseries L2 & 59.5 & 46.0 & 2.0 & 0.0 & 0.0 & 36.5 & 51.5 & {---} & 0.0 & {---} & 1.5 & 71.0 & {---} & {---} & 11.0 & 10.0 & 0.0 & 22.2 \\
\bfseries L3 & 52.0 & 44.0 & 0.5 & 0.0 & 0.0 & 40.5 & 42.5 & {---} & 0.0 & {---} & 5.0 & {---} & {---} & {---} & 24.5 & 9.5 & 0.0 & 18.2 \\
\bfseries L4 & {---} & {---} & {---} & {---} & {---} & {---} & {---} & 19.0 & {---} & 0.5 & {---} & {---} & 0.0 & 15.5 & {---} & {---} & {---} & 8.8 \\
\bottomrule
\end{tabular}
\caption{Per-task average success rate when evaluating performance with \textbf{entirely masked instructions}. This compares models' ability to recover from mistakes (top) versus acting without making mistakes (bottom). All models are trained \textbf{on paraphrased instructions}. This table corresponds to \cref{tab:perf-no-instructions-trained-para}.}
\label{tab:no-instruct-full}
\end{table*}

\begin{table*}[tbh]
\centering
\scriptsize
\sisetup{table-format=2.1}
\renewcommand{\arraystretch}{1.2}
\setlength{\tabcolsep}{4.5pt}
\begin{tabular}{@{} l *{18}{S} @{}}
\toprule
& {T1} & {T2} & {T3} & {T4} & {T5} & {T6} & {T7} & {T8} & {T9} & {T10} & {T11} & {T12} & {T13} & {T14} & {T15} & {T16} & {T17} & {Avg.} \\
\midrule
\multicolumn{18}{@{}l}{\textit{\textbf{Distracting}}} \\
\multicolumn{18}{@{}l}{\textit{Cross-Attn + Obj-Centric}} \\
\bfseries L1 & 65.5 & 84.5 & 99.0 & 0.5 & 0.0 & 87.0 & 73.5 & {---} & 4.5 & {---} & 88.0 & 74.5 & {---} & {---} & 78.0 & 44.5 & 0.0 & 53.8 \\
\bfseries L2 & 66.5 & 86.0 & 97.5 & 1.0 & 0.0 & 84.0 & 72.5 & {---} & 2.5 & {---} & 86.5 & 70.0 & {---} & {---} & 69.0 & 46.0 & 0.0 & 52.4 \\
\bfseries L3 & 59.0 & 70.0 & 98.0 & 1.5 & 0.0 & 83.0 & 70.5 & {---} & 5.0 & {---} & 83.5 & {---} & {---} & {---} & 44.5 & 44.5 & 0.0 & 46.6 \\
\bfseries L4 & {---} & {---} & {---} & {---} & {---} & {---} & {---} & 79.5 & {---} & 0.0 & {---} & {---} & 0.0 & 59.5 & {---} & {---} & {---} & 34.8 \\
\multicolumn{18}{@{}l}{\textit{Cross-Attn + Patches}} \\
\bfseries L1 & 22.5 & 25.0 & 19.5 & 0.0 & 0.0 & 16.0 & 22.5 & {---} & 1.0 & {---} & 87.5 & 56.0 & {---} & {---} & 25.0 & 87.5 & 0.0 & 27.9 \\
\bfseries L2 & 20.5 & 15.0 & 21.0 & 0.0 & 0.0 & 15.0 & 22.0 & {---} & 0.5 & {---} & 88.0 & 57.0 & {---} & {---} & 26.5 & 91.0 & 0.0 & 27.4 \\
\bfseries L3 & 11.0 & 20.0 & 12.5 & 0.0 & 0.0 & 12.5 & 9.5 & {---} & 0.5 & {---} & 84.5 & {---} & {---} & {---} & 9.0 & 59.5 & 0.0 & 18.2 \\
\bfseries L4 & {---} & {---} & {---} & {---} & {---} & {---} & {---} & 12.0 & {---} & 1.0 & {---} & {---} & 0.0 & 2.0 & {---} & {---} & {---} & 3.8 \\
\multicolumn{18}{@{}l}{\textit{Concatenate + Obj-Centric}} \\
\bfseries L1 & 98.5 & 99.5 & 99.0 & 1.5 & 0.0 & 96.0 & 99.0 & {---} & 3.5 & {---} & 86.0 & 76.5 & {---} & {---} & 79.0 & 44.5 & 0.0 & 60.2 \\
\bfseries L2 & 99.5 & 98.0 & 99.5 & 0.0 & 0.0 & 98.0 & 99.5 & {---} & 4.5 & {---} & 85.0 & 73.5 & {---} & {---} & 73.0 & 47.5 & 0.0 & 59.8 \\
\bfseries L3 & 95.0 & 83.0 & 100.0 & 0.0 & 0.0 & 80.0 & 86.5 & {---} & 4.0 & {---} & 84.5 & {---} & {---} & {---} & 64.0 & 42.5 & 0.0 & 53.3 \\
\bfseries L4 & {---} & {---} & {---} & {---} & {---} & {---} & {---} & 87.0 & {---} & 0.0 & {---} & {---} & 0.0 & 69.0 & {---} & {---} & {---} & 39.0 \\
\multicolumn{18}{@{}l}{\textit{Concatenate + Patches}} \\
\bfseries L1 & 35.5 & 30.0 & 18.0 & 0.0 & 0.0 & 18.0 & 21.5 & {---} & 7.5 & {---} & 84.0 & 55.0 & {---} & {---} & 12.5 & 98.0 & 0.0 & 29.2 \\
\bfseries L2 & 27.0 & 20.0 & 21.5 & 0.0 & 0.0 & 15.5 & 14.5 & {---} & 2.5 & {---} & 84.5 & 61.0 & {---} & {---} & 12.5 & 91.5 & 0.0 & 27.0 \\
\bfseries L3 & 13.0 & 20.5 & 13.5 & 0.0 & 0.0 & 16.0 & 10.0 & {---} & 2.5 & {---} & 81.0 & {---} & {---} & {---} & 2.5 & 62.0 & 0.0 & 18.4 \\
\bfseries L4 & {---} & {---} & {---} & {---} & {---} & {---} & {---} & 14.5 & {---} & 3.0 & {---} & {---} & 0.0 & 3.0 & {---} & {---} & {---} & 5.1 \\
\midrule[0.2pt]
\addlinespace[2pt]
\multicolumn{18}{@{}l}{\textit{\textbf{Extreme}}} \\
\multicolumn{18}{@{}l}{\textit{Cross-Attn + Obj-Centric}} \\
\bfseries L1 & 97.5 & 97.5 & 74.5 & 96.0 & 5.5 & 98.5 & 97.0 & {---} & 12.0 & {---} & 2.0 & 33.0 & {---} & {---} & 77.0 & 0.0 & 0.0 & 53.1 \\
\bfseries L2 & 98.0 & 96.5 & 80.0 & 95.0 & 7.0 & 93.5 & 98.0 & {---} & 13.0 & {---} & 0.5 & 34.0 & {---} & {---} & 79.5 & 0.0 & 0.0 & 53.5 \\
\bfseries L3 & 100.0 & 99.5 & 79.5 & 98.5 & 5.0 & 95.0 & 99.5 & {---} & 16.0 & {---} & 0.5 & {---} & {---} & {---} & 72.5 & 0.0 & 0.0 & 55.5 \\
\bfseries L4 & {---} & {---} & {---} & {---} & {---} & {---} & {---} & 70.5 & {---} & 0.5 & {---} & {---} & 0.0 & 75.5 & {---} & {---} & {---} & 36.6 \\
\multicolumn{18}{@{}l}{\textit{Cross-Attn + Patches}} \\
\bfseries L1 & 19.0 & 7.0 & 52.0 & 8.5 & 1.0 & 13.0 & 9.0 & {---} & 9.0 & {---} & 0.5 & 33.5 & {---} & {---} & 16.5 & 0.0 & 0.0 & 13.0 \\
\bfseries L2 & 15.0 & 7.5 & 49.0 & 10.0 & 0.5 & 12.0 & 8.5 & {---} & 5.0 & {---} & 2.0 & 33.5 & {---} & {---} & 20.0 & 0.0 & 0.0 & 12.5 \\
\bfseries L3 & 11.0 & 7.0 & 46.5 & 6.0 & 0.5 & 10.5 & 8.0 & {---} & 8.0 & {---} & 2.5 & {---} & {---} & {---} & 16.5 & 0.0 & 0.0 & 9.7 \\
\bfseries L4 & {---} & {---} & {---} & {---} & {---} & {---} & {---} & 3.0 & {---} & 11.0 & {---} & {---} & 0.5 & 22.5 & {---} & {---} & {---} & 9.2 \\
\multicolumn{18}{@{}l}{\textit{Concatenate + Obj-Centric}} \\
\bfseries L1 & 25.0 & 4.0 & 77.0 & 99.0 & 4.5 & 0.5 & 2.5 & {---} & 15.5 & {---} & 1.5 & 32.5 & {---} & {---} & 30.5 & 0.0 & 0.0 & 22.5 \\
\bfseries L2 & 22.0 & 1.5 & 71.5 & 99.0 & 7.0 & 0.5 & 3.0 & {---} & 13.5 & {---} & 4.0 & 36.5 & {---} & {---} & 40.0 & 0.0 & 0.0 & 23.0 \\
\bfseries L3 & 30.5 & 4.0 & 77.0 & 100.0 & 7.0 & 1.0 & 2.0 & {---} & 15.5 & {---} & 3.0 & {---} & {---} & {---} & 38.0 & 0.0 & 0.0 & 23.2 \\
\bfseries L4 & {---} & {---} & {---} & {---} & {---} & {---} & {---} & 0.5 & {---} & 8.5 & {---} & {---} & 1.5 & 31.5 & {---} & {---} & {---} & 10.5 \\
\multicolumn{18}{@{}l}{\textit{Concatenate + Patches}} \\
\bfseries L1 & 12.5 & 13.5 & 74.5 & 7.0 & 2.5 & 11.5 & 15.5 & {---} & 22.5 & {---} & 3.5 & 28.5 & {---} & {---} & 19.5 & 0.0 & 0.0 & 16.2 \\
\bfseries L2 & 13.5 & 11.0 & 66.5 & 6.5 & 0.5 & 19.5 & 7.5 & {---} & 18.0 & {---} & 5.0 & 29.0 & {---} & {---} & 18.5 & 0.0 & 0.0 & 15.0 \\
\bfseries L3 & 15.0 & 13.0 & 51.0 & 7.0 & 1.0 & 15.0 & 5.0 & {---} & 16.5 & {---} & 3.5 & {---} & {---} & {---} & 18.0 & 0.0 & 0.0 & 12.1 \\
\bfseries L4 & {---} & {---} & {---} & {---} & {---} & {---} & {---} & 4.5 & {---} & 18.0 & {---} & {---} & 0.0 & 26.0 & {---} & {---} & {---} & 12.1 \\
\midrule[0.2pt]
\addlinespace[2pt]
\multicolumn{18}{@{}l}{\textit{\textbf{Extremely Distracting}}} \\
\multicolumn{18}{@{}l}{\textit{Cross-Attn + Obj-Centric}} \\
\bfseries L1 & 48.0 & 71.5 & 78.5 & 0.0 & 0.0 & 58.0 & 41.0 & {---} & 3.0 & {---} & 2.0 & 33.0 & {---} & {---} & 57.0 & 0.0 & 0.0 & 30.2 \\
\bfseries L2 & 43.5 & 68.5 & 80.5 & 0.0 & 0.0 & 60.0 & 47.0 & {---} & 4.5 & {---} & 0.5 & 36.0 & {---} & {---} & 58.5 & 0.0 & 0.0 & 30.7 \\
\bfseries L3 & 49.0 & 77.5 & 74.0 & 0.0 & 0.0 & 67.5 & 50.0 & {---} & 4.0 & {---} & 0.5 & {---} & {---} & {---} & 73.0 & 0.0 & 0.0 & 33.0 \\
\bfseries L4 & {---} & {---} & {---} & {---} & {---} & {---} & {---} & 58.0 & {---} & 0.0 & {---} & {---} & 0.5 & 68.5 & {---} & {---} & {---} & 31.8 \\
\multicolumn{18}{@{}l}{\textit{Cross-Attn + Patches}} \\
\bfseries L1 & 3.5 & 1.0 & 10.0 & 0.0 & 0.0 & 4.0 & 1.5 & {---} & 1.5 & {---} & 0.5 & 32.0 & {---} & {---} & 4.0 & 0.0 & 0.0 & 4.5 \\
\bfseries L2 & 1.5 & 0.0 & 9.5 & 0.0 & 0.0 & 2.0 & 1.0 & {---} & 0.0 & {---} & 1.5 & 32.5 & {---} & {---} & 2.0 & 0.0 & 0.0 & 3.8 \\
\bfseries L3 & 2.0 & 4.0 & 8.5 & 0.0 & 0.0 & 5.0 & 2.0 & {---} & 1.0 & {---} & 0.0 & {---} & {---} & {---} & 3.0 & 0.0 & 0.0 & 2.1 \\
\bfseries L4 & {---} & {---} & {---} & {---} & {---} & {---} & {---} & 2.5 & {---} & 8.5 & {---} & {---} & 0.0 & 0.5 & {---} & {---} & {---} & 2.9 \\
\multicolumn{18}{@{}l}{\textit{Concatenate + Obj-Centric}} \\
\bfseries L1 & 27.5 & 1.0 & 75.0 & 0.0 & 0.0 & 0.5 & 4.0 & {---} & 6.5 & {---} & 4.0 & 36.0 & {---} & {---} & 35.5 & 0.0 & 0.0 & 14.6 \\
\bfseries L2 & 26.5 & 0.5 & 75.0 & 0.0 & 0.0 & 1.5 & 3.5 & {---} & 6.5 & {---} & 0.5 & 41.0 & {---} & {---} & 31.0 & 0.0 & 0.0 & 14.3 \\
\bfseries L3 & 19.5 & 0.5 & 68.5 & 0.0 & 0.0 & 2.5 & 1.5 & {---} & 7.5 & {---} & 1.5 & {---} & {---} & {---} & 28.0 & 0.0 & 0.0 & 10.8 \\
\bfseries L4 & {---} & {---} & {---} & {---} & {---} & {---} & {---} & 0.5 & {---} & 0.0 & {---} & {---} & 0.0 & 33.5 & {---} & {---} & {---} & 8.5 \\
\multicolumn{18}{@{}l}{\textit{Concatenate + Patches}} \\
\bfseries L1 & 3.5 & 5.0 & 11.0 & 0.0 & 0.0 & 4.5 & 3.0 & {---} & 9.5 & {---} & 2.0 & 32.0 & {---} & {---} & 3.0 & 0.0 & 0.0 & 5.7 \\
\bfseries L2 & 4.0 & 4.0 & 14.0 & 0.0 & 0.0 & 2.5 & 4.0 & {---} & 3.0 & {---} & 1.5 & 31.5 & {---} & {---} & 2.5 & 0.0 & 0.0 & 5.2 \\
\bfseries L3 & 3.5 & 7.0 & 9.0 & 0.0 & 0.0 & 4.5 & 1.0 & {---} & 5.5 & {---} & 3.0 & {---} & {---} & {---} & 0.0 & 0.0 & 0.0 & 2.8 \\
\bfseries L4 & {---} & {---} & {---} & {---} & {---} & {---} & {---} & 3.5 & {---} & 12.5 & {---} & {---} & 0.0 & 0.5 & {---} & {---} & {---} & 4.1 \\
\bottomrule
\end{tabular}
\caption{Per-task average success rate when evaluating performance \textbf{across each difficulty level, without making mistakes.} This corresponds to the top of \cref{tab:perf-difficulty-on-para}. All models were trained \textbf{on paraphrased instructions}.}
\label{tab:difficulty-full}
\end{table*}

\begin{table*}[tbh]
\centering
\scriptsize
\sisetup{table-format=2.1}
\renewcommand{\arraystretch}{1.2}
\setlength{\tabcolsep}{4.5pt}
\begin{tabular}{@{} l *{18}{S} @{}}
\toprule
& {T1} & {T2} & {T3} & {T4} & {T5} & {T6} & {T7} & {T8} & {T9} & {T10} & {T11} & {T12} & {T13} & {T14} & {T15} & {T16} & {T17} & {Avg.} \\
\midrule
\multicolumn{18}{@{}l}{\textit{\textbf{Distracting}}} \\
\multicolumn{18}{@{}l}{\textit{Cross-Attn + Obj-Centric}} \\
\bfseries L1 & 69.0 & 63.5 & 10.5 & 0.0 & 0.0 & 67.5 & 70.5 & {---} & 0.5 & {---} & 10.5 & 69.0 & {---} & {---} & 47.0 & 21.5 & 0.0 & 33.0 \\
\bfseries L2 & 67.5 & 74.0 & 8.0 & 0.0 & 0.0 & 67.5 & 71.0 & {---} & 0.0 & {---} & 10.0 & 65.5 & {---} & {---} & 38.0 & 17.0 & 0.0 & 32.2 \\
\bfseries L3 & 46.0 & 55.0 & 5.0 & 0.0 & 0.0 & 50.5 & 53.0 & {---} & 1.0 & {---} & 8.5 & {---} & {---} & {---} & 20.5 & 13.0 & 0.0 & 21.0 \\
\bfseries L4 & {---} & {---} & {---} & {---} & {---} & {---} & {---} & 62.0 & {---} & 0.0 & {---} & {---} & 0.0 & 25.5 & {---} & {---} & {---} & 21.9 \\
\multicolumn{18}{@{}l}{\textit{Cross-Attn + Patches}} \\
\bfseries L1 & 3.5 & 2.0 & 0.5 & 0.0 & 0.0 & 8.0 & 2.5 & {---} & 0.0 & {---} & 2.5 & 56.0 & {---} & {---} & 0.0 & 12.5 & 0.0 & 6.7 \\
\bfseries L2 & 2.5 & 4.5 & 0.0 & 0.0 & 0.0 & 4.0 & 3.5 & {---} & 0.0 & {---} & 3.0 & 62.0 & {---} & {---} & 0.0 & 16.0 & 0.0 & 7.3 \\
\bfseries L3 & 4.0 & 4.5 & 0.0 & 0.0 & 0.0 & 5.5 & 2.0 & {---} & 0.0 & {---} & 0.5 & {---} & {---} & {---} & 0.0 & 12.5 & 0.0 & 2.4 \\
\bfseries L4 & {---} & {---} & {---} & {---} & {---} & {---} & {---} & 7.5 & {---} & 3.0 & {---} & {---} & 0.0 & 0.0 & {---} & {---} & {---} & 2.6 \\
\multicolumn{18}{@{}l}{\textit{Concatenate + Obj-Centric}} \\
\bfseries L1 & 3.5 & 12.0 & 0.0 & 0.0 & 0.0 & 9.0 & 1.5 & {---} & 0.0 & {---} & 1.5 & 42.5 & {---} & {---} & 0.0 & 6.5 & 0.0 & 5.9 \\
\bfseries L2 & 3.5 & 4.0 & 0.0 & 0.0 & 0.0 & 4.0 & 4.5 & {---} & 0.0 & {---} & 1.0 & 40.0 & {---} & {---} & 0.0 & 2.0 & 0.0 & 4.5 \\
\bfseries L3 & 2.5 & 6.5 & 0.0 & 0.0 & 0.0 & 2.5 & 2.0 & {---} & 0.0 & {---} & 0.5 & {---} & {---} & {---} & 0.0 & 0.0 & 0.0 & 1.2 \\
\bfseries L4 & {---} & {---} & {---} & {---} & {---} & {---} & {---} & 2.0 & {---} & 0.0 & {---} & {---} & 0.0 & 0.5 & {---} & {---} & {---} & 0.6 \\
\multicolumn{18}{@{}l}{\textit{Concatenate + Patches}} \\
\bfseries L1 & 2.0 & 13.0 & 0.0 & 0.0 & 0.0 & 4.5 & 5.5 & {---} & 0.0 & {---} & 1.0 & 49.5 & {---} & {---} & 1.0 & 8.5 & 0.0 & 6.5 \\
\bfseries L2 & 3.5 & 12.5 & 0.0 & 0.0 & 0.0 & 4.5 & 3.0 & {---} & 0.0 & {---} & 3.5 & 48.0 & {---} & {---} & 0.5 & 11.5 & 0.0 & 6.7 \\
\bfseries L3 & 6.5 & 17.0 & 0.0 & 0.0 & 0.0 & 6.0 & 3.5 & {---} & 0.0 & {---} & 2.0 & {---} & {---} & {---} & 1.0 & 5.0 & 0.0 & 3.4 \\
\bfseries L4 & {---} & {---} & {---} & {---} & {---} & {---} & {---} & 6.5 & {---} & 0.0 & {---} & {---} & 0.0 & 0.0 & {---} & {---} & {---} & 1.6 \\
\midrule[0.2pt]
\addlinespace[2pt]
\multicolumn{18}{@{}l}{\textit{\textbf{Extreme}}} \\
\multicolumn{18}{@{}l}{\textit{Cross-Attn + Obj-Centric}} \\
\bfseries L1 & 61.0 & 14.5 & 10.5 & 0.0 & 0.0 & 2.5 & 30.5 & {---} & 0.5 & {---} & 0.5 & 39.5 & {---} & {---} & 37.5 & 0.0 & 0.0 & 15.2 \\
\bfseries L2 & 69.0 & 14.5 & 6.5 & 0.0 & 0.0 & 5.0 & 26.5 & {---} & 1.0 & {---} & 0.0 & 36.0 & {---} & {---} & 43.0 & 0.0 & 0.0 & 15.5 \\
\bfseries L3 & 77.0 & 31.0 & 6.0 & 0.0 & 0.0 & 9.0 & 41.0 & {---} & 0.5 & {---} & 2.0 & {---} & {---} & {---} & 42.5 & 0.0 & 0.0 & 17.4 \\
\bfseries L4 & {---} & {---} & {---} & {---} & {---} & {---} & {---} & 10.0 & {---} & 0.0 & {---} & {---} & 0.0 & 35.0 & {---} & {---} & {---} & 11.2 \\
\multicolumn{18}{@{}l}{\textit{Cross-Attn + Patches}} \\
\bfseries L1 & 4.0 & 4.0 & 3.5 & 0.0 & 0.0 & 6.0 & 6.0 & {---} & 1.5 & {---} & 1.0 & 37.5 & {---} & {---} & 10.5 & 0.0 & 0.0 & 5.7 \\
\bfseries L2 & 3.5 & 2.5 & 3.5 & 0.0 & 0.0 & 5.0 & 2.5 & {---} & 1.5 & {---} & 0.5 & 34.0 & {---} & {---} & 7.0 & 0.0 & 0.0 & 4.6 \\
\bfseries L3 & 4.5 & 1.0 & 7.5 & 0.0 & 0.0 & 4.5 & 2.5 & {---} & 4.0 & {---} & 0.0 & {---} & {---} & {---} & 8.0 & 0.0 & 0.0 & 2.7 \\
\bfseries L4 & {---} & {---} & {---} & {---} & {---} & {---} & {---} & 2.5 & {---} & 8.0 & {---} & {---} & 0.0 & 8.0 & {---} & {---} & {---} & 4.6 \\
\multicolumn{18}{@{}l}{\textit{Concatenate + Obj-Centric}} \\
\bfseries L1 & 8.5 & 6.0 & 1.0 & 0.0 & 0.0 & 8.0 & 5.0 & {---} & 0.5 & {---} & 0.5 & 29.0 & {---} & {---} & 0.5 & 0.0 & 0.0 & 4.5 \\
\bfseries L2 & 2.5 & 3.5 & 0.0 & 0.0 & 0.0 & 7.0 & 7.5 & {---} & 0.0 & {---} & 0.5 & 21.5 & {---} & {---} & 1.0 & 0.0 & 0.0 & 3.3 \\
\bfseries L3 & 6.0 & 4.0 & 0.0 & 0.0 & 0.0 & 8.5 & 4.5 & {---} & 0.0 & {---} & 0.5 & {---} & {---} & {---} & 0.0 & 0.0 & 0.0 & 2.0 \\
\bfseries L4 & {---} & {---} & {---} & {---} & {---} & {---} & {---} & 3.5 & {---} & 5.0 & {---} & {---} & 0.5 & 0.0 & {---} & {---} & {---} & 2.2 \\
\multicolumn{18}{@{}l}{\textit{Concatenate + Patches}} \\
\bfseries L1 & 4.5 & 3.0 & 1.5 & 0.0 & 0.0 & 4.0 & 5.0 & {---} & 0.0 & {---} & 1.5 & 20.5 & {---} & {---} & 11.0 & 0.0 & 0.0 & 3.9 \\
\bfseries L2 & 7.0 & 5.5 & 0.5 & 0.0 & 0.0 & 6.0 & 6.0 & {---} & 0.0 & {---} & 1.0 & 18.5 & {---} & {---} & 9.0 & 0.0 & 0.0 & 4.1 \\
\bfseries L3 & 8.0 & 2.0 & 0.0 & 0.0 & 0.0 & 5.5 & 4.5 & {---} & 0.0 & {---} & 0.0 & {---} & {---} & {---} & 12.0 & 0.0 & 0.0 & 2.7 \\
\bfseries L4 & {---} & {---} & {---} & {---} & {---} & {---} & {---} & 2.5 & {---} & 5.0 & {---} & {---} & 0.0 & 22.0 & {---} & {---} & {---} & 7.4 \\
\midrule[0.2pt]
\addlinespace[2pt]
\multicolumn{18}{@{}l}{\textit{\textbf{Extremely Distracting}}} \\
\multicolumn{18}{@{}l}{\textit{Cross-Attn + Obj-Centric}} \\
\bfseries L1 & 22.5 & 3.0 & 9.0 & 0.0 & 0.0 & 8.5 & 17.5 & {---} & 1.0 & {---} & 2.5 & 36.5 & {---} & {---} & 31.0 & 0.0 & 0.0 & 10.1 \\
\bfseries L2 & 14.5 & 6.0 & 7.0 & 0.0 & 0.0 & 6.5 & 14.0 & {---} & 0.5 & {---} & 1.0 & 35.0 & {---} & {---} & 29.0 & 0.0 & 0.0 & 8.7 \\
\bfseries L3 & 23.0 & 10.5 & 7.5 & 0.0 & 0.0 & 6.0 & 17.5 & {---} & 0.5 & {---} & 2.0 & {---} & {---} & {---} & 37.0 & 0.0 & 0.0 & 8.7 \\
\bfseries L4 & {---} & {---} & {---} & {---} & {---} & {---} & {---} & 9.5 & {---} & 0.0 & {---} & {---} & 0.0 & 34.0 & {---} & {---} & {---} & 10.9 \\
\multicolumn{18}{@{}l}{\textit{Cross-Attn + Patches}} \\
\bfseries L1 & 1.0 & 2.5 & 0.0 & 0.0 & 0.0 & 1.5 & 1.0 & {---} & 0.0 & {---} & 0.0 & 32.0 & {---} & {---} & 0.0 & 0.0 & 0.0 & 2.9 \\
\bfseries L2 & 1.0 & 1.0 & 0.0 & 0.0 & 0.0 & 1.0 & 0.5 & {---} & 0.0 & {---} & 0.5 & 32.5 & {---} & {---} & 0.5 & 0.0 & 0.0 & 2.8 \\
\bfseries L3 & 1.5 & 0.5 & 0.0 & 0.0 & 0.0 & 0.0 & 2.0 & {---} & 0.0 & {---} & 0.5 & {---} & {---} & {---} & 0.0 & 0.0 & 0.0 & 0.4 \\
\bfseries L4 & {---} & {---} & {---} & {---} & {---} & {---} & {---} & 2.0 & {---} & 12.0 & {---} & {---} & 0.5 & 0.0 & {---} & {---} & {---} & 3.6 \\
\multicolumn{18}{@{}l}{\textit{Concatenate + Obj-Centric}} \\
\bfseries L1 & 9.5 & 5.0 & 0.5 & 0.0 & 0.0 & 6.0 & 6.5 & {---} & 0.5 & {---} & 0.0 & 28.0 & {---} & {---} & 0.0 & 0.0 & 0.0 & 4.3 \\
\bfseries L2 & 3.5 & 5.0 & 0.0 & 0.0 & 0.0 & 6.5 & 7.5 & {---} & 0.0 & {---} & 0.0 & 32.5 & {---} & {---} & 0.5 & 0.0 & 0.0 & 4.3 \\
\bfseries L3 & 3.0 & 7.5 & 0.0 & 0.0 & 0.0 & 7.0 & 5.0 & {---} & 0.0 & {---} & 1.0 & {---} & {---} & {---} & 0.5 & 0.0 & 0.0 & 2.0 \\
\bfseries L4 & {---} & {---} & {---} & {---} & {---} & {---} & {---} & 3.0 & {---} & 2.5 & {---} & {---} & 0.0 & 0.0 & {---} & {---} & {---} & 1.4 \\
\multicolumn{18}{@{}l}{\textit{Concatenate + Patches}} \\
\bfseries L1 & 1.5 & 2.0 & 0.0 & 0.0 & 0.0 & 1.5 & 0.0 & {---} & 0.0 & {---} & 0.0 & 16.5 & {---} & {---} & 0.5 & 0.0 & 0.0 & 1.7 \\
\bfseries L2 & 1.0 & 1.0 & 0.0 & 0.0 & 0.0 & 4.0 & 1.0 & {---} & 0.0 & {---} & 0.0 & 13.5 & {---} & {---} & 0.0 & 0.0 & 0.0 & 1.6 \\
\bfseries L3 & 0.5 & 0.5 & 0.0 & 0.0 & 0.0 & 1.0 & 1.0 & {---} & 0.0 & {---} & 0.5 & {---} & {---} & {---} & 1.0 & 0.0 & 0.0 & 0.4 \\
\bfseries L4 & {---} & {---} & {---} & {---} & {---} & {---} & {---} & 2.5 & {---} & 2.0 & {---} & {---} & 0.5 & 0.0 & {---} & {---} & {---} & 1.2 \\
\bottomrule
\end{tabular}
\caption{Per-task average success rate when evaluating performance \textbf{across each difficulty level} when the \textbf{instruction is entirely masked}. The model must perform \textbf{without making mistakes}. This corresponds to the bottom of \cref{tab:perf-difficulty-on-para}. All models were trained \textbf{on paraphrased instructions}.}
\label{tab:difficulty-no-prompt-full}
\end{table*}

\begin{table*}[tbph]
\centering
\scriptsize
\sisetup{table-format=2.1}
\renewcommand{\arraystretch}{1.2}
\setlength{\tabcolsep}{4.5pt}
\begin{tabular}{@{} l *{18}{S} @{}}
\toprule
& {T1} & {T2} & {T3} & {T4} & {T5} & {T6} & {T7} & {T8} & {T9} & {T10} & {T11} & {T12} & {T13} & {T14} & {T15} & {T16} & {T17} & {Avg.} \\
\midrule
\multicolumn{18}{@{}l}{\textit{\textbf{Permute Object Order; Can Recover From Mistakes}}} \\
\multicolumn{18}{@{}l}{\textit{Cross-Attn + Obj-Centric}} \\
\bfseries L1 & 60.5 & 58.5 & 56.0 & 9.5 & 0.5 & 56.0 & 63.0 & {---} & 14.5 & {---} & 92.0 & 55.0 & {---} & {---} & 33.5 & 28.5 & 4.5 & 40.9 \\
\bfseries L2 & 60.0 & 51.5 & 61.5 & 12.0 & 0.0 & 53.5 & 54.0 & {---} & 8.5 & {---} & 91.0 & 49.5 & {---} & {---} & 34.5 & 28.5 & 3.5 & 39.1 \\
\bfseries L3 & 42.5 & 38.5 & 58.5 & 11.0 & 1.0 & 47.5 & 44.5 & {---} & 12.0 & {---} & 91.0 & {---} & {---} & {---} & 27.5 & 25.5 & 0.0 & 33.3 \\
\bfseries L4 & {---} & {---} & {---} & {---} & {---} & {---} & {---} & 20.5 & {---} & 0.0 & {---} & {---} & 0.0 & 26.5 & {---} & {---} & {---} & 11.8 \\
\multicolumn{18}{@{}l}{\textit{Concatenate + Obj-Centric}} \\
\bfseries L1 & 59.5 & 61.0 & 50.5 & 19.5 & 3.5 & 63.0 & 60.0 & {---} & 8.5 & {---} & 91.5 & 42.0 & {---} & {---} & 30.5 & 35.0 & 3.0 & 40.6 \\
\bfseries L2 & 65.5 & 56.5 & 56.5 & 13.0 & 1.0 & 56.0 & 58.5 & {---} & 10.5 & {---} & 96.5 & 42.5 & {---} & {---} & 32.0 & 35.0 & 4.5 & 40.6 \\
\bfseries L3 & 53.5 & 49.0 & 59.0 & 17.0 & 3.0 & 42.5 & 44.0 & {---} & 8.5 & {---} & 93.5 & {---} & {---} & {---} & 29.5 & 35.5 & 0.5 & 36.3 \\
\bfseries L4 & {---} & {---} & {---} & {---} & {---} & {---} & {---} & 21.5 & {---} & 1.0 & {---} & {---} & 0.0 & 35.5 & {---} & {---} & {---} & 14.5 \\
\midrule[0.2pt]
\addlinespace[2pt]
\multicolumn{18}{@{}l}{\textit{\textbf{Permute Object Order; No Mistakes Allowed}}} \\
\multicolumn{18}{@{}l}{\textit{Cross-Attn + Obj-Centric}} \\
\bfseries L1 & 22.5 & 19.0 & 44.5 & 6.5 & 0.0 & 23.5 & 20.5 & {---} & 5.0 & {---} & 91.0 & 41.5 & {---} & {---} & 24.5 & 25.5 & 0.0 & 24.9 \\
\bfseries L2 & 19.5 & 19.5 & 49.0 & 8.5 & 0.0 & 23.5 & 20.5 & {---} & 1.5 & {---} & 91.0 & 40.5 & {---} & {---} & 20.0 & 26.0 & 0.0 & 24.6 \\
\bfseries L3 & 14.5 & 13.0 & 47.5 & 6.0 & 0.0 & 17.5 & 17.0 & {---} & 5.0 & {---} & 90.5 & {---} & {---} & {---} & 13.0 & 24.5 & 0.0 & 20.7 \\
\bfseries L4 & {---} & {---} & {---} & {---} & {---} & {---} & {---} & 7.5 & {---} & 0.0 & {---} & {---} & 0.0 & 16.0 & {---} & {---} & {---} & 5.9 \\
\multicolumn{18}{@{}l}{\textit{Concatenate + Obj-Centric}} \\
\bfseries L1 & 21.0 & 27.0 & 40.5 & 19.0 & 2.0 & 41.0 & 30.0 & {---} & 2.0 & {---} & 89.5 & 39.0 & {---} & {---} & 18.0 & 29.5 & 0.0 & 27.6 \\
\bfseries L2 & 20.5 & 28.0 & 49.5 & 12.0 & 0.5 & 34.0 & 27.5 & {---} & 6.5 & {---} & 94.0 & 39.5 & {---} & {---} & 17.0 & 32.0 & 0.0 & 27.8 \\
\bfseries L3 & 17.0 & 20.5 & 52.5 & 16.5 & 2.0 & 24.0 & 16.5 & {---} & 6.0 & {---} & 90.5 & {---} & {---} & {---} & 17.0 & 35.5 & 0.0 & 24.8 \\
\bfseries L4 & {---} & {---} & {---} & {---} & {---} & {---} & {---} & 13.0 & {---} & 0.0 & {---} & {---} & 0.0 & 20.0 & {---} & {---} & {---} & 8.2 \\
\bottomrule
\end{tabular}
\caption{Per-task average success rate when evaluating performance for \textbf{object-centric models with a permuted object order} per observation during inference. This table compares a models' ability to recover from mistakes (top) versus acting without making mistakes (bottom). All models are trained \textbf{on paraphrased instructions}.}
\label{tab:shuffle-para-full}
\end{table*}

\begin{table*}[tbph]
\centering
\scriptsize
\sisetup{table-format=2.1}
\renewcommand{\arraystretch}{1.2}
\setlength{\tabcolsep}{4.5pt}
\begin{tabular}{@{} l *{18}{S} @{}}
\toprule
& {T1} & {T2} & {T3} & {T4} & {T5} & {T6} & {T7} & {T8} & {T9} & {T10} & {T11} & {T12} & {T13} & {T14} & {T15} & {T16} & {T17} & {Avg.} \\
\midrule
\multicolumn{18}{@{}l}{\textit{\textbf{Permute Object Order; Distracting; Can Recover From Mistakes}}} \\
\multicolumn{18}{@{}l}{\textit{Cross-Attn + Obj-Centric}} \\
\bfseries L1 & 12.5 & 26.0 & 17.0 & 0.0 & 0.0 & 13.0 & 13.0 & {---} & 0.5 & {---} & 89.5 & 46.5 & {---} & {---} & 5.0 & 29.0 & 0.0 & 19.4 \\
\bfseries L2 & 11.0 & 23.5 & 19.5 & 0.5 & 0.0 & 10.5 & 9.5 & {---} & 1.5 & {---} & 88.5 & 46.0 & {---} & {---} & 4.5 & 25.5 & 0.0 & 18.5 \\
\bfseries L3 & 7.5 & 10.5 & 19.0 & 0.5 & 0.0 & 10.0 & 9.0 & {---} & 1.0 & {---} & 87.0 & {---} & {---} & {---} & 6.5 & 33.0 & 0.0 & 15.3 \\
\bfseries L4 & {---} & {---} & {---} & {---} & {---} & {---} & {---} & 8.0 & {---} & 0.0 & {---} & {---} & 0.0 & 2.5 & {---} & {---} & {---} & 2.6 \\
\multicolumn{18}{@{}l}{\textit{Concatenate + Obj-Centric}} \\
\bfseries L1 & 6.0 & 24.5 & 11.5 & 0.0 & 0.0 & 21.0 & 6.5 & {---} & 1.0 & {---} & 91.0 & 20.5 & {---} & {---} & 1.5 & 35.0 & 0.0 & 16.8 \\
\bfseries L2 & 5.5 & 25.5 & 12.5 & 0.0 & 0.0 & 12.0 & 9.0 & {---} & 1.5 & {---} & 86.0 & 21.5 & {---} & {---} & 2.0 & 31.5 & 0.0 & 15.9 \\
\bfseries L3 & 6.5 & 18.0 & 12.0 & 0.0 & 0.0 & 10.5 & 8.0 & {---} & 4.5 & {---} & 87.5 & {---} & {---} & {---} & 1.5 & 32.0 & 0.0 & 15.0 \\
\bfseries L4 & {---} & {---} & {---} & {---} & {---} & {---} & {---} & 10.0 & {---} & 0.0 & {---} & {---} & 0.0 & 2.5 & {---} & {---} & {---} & 3.1 \\
\midrule[0.2pt]
\addlinespace[2pt]
\multicolumn{18}{@{}l}{\textit{\textbf{Permute Object Order; Extreme; Can Recover From Mistakes}}} \\
\multicolumn{18}{@{}l}{\textit{Cross-Attn + Obj-Centric}} \\
\bfseries L1 & 37.0 & 33.5 & 48.0 & 12.0 & 1.5 & 24.0 & 33.5 & {---} & 15.0 & {---} & 1.0 & 31.0 & {---} & {---} & 34.0 & 0.0 & 0.5 & 20.8 \\
\bfseries L2 & 37.0 & 37.0 & 49.0 & 12.5 & 1.0 & 23.0 & 33.0 & {---} & 8.5 & {---} & 1.5 & 29.0 & {---} & {---} & 31.0 & 0.0 & 1.5 & 20.3 \\
\bfseries L3 & 38.0 & 34.0 & 49.5 & 15.0 & 1.5 & 25.5 & 34.0 & {---} & 11.0 & {---} & 0.5 & {---} & {---} & {---} & 25.0 & 0.0 & 0.0 & 19.5 \\
\bfseries L4 & {---} & {---} & {---} & {---} & {---} & {---} & {---} & 10.5 & {---} & 5.5 & {---} & {---} & 0.5 & 31.5 & {---} & {---} & {---} & 12.0 \\
\multicolumn{18}{@{}l}{\textit{Concatenate + Obj-Centric}} \\
\bfseries L1 & 37.0 & 23.0 & 40.0 & 14.0 & 2.5 & 18.5 & 25.0 & {---} & 10.5 & {---} & 5.0 & 17.0 & {---} & {---} & 38.5 & 0.0 & 1.5 & 17.9 \\
\bfseries L2 & 30.5 & 22.5 & 41.0 & 14.0 & 2.0 & 18.5 & 30.0 & {---} & 10.0 & {---} & 3.5 & 15.5 & {---} & {---} & 27.0 & 0.0 & 0.0 & 16.5 \\
\bfseries L3 & 36.0 & 22.0 & 40.5 & 12.0 & 2.0 & 21.5 & 22.0 & {---} & 9.0 & {---} & 6.0 & {---} & {---} & {---} & 29.0 & 0.5 & 0.0 & 16.7 \\
\bfseries L4 & {---} & {---} & {---} & {---} & {---} & {---} & {---} & 12.0 & {---} & 20.0 & {---} & {---} & 2.0 & 31.0 & {---} & {---} & {---} & 16.2 \\
\midrule[0.2pt]
\addlinespace[2pt]
\multicolumn{18}{@{}l}{\textit{\textbf{Permute Object Order; Extremely Distracting; Can Recover From Mistakes}}} \\
\multicolumn{18}{@{}l}{\textit{Cross-Attn + Obj-Centric}} \\
\bfseries L1 & 4.0 & 11.0 & 19.5 & 0.0 & 0.0 & 8.5 & 4.0 & {---} & 2.0 & {---} & 0.5 & 25.0 & {---} & {---} & 3.0 & 0.0 & 0.0 & 6.0 \\
\bfseries L2 & 5.0 & 12.5 & 13.5 & 0.0 & 0.0 & 9.0 & 7.5 & {---} & 1.0 & {---} & 2.0 & 36.5 & {---} & {---} & 4.5 & 0.0 & 0.5 & 7.1 \\
\bfseries L3 & 7.0 & 13.0 & 21.5 & 0.0 & 0.0 & 7.0 & 6.0 & {---} & 1.5 & {---} & 1.5 & {---} & {---} & {---} & 4.0 & 0.0 & 0.0 & 5.1 \\
\bfseries L4 & {---} & {---} & {---} & {---} & {---} & {---} & {---} & 8.0 & {---} & 0.0 & {---} & {---} & 1.0 & 4.0 & {---} & {---} & {---} & 3.2 \\
\multicolumn{18}{@{}l}{\textit{Concatenate + Obj-Centric}} \\
\bfseries L1 & 3.5 & 10.0 & 8.5 & 0.0 & 0.0 & 3.5 & 5.0 & {---} & 3.0 & {---} & 4.5 & 13.5 & {---} & {---} & 1.5 & 0.5 & 0.0 & 4.1 \\
\bfseries L2 & 2.5 & 8.5 & 9.5 & 0.0 & 0.0 & 7.0 & 6.0 & {---} & 0.0 & {---} & 4.0 & 16.5 & {---} & {---} & 3.0 & 1.5 & 0.0 & 4.5 \\
\bfseries L3 & 3.5 & 6.5 & 13.0 & 0.0 & 0.0 & 4.5 & 6.0 & {---} & 0.0 & {---} & 6.0 & {---} & {---} & {---} & 3.0 & 0.0 & 0.0 & 3.5 \\
\bfseries L4 & {---} & {---} & {---} & {---} & {---} & {---} & {---} & 8.0 & {---} & 1.0 & {---} & {---} & 0.0 & 2.0 & {---} & {---} & {---} & 2.8 \\
\bottomrule
\end{tabular}
\caption{Per-task average success rate when evaluating performance for models with \textbf{permuted object order per observation across each difficulty level}. All models are trained \textbf{on paraphrased instructions} and \textbf{can recover from mistakes} during inference. This table corresponds to the middle section of \cref{tab:perf-eval-shuffle-on-para}, and also provides per-task results for the \textit{Extremely Distracting} difficulty level.}
\label{tab:shuffle-difficulty-full}
\end{table*}

\begin{table*}[tbph]
\centering
\scriptsize
\sisetup{table-format=2.1}
\renewcommand{\arraystretch}{1.2}
\setlength{\tabcolsep}{4.5pt}
\begin{tabular}{@{} l *{18}{S} @{}}
\toprule
& {T1} & {T2} & {T3} & {T4} & {T5} & {T6} & {T7} & {T8} & {T9} & {T10} & {T11} & {T12} & {T13} & {T14} & {T15} & {T16} & {T17} & {Avg.} \\
\midrule
\multicolumn{18}{@{}l}{\textit{\textbf{Default Difficulty; Can Recover From Mistakes}}} \\
\multicolumn{18}{@{}l}{\textit{Cross-Attn + Obj-Centric}} \\
\bfseries L1 & 84.5 & 94.0 & 16.0 & 90.5 & 10.0 & 60.5 & 82.5 & {---} & 10.5 & {---} & 93.0 & 91.5 & {---} & {---} & 92.5 & 48.0 & 2.5 & 59.7 \\
\bfseries L2 & 31.5 & 41.0 & 16.0 & 73.0 & 7.0 & 68.0 & 24.5 & {---} & 5.5 & {---} & 93.0 & 86.0 & {---} & {---} & 53.0 & 46.5 & 2.0 & 42.1 \\
\bfseries L3 & 42.5 & 55.0 & 19.0 & 46.5 & 4.0 & 54.0 & 42.0 & {---} & 8.0 & {---} & 93.0 & {---} & {---} & {---} & 49.5 & 42.5 & 1.0 & 38.1 \\
\bfseries L4 & {---} & {---} & {---} & {---} & {---} & {---} & {---} & 23.5 & {---} & 0.0 & {---} & {---} & 0.0 & 34.0 & {---} & {---} & {---} & 14.4 \\
\multicolumn{18}{@{}l}{\textit{Concatenate + Obj-Centric}} \\
\bfseries L1 & 96.0 & 99.0 & 46.5 & 93.0 & 22.5 & 63.0 & 89.0 & {---} & 78.0 & {---} & 95.0 & 87.5 & {---} & {---} & 97.5 & 44.5 & 6.0 & 70.6 \\
\bfseries L2 & 35.5 & 44.5 & 52.0 & 88.0 & 16.5 & 54.0 & 20.5 & {---} & 39.5 & {---} & 94.0 & 87.5 & {---} & {---} & 65.0 & 47.5 & 4.0 & 49.9 \\
\bfseries L3 & 37.0 & 63.0 & 49.0 & 76.5 & 12.0 & 39.5 & 27.0 & {---} & 60.5 & {---} & 94.5 & {---} & {---} & {---} & 32.0 & 45.0 & 0.0 & 44.7 \\
\bfseries L4 & {---} & {---} & {---} & {---} & {---} & {---} & {---} & 29.0 & {---} & 6.0 & {---} & {---} & 0.0 & 23.0 & {---} & {---} & {---} & 14.5 \\
\midrule[0.2pt]
\addlinespace[2pt]
\multicolumn{18}{@{}l}{\textit{\textbf{Distracting Difficulty; Can Recover From Mistakes}}} \\
\multicolumn{18}{@{}l}{\textit{Cross-Attn + Obj-Centric}} \\
\bfseries L1 & 31.0 & 71.0 & 1.0 & 7.5 & 0.0 & 20.5 & 38.0 & {---} & 4.0 & {---} & 92.5 & 82.0 & {---} & {---} & 92.5 & 45.0 & 0.0 & 37.3 \\
\bfseries L2 & 1.0 & 11.5 & 0.5 & 4.5 & 0.0 & 6.5 & 1.0 & {---} & 1.5 & {---} & 93.5 & 75.5 & {---} & {---} & 27.0 & 49.0 & 0.0 & 20.9 \\
\bfseries L3 & 7.5 & 31.0 & 0.5 & 3.5 & 0.0 & 13.5 & 8.0 & {---} & 3.5 & {---} & 88.0 & {---} & {---} & {---} & 13.0 & 44.5 & 0.0 & 17.8 \\
\bfseries L4 & {---} & {---} & {---} & {---} & {---} & {---} & {---} & 8.5 & {---} & 0.0 & {---} & {---} & 0.0 & 2.5 & {---} & {---} & {---} & 2.8 \\
\multicolumn{18}{@{}l}{\textit{Concatenate + Obj-Centric}} \\
\bfseries L1 & 42.0 & 93.0 & 6.5 & 0.0 & 0.0 & 26.5 & 34.5 & {---} & 27.5 & {---} & 93.0 & 73.0 & {---} & {---} & 71.0 & 46.5 & 0.0 & 39.5 \\
\bfseries L2 & 0.5 & 15.0 & 7.0 & 0.0 & 0.0 & 15.5 & 1.5 & {---} & 6.5 & {---} & 91.5 & 81.0 & {---} & {---} & 20.0 & 40.5 & 0.0 & 21.5 \\
\bfseries L3 & 7.5 & 27.0 & 7.0 & 2.0 & 0.0 & 8.0 & 6.0 & {---} & 15.0 & {---} & 93.5 & {---} & {---} & {---} & 2.0 & 46.0 & 0.0 & 17.8 \\
\bfseries L4 & {---} & {---} & {---} & {---} & {---} & {---} & {---} & 11.5 & {---} & 1.0 & {---} & {---} & 0.0 & 2.0 & {---} & {---} & {---} & 3.6 \\
\midrule[0.2pt]
\addlinespace[2pt]
\multicolumn{18}{@{}l}{\textit{\textbf{Extreme Difficulty; Can Recover From Mistakes}}} \\
\multicolumn{18}{@{}l}{\textit{Cross-Attn + Obj-Centric}} \\
\bfseries L1 & 32.0 & 41.5 & 17.5 & 81.5 & 9.5 & 17.5 & 39.5 & {---} & 11.5 & {---} & 5.0 & 52.5 & {---} & {---} & 22.5 & 0.0 & 2.0 & 25.6 \\
\bfseries L2 & 5.5 & 22.5 & 9.5 & 74.0 & 5.5 & 20.0 & 3.5 & {---} & 6.5 & {---} & 2.0 & 56.0 & {---} & {---} & 15.5 & 0.0 & 2.5 & 17.2 \\
\bfseries L3 & 20.5 & 26.5 & 19.0 & 45.0 & 6.5 & 20.5 & 22.0 & {---} & 9.5 & {---} & 2.0 & {---} & {---} & {---} & 21.0 & 0.5 & 1.5 & 16.2 \\
\bfseries L4 & {---} & {---} & {---} & {---} & {---} & {---} & {---} & 9.5 & {---} & 1.0 & {---} & {---} & 0.5 & 24.0 & {---} & {---} & {---} & 8.8 \\
\multicolumn{18}{@{}l}{\textit{Concatenate + Obj-Centric}} \\
\bfseries L1 & 17.5 & 47.5 & 31.0 & 93.0 & 17.5 & 29.0 & 22.0 & {---} & 73.0 & {---} & 9.5 & 51.0 & {---} & {---} & 21.5 & 0.5 & 2.5 & 32.0 \\
\bfseries L2 & 21.5 & 42.0 & 37.0 & 93.0 & 15.5 & 31.0 & 11.5 & {---} & 38.5 & {---} & 9.0 & 45.5 & {---} & {---} & 27.0 & 0.5 & 2.5 & 28.8 \\
\bfseries L3 & 28.5 & 49.5 & 37.0 & 73.0 & 13.0 & 24.5 & 22.0 & {---} & 56.5 & {---} & 7.0 & {---} & {---} & {---} & 33.5 & 0.0 & 0.0 & 28.7 \\
\bfseries L4 & {---} & {---} & {---} & {---} & {---} & {---} & {---} & 14.5 & {---} & 18.5 & {---} & {---} & 0.0 & 19.5 & {---} & {---} & {---} & 13.1 \\
\midrule[0.2pt]
\addlinespace[2pt]
\multicolumn{18}{@{}l}{\textit{\textbf{Extremely Distracting Difficulty; Can Recover From Mistakes}}} \\
\multicolumn{18}{@{}l}{\textit{Cross-Attn + Obj-Centric}} \\
\bfseries L1 & 5.0 & 18.5 & 0.5 & 0.0 & 0.0 & 9.5 & 12.5 & {---} & 3.0 & {---} & 2.0 & 48.0 & {---} & {---} & 9.5 & 1.0 & 0.0 & 8.4 \\
\bfseries L2 & 1.5 & 10.5 & 0.5 & 0.0 & 0.0 & 5.5 & 0.0 & {---} & 0.5 & {---} & 1.5 & 46.0 & {---} & {---} & 2.0 & 0.0 & 0.0 & 5.2 \\
\bfseries L3 & 7.0 & 10.0 & 1.5 & 0.0 & 0.0 & 10.5 & 6.5 & {---} & 1.0 & {---} & 1.0 & {---} & {---} & {---} & 2.5 & 0.0 & 0.0 & 3.3 \\
\bfseries L4 & {---} & {---} & {---} & {---} & {---} & {---} & {---} & 7.5 & {---} & 0.0 & {---} & {---} & 1.0 & 0.5 & {---} & {---} & {---} & 2.2 \\
\multicolumn{18}{@{}l}{\textit{Concatenate + Obj-Centric}} \\
\bfseries L1 & 3.0 & 35.0 & 5.5 & 0.0 & 0.0 & 15.0 & 4.5 & {---} & 29.0 & {---} & 11.0 & 50.5 & {---} & {---} & 1.0 & 0.0 & 0.0 & 11.9 \\
\bfseries L2 & 5.0 & 30.5 & 3.5 & 0.0 & 0.0 & 12.0 & 2.5 & {---} & 7.0 & {---} & 9.0 & 49.0 & {---} & {---} & 2.5 & 0.0 & 0.5 & 9.3 \\
\bfseries L3 & 4.0 & 28.0 & 7.0 & 0.0 & 0.0 & 6.5 & 6.0 & {---} & 12.0 & {---} & 7.5 & {---} & {---} & {---} & 3.5 & 0.5 & 0.0 & 6.2 \\
\bfseries L4 & {---} & {---} & {---} & {---} & {---} & {---} & {---} & 7.5 & {---} & 5.0 & {---} & {---} & 0.0 & 0.5 & {---} & {---} & {---} & 3.2 \\
\bottomrule
\end{tabular}
\caption{Per-task average success rate when evaluating performance for models \textbf{\textit{trained} with a randomised object order per observation} for each difficulty level. Models \textbf{can recover from mistakes} and are trained \textbf{with paraphrased instructions}.}
\label{tab:trained-shuffle-full}
\end{table*}

\begin{table*}[tbh]
\centering
\scriptsize
\sisetup{table-format=2.1}
\renewcommand{\arraystretch}{1.2}
\setlength{\tabcolsep}{4.5pt}
\begin{tabular}{@{} l *{18}{S} @{}}
\toprule
& {T1} & {T2} & {T3} & {T4} & {T5} & {T6} & {T7} & {T8} & {T9} & {T10} & {T11} & {T12} & {T13} & {T14} & {T15} & {T16} & {T17} & {Avg.} \\
\midrule
\multicolumn{18}{@{}l}{\textit{\textbf{Permute Object Order; Default Difficulty; No Mistakes Allowed}}} \\
\multicolumn{18}{@{}l}{\textit{Cross-Attn + Obj-Centric}} \\
\bfseries L1 & 69.0 & 85.0 & 5.0 & 90.0 & 3.5 & 37.5 & 71.5 & {---} & 5.5 & {---} & 92.0 & 85.5 & {---} & {---} & 63.5 & 46.0 & 0.0 & 50.3 \\
\bfseries L2 & 18.5 & 31.0 & 5.0 & 73.0 & 2.0 & 38.5 & 16.5 & {---} & 3.0 & {---} & 92.0 & 75.0 & {---} & {---} & 43.0 & 46.0 & 0.0 & 34.1 \\
\bfseries L3 & 32.5 & 44.0 & 10.5 & 46.5 & 2.0 & 31.0 & 27.0 & {---} & 3.5 & {---} & 92.0 & {---} & {---} & {---} & 29.5 & 42.5 & 0.0 & 30.1 \\
\bfseries L4 & {---} & {---} & {---} & {---} & {---} & {---} & {---} & 15.5 & {---} & 0.0 & {---} & {---} & 0.0 & 24.5 & {---} & {---} & {---} & 10.0 \\
\multicolumn{18}{@{}l}{\textit{Concatenate + Obj-Centric}} \\
\bfseries L1 & 73.0 & 96.0 & 24.0 & 92.5 & 11.0 & 39.5 & 75.5 & {---} & 70.5 & {---} & 92.5 & 72.5 & {---} & {---} & 68.5 & 44.0 & 0.0 & 58.4 \\
\bfseries L2 & 24.0 & 39.5 & 30.0 & 87.5 & 7.5 & 39.5 & 14.5 & {---} & 26.5 & {---} & 91.5 & 75.5 & {---} & {---} & 51.5 & 46.0 & 0.0 & 41.0 \\
\bfseries L3 & 20.0 & 42.0 & 28.5 & 75.5 & 8.0 & 19.0 & 14.0 & {---} & 46.5 & {---} & 93.5 & {---} & {---} & {---} & 22.0 & 45.0 & 0.0 & 34.5 \\
\bfseries L4 & {---} & {---} & {---} & {---} & {---} & {---} & {---} & 18.5 & {---} & 0.0 & {---} & {---} & 0.0 & 14.0 & {---} & {---} & {---} & 8.1 \\
\midrule[0.2pt]
\addlinespace[2pt]
\multicolumn{18}{@{}l}{\textit{\textbf{Permute Object Order; Distracting Difficulty; No Mistakes Allowed}}} \\
\multicolumn{18}{@{}l}{\textit{Cross-Attn + Obj-Centric}} \\
\bfseries L1 & 17.5 & 63.5 & 0.5 & 4.0 & 0.0 & 10.5 & 25.0 & {---} & 2.0 & {---} & 90.5 & 68.5 & {---} & {---} & 75.0 & 42.5 & 0.0 & 30.7 \\
\bfseries L2 & 0.0 & 8.5 & 0.0 & 2.5 & 0.0 & 6.0 & 0.0 & {---} & 1.0 & {---} & 92.0 & 63.5 & {---} & {---} & 24.0 & 47.0 & 0.0 & 18.8 \\
\bfseries L3 & 4.0 & 22.0 & 0.0 & 3.0 & 0.0 & 9.0 & 4.0 & {---} & 2.0 & {---} & 87.5 & {---} & {---} & {---} & 7.5 & 44.5 & 0.0 & 15.3 \\
\bfseries L4 & {---} & {---} & {---} & {---} & {---} & {---} & {---} & 3.5 & {---} & 0.0 & {---} & {---} & 0.0 & 2.5 & {---} & {---} & {---} & 1.5 \\
\multicolumn{18}{@{}l}{\textit{Concatenate + Obj-Centric}} \\
\bfseries L1 & 33.5 & 85.0 & 3.0 & 0.0 & 0.0 & 13.5 & 22.5 & {---} & 25.0 & {---} & 82.5 & 61.5 & {---} & {---} & 61.5 & 45.5 & 0.0 & 33.3 \\
\bfseries L2 & 0.0 & 14.0 & 5.5 & 0.0 & 0.0 & 8.0 & 0.0 & {---} & 3.5 & {---} & 87.5 & 62.5 & {---} & {---} & 15.5 & 39.5 & 0.0 & 18.2 \\
\bfseries L3 & 4.5 & 21.5 & 5.0 & 0.0 & 0.0 & 4.5 & 4.5 & {---} & 11.5 & {---} & 91.0 & {---} & {---} & {---} & 1.5 & 46.0 & 0.0 & 15.8 \\
\bfseries L4 & {---} & {---} & {---} & {---} & {---} & {---} & {---} & 7.5 & {---} & 0.0 & {---} & {---} & 0.0 & 1.0 & {---} & {---} & {---} & 2.1 \\
\midrule[0.2pt]
\addlinespace[2pt]
\multicolumn{18}{@{}l}{\textit{\textbf{Permute Object Order; Extreme Difficulty; No Mistakes Allowed}}} \\
\multicolumn{18}{@{}l}{\textit{Cross-Attn + Obj-Centric}} \\
\bfseries L1 & 2.5 & 17.0 & 7.5 & 81.5 & 3.5 & 4.0 & 3.5 & {---} & 10.0 & {---} & 4.5 & 34.5 & {---} & {---} & 13.5 & 0.0 & 0.0 & 14.0 \\
\bfseries L2 & 0.0 & 4.0 & 1.5 & 73.5 & 2.5 & 5.5 & 0.0 & {---} & 3.0 & {---} & 1.5 & 42.5 & {---} & {---} & 12.0 & 0.0 & 0.0 & 11.2 \\
\bfseries L3 & 3.5 & 10.0 & 4.5 & 44.5 & 3.5 & 4.0 & 3.5 & {---} & 5.0 & {---} & 0.5 & {---} & {---} & {---} & 16.5 & 0.5 & 0.0 & 8.0 \\
\bfseries L4 & {---} & {---} & {---} & {---} & {---} & {---} & {---} & 2.5 & {---} & 0.0 & {---} & {---} & 0.5 & 15.0 & {---} & {---} & {---} & 4.5 \\
\multicolumn{18}{@{}l}{\textit{Concatenate + Obj-Centric}} \\
\bfseries L1 & 8.5 & 36.0 & 14.0 & 93.0 & 9.0 & 9.5 & 12.0 & {---} & 68.5 & {---} & 4.0 & 39.5 & {---} & {---} & 16.5 & 0.5 & 0.0 & 23.9 \\
\bfseries L2 & 0.0 & 18.0 & 18.0 & 93.0 & 7.5 & 6.5 & 0.0 & {---} & 22.0 & {---} & 3.0 & 38.0 & {---} & {---} & 17.0 & 0.5 & 0.0 & 17.2 \\
\bfseries L3 & 4.0 & 28.0 & 18.5 & 72.0 & 8.5 & 6.0 & 7.0 & {---} & 45.0 & {---} & 1.5 & {---} & {---} & {---} & 19.0 & 0.0 & 0.0 & 17.5 \\
\bfseries L4 & {---} & {---} & {---} & {---} & {---} & {---} & {---} & 7.0 & {---} & 5.5 & {---} & {---} & 0.0 & 11.0 & {---} & {---} & {---} & 5.9 \\
\midrule[0.2pt]
\addlinespace[2pt]
\multicolumn{18}{@{}l}{\textit{\textbf{Permute Object Order; Extremely Distracting Difficulty; No Mistakes Allowed}}} \\
\multicolumn{18}{@{}l}{\textit{Cross-Attn + Obj-Centric}} \\
\bfseries L1 & 0.0 & 10.5 & 0.0 & 0.0 & 0.0 & 1.0 & 1.5 & {---} & 1.5 & {---} & 2.0 & 33.0 & {---} & {---} & 8.5 & 0.5 & 0.0 & 4.5 \\
\bfseries L2 & 0.0 & 4.0 & 0.0 & 0.0 & 0.0 & 1.5 & 0.0 & {---} & 0.5 & {---} & 0.5 & 34.5 & {---} & {---} & 2.0 & 0.0 & 0.0 & 3.3 \\
\bfseries L3 & 1.5 & 3.5 & 0.5 & 0.0 & 0.0 & 2.0 & 0.5 & {---} & 0.0 & {---} & 0.5 & {---} & {---} & {---} & 1.5 & 0.0 & 0.0 & 0.8 \\
\bfseries L4 & {---} & {---} & {---} & {---} & {---} & {---} & {---} & 1.0 & {---} & 0.0 & {---} & {---} & 1.0 & 0.5 & {---} & {---} & {---} & 0.6 \\
\multicolumn{18}{@{}l}{\textit{Concatenate + Obj-Centric}} \\
\bfseries L1 & 0.0 & 27.0 & 3.0 & 0.0 & 0.0 & 5.0 & 3.0 & {---} & 26.5 & {---} & 3.0 & 36.0 & {---} & {---} & 1.0 & 0.0 & 0.0 & 8.0 \\
\bfseries L2 & 0.0 & 14.5 & 2.0 & 0.0 & 0.0 & 3.0 & 0.0 & {---} & 3.0 & {---} & 4.5 & 38.5 & {---} & {---} & 2.0 & 0.0 & 0.0 & 5.2 \\
\bfseries L3 & 1.5 & 15.0 & 3.0 & 0.0 & 0.0 & 1.0 & 0.5 & {---} & 10.0 & {---} & 2.0 & {---} & {---} & {---} & 3.0 & 0.5 & 0.0 & 3.0 \\
\bfseries L4 & {---} & {---} & {---} & {---} & {---} & {---} & {---} & 1.5 & {---} & 0.0 & {---} & {---} & 0.0 & 0.5 & {---} & {---} & {---} & 0.5 \\
\bottomrule
\end{tabular}
\caption{Per-task average success rate for models \textbf{\textit{trained} with a randomised object order per observation} for each difficulty level, and then \textbf{evaluated with a permuted object order}. Models are trained \textbf{with paraphrased instructions} and are \textbf{not allowed to make mistakes} during evaluation.}
\label{tab:trained-shuffle-full-strict}
\end{table*}
% Diff+Shuf (ParaShuf) w/ strict time limit

\subsection{Exploring Task Success at Higher Difficulty Levels and Masked Instructions}\label{app:per-task-difficulty-masked}

\cref{tab:difficulty-no-prompt-full} shows that model performance drops to 0 for most tasks without instructions, as expected. 
However, T1 (pick-and-place), T2 (pick-and-place from a frame), and particularly T12 (object sweeping) can still be performed.
T12 shows the best performance, followed by T1 and T2, with T12's performance remaining significantly higher than T1 at increased difficulty levels for all models except \textit{Cross-Attn + Obj-Centric}.

T12 is unique in \VIMABench as the only training task requiring sweeping objects into some boundary. 
Without instructions, the model has a 50/50 chance of choosing the correct object type to sweep. Therefore, the model has likely overfit to perform a sweeping action when using a spatula, as it's the only task with this specific end-effector. 
This explains T12's higher performance across difficulty levels and reinforces the claim that without instructions, models rely on spurious correlations learned during training, such as associating the spatula with sweeping, rather than true task understanding.

\end{document}